\documentclass[10pt,twocolumn,letterpaper]{article}

\usepackage[pagenumbers]{cvpr} % To force page numbers, e.g. for an arXiv version

\usepackage{graphicx}
\usepackage{command}
\definecolor{cvprblue}{rgb}{0.21,0.49,0.74}
\usepackage[pagebackref,breaklinks,colorlinks,allcolors=cvprblue]{hyperref}

\title{Improving the Transferability of Adversarial Attacks by an Input Transpose}

\author{Qing Wan\\
{\tt\small frankqingwan@gmail.com}
\and
Shilong Deng\\
{\tt\small shilongdeng12@gmail.com}
\and
Xun Wang\\
{\tt\small wx@zjgsu.edu.cn}
\\
School of Computer Science and Technology, Zhejiang Gongshang University,\\ 
Zhejiang Province 310018, China\\
}

\begin{document}
\maketitle
\begin{abstract}
	Deep neural networks (DNNs) are highly susceptible to adversarial examples--subtle perturbations applied to inputs that are often imperceptible to humans yet lead to incorrect model predictions. In black-box scenarios, however, existing adversarial examples exhibit limited transferability and struggle to effectively compromise multiple unseen DNN models. Previous strategies enhance the cross-model generalization of adversarial examples by introducing versatility into adversarial perturbations, thereby improving transferability. However, further refining perturbation versatility often demands intricate algorithm development and substantial computation consumption. In this work, we propose an input transpose method that requires almost no additional labor and computation costs but can significantly improve the transferability of existing adversarial strategies. Even without adding adversarial perturbations, our method demonstrates considerable effectiveness in cross-model attacks. Our exploration finds that on specific datasets, a mere $1^\circ$ left or right rotation might be sufficient for most adversarial examples to deceive unseen models. Our further analysis suggests that this transferability improvement triggered by rotating only $1^\circ$ may stem from visible pattern shifts in the DNN's low-level feature maps. Moreover, this transferability exhibits optimal angles that, when identified under unrestricted query conditions, could potentially yield even greater performance.
\end{abstract}    
\section{Introduction}
\label{sec:introduction}
The adversarial attack \cite{goodfellow2014explaining} has been one of the pervasive threats to deep neural networks (DNNs). Given an input image, this attack generates adversarial examples (AEs) by adding small perturbations. These AEs hardly deceive our humans but can easily mislead a DNN into wrong decision-making, which raises the security concerns in DNN-based applications. Adversarial attacks \cite{goodfellow2014explaining} can be generally categorized into two types: white-box attacks and black-box attacks. A white-box attack assumes the attacker can access to the target model's architecture, internal parameters (such as weights and biases), or even gradient information used during the model's training (treat the sufferer as a white box). In contrast to the white-box, a black-box attack operates without any access to the target model's internal (treat the sufferer as a black box). It only allows the adversary to interact with the sufferer by querying it through its input and observing its outputs. Therefore, black-box attacks can be more challenging than white-box ones. In practice, the black-box attack can be implemented by AEs generated from a white-box attack, \ie, some AE crafted on a white-box model could also deceive other unseen DNN models \cite{xie2019improving}. Therefore, an AE crafted on one model has transferability if it can be well generalized across other black-box models. Moreover, the attack method is named transferable adversarial attack if it has transferability.

Enhancing the transferability of an AE is challenging. A usual AE only consumes a few steps as long as we fix a white-box model and utilize a conventional iterative method (like FGSM \cite{goodfellow2014explaining} or BIM \cite{kurakin2018adversarial}). However, this AE may soon overfit to the white-box and not generalize well across black-box sufferers. Extensive efforts have proposed numerous strategies to reduce overfitting and enhancing transferability. A recent study \cite{zhang2024bag} concludes that improving the transferability may heavily depend on tuning parameters. Attempts include loss-function or gradient manipulations \cite{dong2019evading, guo2020backpropagating, wang2021feature, ge2023boosting, wu2023gnp, wang2024gi_fgsm}, input variations \cite{xie2019improving, lin2019nesterov, luo2022frequency, zhang2023improving, lin2024boosting, zhu2024learning}, model augmentations \cite{yang2023generating, wang2023diversifying, byun2023introducing}, etc. These methods rely on intricate ideas, manually crafted algorithms, and careful refinements. Inevitably, current strategies may also consume intensive trials (for example, \cite{wang2024ags} demands $17$ hours of RTX 3090 GPU training just to enable its transferable functionality), resulting in unpredictable scalability difficulties, high computational consumption, and excessive labor costs. The current studies of transferable attacks need a fresh perspective.

To improve the transferability of AEs while maintaining budget algorithm development and computational costs, we introduce a novel strategy termed \textit{input transpose}, which involves applying a simple mathematical transpose operation directly to the input. The implementation of our strategy adds only one line of code but achieves a spectacular bump in transferability according to our NIPS'17 (ImageNet) and CIFAR-10 datasets evaluations:
\begin{itemize}
	\item Without any adversarial techniques, over $40\%$ of clean images successfully fool black-box models;
	\item Existing transferable attacks show up to $803\%$ improvement (average $83\%$) on NIPS’17 and up to $117\%$ (average $61\%$) on CIFAR-10 in a single-model setting;
	\item In an ensemble setting, transferability on NIPS’17 also increases, achieving up to $347\%$ improvement with an average of $44\%$.
\end{itemize}

Based on findings from the input transpose strategy, we explore minimizing the input adjustment by taking a $1^\circ$ rotation. The result shows that it improves the transferability of most mainstream transferable attacks with gains up to $26.5$ percentage points (average $6.5$) under the single model setting; under the ensemble setting, the improvement reaches up to $23.8$ (averaged $11.2$). However, this improvement with a $1^\circ$ rotation is observed exclusively on NIPS'17, prompting us to conduct further investigation. Our follow-up experiments suggest that this boost in transferability may stem from visible pattern fluctuations in the DNN's low-level feature maps, and the transferability may benefit from the angle optimization: If querying to black-box sufferers was always permitted, rotating AEs at an angle between $210^\circ$ and $240^\circ$ could enhance transferability beyond our current results.

\section{Related Work}
\label{sec:related_work}

\subsection{Literature review}
The landmark paper \cite{goodfellow2014explaining} introduced the iconic adversarial `Gibbon'. It demonstrates the vulnerability of DNNs to subtle perturbations and inaugurates a prolific era of research into adversarial attacks and model robustness. We mainly focus on transferable, non-targeted black-box attacks. Here, `non-targeted' means we only care about how to fool a DNN and do not care what direction the DNN mode heads. As we mentioned, the setting of transferable black-box attacks usually requires a white-box model. That model is the source model on which many transferable AEs are crafted. In this way, numerous transferable methods have been proposed recently. Their attempts fall into four categories: loss-function or gradient manipulations, input variations, model augmentations, and generative modeling.

Loss-function or gradient manipulations mainly focus on developing new loss functions or refining gradients for generating transferable perturbations. The loss design includes IR \cite{wang2020unified}, GNP \cite{wu2023gnp}, and PGN \cite{ge2023boosting}. IR proposed an interaction loss to decrease interactions between perturbation units and enhance the adversarial transferability. GNP and PGN shared a similar strategy in penalizing the gradient norm of the training loss to find an AE that lies in a flat local region, thus enhancing the adversarial transferability. The gradient refinements include TIM \cite{dong2019evading}, LinBP \cite{guo2020backpropagating}, FIA \cite{wang2021feature}, MultiANDA \cite{fang2024strong}, and GI-FGSM\cite{wang2024gi_fgsm}. TIM developed a Gaussian kernel to convolve and smooth the gradient used for the transferable adversarial perturbation update. LinBP improved the cross-model effectiveness of AEs by ignoring specific nonlinear activations during the backpropagation and strengthening the linearity of gradient signals. FIA optimized the weighted feature maps and normalized gradient from the decomposition of an intermediate layer to achieve higher adversarial transferability. MultiANDA characterized transferable adversarial perturbations from a distribution learned by the asymptotic normality property of stochastic gradient ascend. GI-FGSM employed the global momentum knowledge to mitigate gradient elimination and improve the success rate of transfer attacks.

Input variations diversify the input images to regularize the adversarial perturbation toward stronger models' adaptability. These methods include DIM \cite{xie2019improving}, SIM \cite{lin2019nesterov}, SSM \cite{long2022frequency}, SSA/SSAH \cite{luo2022frequency}, PAM \cite{zhang2023improving}, BSR \cite{wang2024BSR}, L2T \cite{zhu2024learning}, and DeCoWA \cite{lin2024boosting}. DIM/SIM/BSR/DeCoWA relied on randomly resizing, padding, scaling, block shuffling \& rotation, and deformating the input image to regularize the perturbation update and enhance the transferability of adversarial attacks. SIM/SSA/SSAH improved the transferability of AEs by a mixture of additive noise and the suppression of specific frequency components. PAM blended the adversarial images with the original clean images according to specific ratios derived from a pre-trained semantic predictor. L2T adopted a dynamic strategy trained from the policy gradient in reinforcement learning to adjust the input for better adversarial transferability adaptively.

Model augmentations mainly generate AEs using the variants of a white-box model. Methods in this category can simulate more transparent models that slightly differ from the white-box one. Thus more transferable properties can be incorporated into AEs using these methods. It is similar to the ensemble-based white-box attacks. SGM \cite{wu2020skip}, DHF \cite{wang2023diversifying}, and CFM \cite{byun2023introducing} are the typical works in this category. SGM improved adversarial transferability by collecting more gradient flow from skip connections than the usual layer backpropagation. DHF randomly transformed the high-level layer’s features from the AEs and mixed them with clean images to enhance the adversarial transferability. CFM decomposed the intermediate layers in a white-box model to allow the mixture of clean and adversarial features with randomness.

Generative modeling \cite{zhu2024ge} mainly fine tune generative models (like GAN or Diffusion) to learn to generate transferable AEs. Typical works include CDTP \cite{naseer2019cross} that utilized a pretrained GAN to learn domain-invariant adversarial perturbations, and GE-AdvGAN \cite{zhu2024ge} incorporated the frequency-based gradient editing for generating transferable adversarial perturbations.

\subsection{Motivation}
Previous approaches have sought to improve the transferability of AEs by introducing variations in perturbations, aiming to boost cross-model generalization. However, further refining these perturbation variations often requires complex human engineering and substantial computation resources. This challenge motivates us to consider an alternative: Could we employ operations that do not rely on additive ways, thereby bypassing the need for extensive refinement? This line of inquiry motivated our method, which generates transferable AEs by directly modifying the input, offering a novel approach to enhancing transferability in adversarial attacks.
\section{Methodology \& Experiments}
\label{sec:meth_expts}
This section discusses how we conduct the experiments using our methods.

\subsection{Experiment setup}
\label{sec:meth_expts:expt_setup}
\subsubsection{Datasets}
\label{sec:meth_expts:expt_setup:datasets}
Following the previous studies \cite{xie2019improving, dong2019evading, ge2023boosting, wang2024gi_fgsm, fang2024strong}, we use the NIPS'17 Competition dataset (NIPS'17 for short). It is an ImageNet-compatible dataset that contains $1000$ $299 \times 299$ color images randomly drawn from the ImageNet dataset. We also use the CIFAR-10 dataset \cite{krizhevsky2009learning} (CIFAR-10 for short). It includes $10,000$ $32 \times 32$ color images for testing. We use these two datasets for benchmark running.

\subsubsection{Models}
\label{sec:meth_expts:expt_setup:models}
For experiments on NIPS'17, we utilize four widely used standard models: Inception-v3 (Inc-v3) \cite{szegedy2016rethinking}, Inception-v4 (Inc-v4) \cite{szegedy2017inceptionv4}, Inception-ResNet-v2 (IncRes-v2) \cite{szegedy2017inceptionv4}, and ResNet-v2-101 (Res-101) \cite{he2016resnet}. We also utilize three widely recognized defense models, $\text{Inc-v3}_{ens3}$, $\text{Inc-v3}_{ens4}$, and $\text{IncRes-v2}_{ens}$ \cite{tramer2017ensemble}, to evaluate the transferability of adversarial attacks.

For experiments on CIFAR-10, we utilize four mainstream models: VGG \cite{simonyan2014very} with the Batch Normalization technique \cite{ioffe2017batch}, WRN (Wide ResNet) \cite{zagoruyko2016wide}, ResNeXt \cite{xie2017aggregated}, and DenseNet \cite{huang2017densely}. More specifically, they are VGG-19BN, WRN-28-10, ResNeXt-29, and DenseNet-BC. 

These models may serve as either white-box or black-box models (they do not play both roles at the same time), depending on the needs of our experiments.

\subsubsection{Baselines}
\label{sec:meth_expts:expt_setup:baselines}
For experiments on NIPS'17, we select seven frequently used transferable attacks as our baselines: DIM \cite{xie2019improving}, TIM \cite{dong2019evading}, SIM \cite{lin2019nesterov}, PGN \cite{ge2023boosting}, BSR \cite{wang2024BSR}, GI-FGSM \cite{wang2024gi_fgsm}, and MultiANDA \cite{fang2024strong}.

For the experiments on CIFAR-10, we choose the MI-FGSM \cite{guo2020backpropagating} method as our attack baseline.

These baselines come from different categories of the domain which studies transferable attacks. They also utilize similar models that we discussed in Section \ref{sec:meth_expts:expt_setup:models}. We use their source codes for our benchmark comparisons.

\subsubsection{Devices}
\label{sec:meth_expts:expt_setup:devices}
We did our experiments on NVIDIA CUDA 11.8 with cuDNN 8.8.1.3. Some baselines (for example, DIM, TIM, SIM) have source codes developed on TensorFlow 1 \cite{abadi2016tensorflow}. To reproduce the baseline attacks, we built TensorFlow 1.15 with Bazel-0.26.1 from scratch to support their source code packages.

\subsubsection{Parameter settings}
\label{sec:meth_expts:expt_setup:params}
We generally follow the default settings of the baselines.

For experiments on NIPS'17, we set the image size to $299$, the maximum perturbation $\epsilon$ to $16$, the number of iterations to $10$, and the momentum $\mu$ to $1.0$ for common iterations of baselines. DIM has its diversity probability $p=0.5$. TIM has a kernel size $=7$. SIM has $5$ copies. PGN has a sample size $=20$, balance coefficient $=0.5$, and upper bound of neighborhood $=3.0$. BSR has a sample size $=20$, $2 \times 2$ blocks, num\_copies $=20$, and max angles $=0.2$ (Radians). GI-FGSM has pre-convergence iterations $=5$ and global search factor $=10$. MultiANDA has augmentation sample numbers $=25$ and max augmentation $=0.3$.

For experiments on CIFAR-10, we set the iterations to $10$, the momentum $\mu =1.0$, the max epsilon to $4$, the step size to $0.4$. We slightly adjust the parameters (the max epsilon and the step size in \cite{guo2020backpropagating} are $16$ and $1.6$, respectively) to lower the baseline in \cite{guo2020backpropagating}. This adjustment was necessary because the initial baseline was already highly saturated, which makes it hard to observe performance comparisons between baselines and ours.

\subsection{Transpose can enhance the transferability}
\label{sec:meth_expts:transpose}
We discuss the implementation of our transpose. A batch of color images has its shape $[b, h, w, c]$, where $b$ is the batch size. $h, w$ is the height and width of the images, respectively. $c$ is the channels of the images ($c=3$ for colored images). After the transpose operation, the batch of images should have a shape of $[b, w, h, c]$, and the transpose we use is just the usual transpose in Linear Algebra. Both TensorFlow and PyTorch \cite{paszke2019pytorch} provide this transpose for easy invocation. This transpose can also be implemented without any GPU. Then, the transposed images are fed into different black-box models for running benchmarks. We evaluate the transferability of our method by three experiments: 1. Transpose of purely clean images. 2. Transpose of the input AEs under a single model setting. 3. Transpose of the input AEs under an ensemble setting.

\subsubsection{Transpose of clean images}
\label{sec:meth_expts:transpose_of_clean}
Under this setting, we do not add any adversarial noise. Images from each dataset only get a transpose before being fed into black-box models for evaluation. We report the attack success rates ($\%$) for comparisons.

Table \ref{tab:cleantranspose:nips17+cifar10} shows that a simple transpose operation demonstrates considerable effectiveness in cross-model attacks: applying a transpose to clean images increases their transferability from near $0$ to over $40\%$, with almost no additional computation.

\subsubsection{Transpose of AEs (Single model)}
\label{sec:meth_expts:transpose:single}
We use the AEs crafted on a single white-box model for this evaluation. On NIPS'17, we first select one model from the four (Inc-v3, Inc-v4, IncRes-v2, and Res-101) as the white-box model. Then, we loop seven baseline attacks. At each loop, we generate AEs on a baseline and transpose them before feeding them into six other black-box sufferers and evaluating of our method. On CIFAR-10, we repeat the experiment similarly by selecting from another four models (VGG-19BN, WRN-28-10, ResNeXt-29, and DenseNet-BC) and using the MI-FGSM baseline attack. We report the attack success rates ($\%$) with and without a transpose on the input of AEs for comparisons. We do not test any white-box attack since we focus on the transferability that measures the effectiveness of AEs across different, unseen models.

Tables \ref{tab:singlemodel:nips17} and \ref{tab:singlemodel:cifar10} show that a simple transpose can substantially and consistently enhance the transferability of the existing attacks under the single model setting. We calculate ratios ($\frac{\text{our percentages}}{\text{baseline percentages}}$) on values in each table. Statistics in Table \ref{tab:singlemodel:nips17} indicate that the maximum and average ratios are $9.03$ and $1.83$, respectively, suggesting that adding a transpose to the AEs crafted on existing transferable attacks can achieve up to an $803\%$ increase in cross-model transferability on NIPS’17, with an average increase of $83\%$. Table \ref{tab:singlemodel:cifar10} reveals similar findings for CIFAR-10, where this boost reaches up to $117\%$, with an average increase of $61\%$.

\subsubsection{Transpose of AEs (Ensemble model)}
\label{sec:meth_expts:transpose:ensemble}
We use the AEs crafted from an ensemble attack. We perform this evaluation on NIPS'17 only. We generally follow the setting in \cite{lin2019nesterov} and use four white-box models (Inc-v3, Inc-v4, IncRes-v2, and Res-101) as an ensemble for the baseline reproduction. We loop seven baseline attacks, and at each loop, we generate AEs on the ensemble and transpose them before feeding them into three defensive black-box sufferers for our method evaluation. Again, we report the attack success rates for comparisons. We utilize the same ratio calculation in Section \ref{sec:meth_expts:transpose:single} for statistics.

Table \ref{tab:ensembles:nips17} shows consistent improvements comparable to those under the single-model setting. Applying an input transpose to AEs significantly enhances the transferability of existing ensemble attacks across defense models, achieving up to a $347\%$ increase, with an average improvement of $44\%$.

% Exp Transpose tables
\begin{table*}
	\begin{center}
		\begin{tabular}{c>{\rowmac}c>{\rowmac}c>{\rowmac}c>{\rowmac}c>{\rowmac}c>{\rowmac}c>{\rowmac}c>{\rowmac}c}
			\specialrule{1.0pt}{0pt}{0pt}
			Model & Inc-v3 & Inc-v4 & IncRes-v2 & Res-101 & Inc-v3$_{ens3}$ & Inc-v3$_{ens4}$ & IncRes-v2$_{ens}$ & \\
			%\midrule
			Transpose (Clean) & \textbf{50.4}(3.6) & \textbf{45.0}(2.4) & \textbf{43.2}(0.0) & \textbf{50.8}(3.5) & \textbf{53.3}(6.1) & \textbf{54.6}(8.2) & \textbf{46.2}(2.2) &\\
			\specialrule{1.0pt}{0pt}{0pt}
			Model & \multicolumn{2}{c}{VGG-19BN} & \multicolumn{2}{c}{WRN-28-10} & \multicolumn{2}{c}{ResNext-29} & \multicolumn{2}{c}{DenseNet-BC} \\
			%\midrule
			Transpose (Clean) & \multicolumn{2}{c}{\textbf{61.98}(6.66)} & \multicolumn{2}{c}{\textbf{58.37}(3.79)} & \multicolumn{2}{c}{\textbf{55.08}(3.76)} & \multicolumn{2}{c}{\textbf{55.97}(3.32)} \\
			\specialrule{1.0pt}{0pt}{0pt}
		\end{tabular}
		\caption{Success rates (\%) of our transposed attack applied to clean images. The upper subtable presents evaluations across seven models on NIPS'17, while the lower subtable covers four models on CIFAR-10. Values with or without parentheses indicate results using transposed clean images or results using unaltered clean images, respectively. Both subtables demonstrate considerable transferability on unseen sufferers when applying our transpose method to the input.}
		\label{tab:cleantranspose:nips17+cifar10}
	\end{center}
\end{table*}
\begin{table*}
	\begin{center}
		\resizebox{0.98\textwidth}{!}{
			\begin{tabular}{c>{\rowmac}c>{\rowmac}c>{\rowmac}c>{\rowmac}c>{\rowmac}c>{\rowmac}c>{\rowmac}c}
				\specialrule{1.0pt}{0pt}{0pt}
				Model & Attack & Inc-v4 & IncRes-v2 & Res-101 & Inc-v3$_{ens3}$ & Inc-v3$_{ens4}$ & IncRes-v2$_{ens}$ \\
				\midrule 
				\multirow{6}{*}{Inc-v3}& DIM & \textbf{84.3} (65.4) & \textbf{81.2} (57.8) & \textbf{80.8} (51.2) & \textbf{69.6} (27.1) & \textbf{70.4} (27.4) & \textbf{62.8} (13.7) \\ 
				& TIM & \textbf{76.1} (38.2) & \textbf{72.7} (27.8) & \textbf{74.1} (28.4) & \textbf{74.1} (29.9) & \textbf{74.0} (30.9) & \textbf{70.5} (22.5) \\
				& SIM & \textbf{89.9} (75.3) & \textbf{88.0} (72.8) & \textbf{86.4} (66.8) & \textbf{74.4} (35.8) & \textbf{74.9} (35.2) & \textbf{66.3} (19.8) \\
				& PGN & \textbf{93.8} (88.4) & \textbf{93.1} (86.5) & \textbf{92.3} (79.5) & \textbf{83.0} (57.5) & \textbf{83.9} (57.6) & \textbf{76.2} (38.1) \\
				& BSR & \textbf{96.8} (94.2) & \textbf{95.4} (91.2) & \textbf{93.6} (85.3) & \textbf{83.3} (48.4) & \textbf{83.0} (49.4) & \textbf{72.7} (27.4) \\
				& GI-FGSM & \textbf{82.7} (52.6) & \textbf{82.4} (50.7) & \textbf{79.6} (42.3) & \textbf{64.1} (15.5) & \textbf{64.1} (14.9) & \textbf{56.0} (6.2) \\
				& MultiANDA & \textbf{93.0} (87.3) & \textbf{90.8} (84.4) & \textbf{89.0} (75.1) & \textbf{80.0} (48.8) & \textbf{79.3} (47.9) & \textbf{71.7} (30.6) \\
				
				\specialrule{1.0pt}{0pt}{0pt}
				Model & Attack & Inc-v3 & IncRes-v2 & Res-101 & Inc-v3$_{ens3}$ & Inc-v3$_{ens4}$ & IncRes-v2$_{ens}$ \\
				\midrule
				\multirow{6}{*}{Inc-v4} & DIM & \textbf{88.8} (75.3) & \textbf{84.6} (62.5) & \textbf{83.3} (53.9) & \textbf{70.8} (29.4) & \textbf{70.2} (28.3) & \textbf{63.2} (17.5) \\
				& TIM & \textbf{80.8} (46.6) & \textbf{73.3} (31.4) & \textbf{74.7} (33.1) & \textbf{75.4} (30.9) & \textbf{74.6} (32.8) & \textbf{69.9} (24.6) \\
				& SIM & \textbf{93.4} (84.2) & \textbf{90.2} (77.0) & \textbf{89.3} (70.0) & \textbf{77.8} (43.9) & \textbf{76.7} (41.1) & \textbf{69.6} (26.9) \\
				& PGN & \textbf{95.4} (92.1) & \textbf{92.0} (88.1) & \textbf{92.5} (83.4) & \textbf{86.2} (66.2) & \textbf{84.6} (64.7) & \textbf{78.6} (49.7) \\
				& BSR & \textbf{95.7} (94.0) & \textbf{93.9} (89.5) & \textbf{93.2} (84.9) & \textbf{83.5} (55.3) & \textbf{82.7} (52.4) & \textbf{75.1} (33.5) \\
				& GI-FGSM & \textbf{86.7} (68.0) & \textbf{82.2} (55.3) & \textbf{81.5} (49.5) & \textbf{65.4} (16.2) & \textbf{64.3} (14.7) & \textbf{58.2} (7.7) \\
				& MultiANDA & \textbf{94.0} (89.5) & \textbf{91.3} (84.5) & \textbf{90.0} (75.5) & \textbf{81.8} (53.0) & \textbf{79.4} (52.2) & \textbf{73.5} (35.9) \\
				
				\specialrule{1.0pt}{0pt}{0pt}
				Model & Attack & Inc-v3 & Inc-v4 & Res-101 & Inc-v3$_{ens3}$ & Inc-v3$_{ens4}$ & IncRes-v2$_{ens}$ \\
				\midrule
				\multirow{6}{*}{IncRes-v2} & DIM & \textbf{90.8} (79.3) & \textbf{88.3} (74.2) & \textbf{85.0} (65.9) & \textbf{75.6} (40.5) & \textbf{74.1} (36.0) & \textbf{69.9} (28.8) \\
				& TIM & \textbf{83.5} (56.1) & \textbf{81.2} (50.6) & \textbf{79.6} (43.1) & \textbf{77.5} (43.3) & \textbf{77.8} (42.8) & \textbf{75.7} (41.8) \\
				& SIM & \textbf{94.3} (86.3) & \textbf{92.1} (81.9) & \textbf{90.8} (76.7) & \textbf{82.3} (52.6) & \textbf{79.9} (47.3) & \textbf{75.5} (38.4) \\
				& PGN & \textbf{95.6} (93.8) & \textbf{93.6} (92.5) & \textbf{93.1} (88.7) & \textbf{88.6} (76.4) & \textbf{86.4} (70.6) & \textbf{83.8} (66.9) \\
				& BSR & \textbf{96.4} (95.2) & \textbf{96.5} (93.9) & \textbf{95.2} (89.9) & \textbf{89.6} (71.2) & \textbf{86.5} (65.2) & \textbf{82.1} (51.6) \\
				& GI-FGSM & \textbf{89.3} (68.6) & \textbf{84.9} (60.8) & \textbf{81.9} (48.5) & \textbf{66.7} (15.5) & \textbf{65.0} (15.3) & \textbf{57.0} (9.4) \\
				& MultiANDA & \textbf{95.7} (92.3) & \textbf{95.0} (90.4) & \textbf{92.0} (84.8) & \textbf{83.1} (62.0) & \textbf{82.3} (58.7) & \textbf{77.1} (47.2) \\
				
				\specialrule{1.0pt}{0pt}{0pt}
				Model & Attack & Inc-v3 & Inc-v4 & IncRes-v2 & Inc-v3$_{ens3}$ & Inc-v3$_{ens4}$ & IncRes-v2$_{ens}$ \\
				\midrule
				\multirow{6}{*}{Res-101} & DIM & \textbf{90.9} (82.9) & \textbf{88.8} (76.9) & \textbf{88.5} (77.7) & \textbf{80.4} (50.2) & \textbf{78.6} (45.0) & \textbf{72.4} (31.7) \\
				& TIM & \textbf{81.0} (48.6) & \textbf{76.7} (42.0) & \textbf{75.4} (35.3) & \textbf{79.0} (40.5) & \textbf{79.8} (39.8) & \textbf{74.4} (33.3) \\
				& SIM & \textbf{89.4} (78.5) & \textbf{87.1} (71.9) & \textbf{87.5} (72.0) & \textbf{77.4} (43.9) & \textbf{75.9} (38.8) & \textbf{70.2} (25.3) \\
				& PGN & \textbf{93.6} (87.2) & \textbf{92.0} (83.9) & \textbf{92.2} (83.7) & \textbf{88.4} (71.7) & \textbf{86.3} (69.2) & \textbf{82.8} (58.7) \\
				& BSR & \textbf{96.9} (95.4) & \textbf{96.5} (93.9) & \textbf{95.6} (93.9) & \textbf{90.8} (75.6) & \textbf{89.2} (71.1) & \textbf{83.4} (52.8) \\
				& GI-FGSM & \textbf{88.2} (69.7) & \textbf{84.6} (61.9) & \textbf{85.0} (60.4) & \textbf{69.0} (23.6) & \textbf{68.0} (21.3) & \textbf{60.3} (11.5) \\
				& MultiANDA & \textbf{95.7} (95.3) & \textbf{94.7} (93.2) & \textbf{95.1} (93.2) & \textbf{88.1} (72.2) & \textbf{85.7} (67.1) & \textbf{79.4} (52.6) \\
				\specialrule{1.0pt}{0pt}{0pt}
		\end{tabular}}
		\caption{Success rates (\%) of transferable attacks (with v.s. without transpose) on seven models under the single model setting (NIPS'17). Values without parentheses denote our results using the input transpose; Values with parentheses denote the baselines. The AEs are crafted on Inc-v3, Inc-v4, IncRes-v2, and Res-101, respectively. Please note that we report black-box attacks only. We do not test any white-box attack since we focus on the transferability that measures the effectiveness of AEs across unseen sufferers.}
		\label{tab:singlemodel:nips17}
	\end{center}
\end{table*}

\begin{table*}
	\begin{center}
		\begin{tabular}{c>{\rowmac}c>{\rowmac}c>{\rowmac}c>{\rowmac}c}
			\specialrule{1.0pt}{0pt}{0pt}
			Model & Attack & WRN-28-10 & ResNeXt-29 & DenseNet-BC \\
			\midrule
			VGG-19BN & MI-FGSM & \textbf{71.36} (40.32) & \textbf{71.16} (39.35) & \textbf{71.45} (37.59) \\
			
			\specialrule{1.0pt}{0pt}{0pt}
			Model & Attack & VGG-19BN & ResNeXt-29 & DenseNet-BC \\
			WRN-28-10 & MI-FGSM &\textbf{70.64} (37.07) & \textbf{73.23} (52.5) & \textbf{73.32} (52.15) \\
			
			\specialrule{1.0pt}{0pt}{0pt}
			Model & Attack & VGG-19BN & WRN-28-20 & DenseNet-BC \\
			ResNeXt-29 & MI-FGSM & \textbf{72.63} (40.32) & \textbf{76.65} (59.66) & \textbf{78.19} (66.5) \\
			
			\specialrule{1.0pt}{0pt}{0pt}
			Model & Attack & VGG-19BN & WRN-28-10 & ResNeXt-29 \\
			DenseNet-BC & MI-FGSM & \textbf{72.55} (33.5) & \textbf{76.52} (54.2) & \textbf{79.35} (64.16) \\
			\specialrule{1.0pt}{0pt}{0pt}
		\end{tabular}
		\caption{Success rates (\%) of transferable attacks (with v.s. without transpose) on four models under the single model setting (CIFAR-10). Please refer to Table \ref{tab:singlemodel:nips17} for the usage of parentheses.}
		\label{tab:singlemodel:cifar10}
	\end{center}
\end{table*}
\begin{table}
	\begin{center}
		\begin{tabular}{c>{\rowmac}c>{\rowmac}c>{\rowmac}c}
			\specialrule{1.0pt}{0pt}{0pt}
			Attack & Inc-v3$_{ens3}$ & Inc-v3$_{ens4}$ & IncRes-v2$_{ens}$ \\
			\midrule
			DIM & \textbf{89.1} (79.0) & \textbf{87.0} (74.6) & \textbf{81.9} (59.7) \\ 
			TIM & \textbf{92.0} (87.1) & \textbf{93.0} (86.6) & \textbf{91.0} (83.0) \\
			SIM & \textbf{91.6} (83.4) & \textbf{90.5} (79.0) & \textbf{84.8} (63.1) \\
			PGN & \textbf{94.2} (89.8) & \textbf{92.7} (88.4) & \textbf{89.6} (83.4) \\
			BSR & \textbf{96.4} (91.8) & \textbf{94.6} (89.2) & \textbf{90.7} (76.5) \\
			GI-FGSM & \textbf{71.5} (28.6) & \textbf{69.8} (26.5) & \textbf{62.6} (14.0) \\
			MultiANDA & \textbf{88.3} (77.7) & \textbf{87.4} (73.8) & \textbf{80.8} (58.1) \\
			\specialrule{1.0pt}{0pt}{0pt}
		\end{tabular}
		\caption{Success rates (\%) of transferable attacks (with v.s. without transpose) on seven models under the ensemble setting (NIPS'17). The AEs are crafted on an ensemble of four white-box models (Inc-v3, Inc-v4, IncRes-v2, and Res-101).}
		\label{tab:ensembles:nips17}
	\end{center}
\end{table}
\section{Further Studies}
\label{sec:further_study}

\subsection{$1^\circ$ rotation can improve the transferability}
\label{sec:further_study:1_degree_rotation}
Our previous results demonstrate that applying a transpose to an input significantly enhances the transferability of AEs. Since a transpose operation is equivalent to a combination of a left-right flip and a $90^\circ$ left rotation, this raises the question of whether more minor modifications to the input can similarly improve adversarial transferability. In this subsection, we further explore the impact of minor input operations on the transferability of adversarial attacks.

We repeat the evaluations conducted in Sections \ref{sec:meth_expts:transpose:single} (single model setting) and \ref{sec:meth_expts:transpose:ensemble} (ensemble model setting), but this time, we replace the transpose operation with two rotations: $1^\circ$ left and $1^\circ$ right. The results are presented using the same table format from those sections, with the attack success rates of the two rotations reported jointly using a slash (`/') symbol.

Tables \ref{tab:rotations:nips17} and \ref{tab:rotation_ensembles:nips17} demonstrate that a minimal $1^\circ$ left or right rotation can improve the transferability of most existing attacks on NIPS'17. In Table \ref{tab:rotations:nips17}, $84.23\%$ of our attack success rates outperform baseline values, improving baseline performance by up to $26.5$ percentage points (calculated as the difference between our results and baseline values), with an average increase of $6.5$ percentage points. This trend is even more pronounced in Table \ref{tab:rotation_ensembles:nips17}, where all results exceed baselines, achieving improvements of up to $23.8$ percentage points with an average increase of $11.2$ percentage points.

In contrast, on CIFAR-10, the results with a $1^\circ$ rotation show no improvement and are identical to the transpose benchmark reported in Table \ref{tab:singlemodel:cifar10}.

% Exp Rotation tables
\begin{table*}
	\begin{center}
		\resizebox{1.0\textwidth}{!}{
			\begin{tabular}{c>{\rowmac}c>{\rowmac}c>{\rowmac}c>{\rowmac}c>{\rowmac}c>{\rowmac}c>{\rowmac}c}
				\specialrule{1.0pt}{0pt}{0pt}
				Model & Attack & Inc-v4 & IncRes-v2 & Res-101 & Inc-v3$_{ens3}$ & Inc-v3$_{ens4}$ & IncRes-v2$_{ens}$ \\
				\midrule 
				\multirow{6}{*}{Inc-v3} & DIM & \textbf{69.8/68.4} (65.4) & \textbf{60.0/60.1} (57.8) & \textbf{52.9/52.6} (51.2) & \textbf{37.9/37.7} (27.1) & \textbf{38.1/37.5} (27.4) & \textbf{23.8/24.3} (13.7) \\ 
				& TIM & \textbf{41.9/41.6} (38.2) & \textbf{31.0/29.2} (27.8) & \textbf{33.5/32.0} (28.4) & \textbf{33.8/33.9} (29.9) & \textbf{33.4/34.9} (30.9) & \textbf{25.0/24.6} (22.5) \\
				& SIM & 73.9/73.3 (75.3) & 69.9/70.9 (72.8) & 66.8/\textbf{67.7} (66.8) & \textbf{48.3/50.0} (35.8) & \textbf{47.8/47.8} (35.2) & \textbf{32.2/32.8} (19.8) \\
				& PGN & \textbf{89.0/88.5} (88.4) & 84.6/84.5 (86.5) & \textbf{81.5/82.6} (79.5) & \textbf{73.5/74.0} (57.5) & \textbf{71.9/72.4} (57.6) & \textbf{57.2/58.3} (38.1) \\
				& BSR & \textbf{95.5/95.3} (94.2) & \textbf{91.7/93.1} (91.2) & \textbf{87.9/87.7} (85.3) & \textbf{72.5/71.4} (48.4) & \textbf{68.5/68.2} (49.4) & \textbf{51.8/51.4} (27.4) \\
				& GI-FGSM & \textbf{53.3/52.7} (52.6) & 48.9/47.1 (50.7) & \textbf{43.0/42.9} (42.3) & \textbf{23.8/21.8} (15.5) & \textbf{23.6/23.1} (14.9) & \textbf{13.0/12.3} (6.2) \\
				& MultiANDA & \textbf{89.2/88.3} (87.3) & 83.9/\textbf{84.8} (84.4) & \textbf{78.1/76.4} (75.1) & \textbf{63.3/63.3} (48.8) & \textbf{62.3/61.0} (47.9) & \textbf{48.7/47.0} (30.6) \\
				
				\specialrule{1.0pt}{0pt}{0pt}
				Model & Attack & Inc-v3 & IncRes-v2 & Res-101 & Inc-v3$_{ens3}$ & Inc-v3$_{ens4}$ & IncRes-v2$_{ens}$ \\
				\midrule
				\multirow{6}{*}{Inc-v4} & DIM & \textbf{77.1/76.6} (75.3) & \textbf{64.6/64.4} (62.5) & \textbf{56.5/56.1} (53.9) & \textbf{40.4/40.4} (29.4) & \textbf{36.8/38.2} (28.3) & \textbf{28.1/27.2} (17.5) \\ 
				& TIM & \textbf{50.3/50.7} (46.6) & \textbf{34.3/34.9} (31.4) & \textbf{34.8/35.7} (33.1) & \textbf{34.0/34.6} (30.9) & \textbf{35.7/36.3} (32.8) & \textbf{28.8/27.8} (24.6) \\
				& SIM & 83.2/83.3 (84.2) & 76.0/75.6 (77.0) & 69.0/69.1 (70.0) & \textbf{55.0/53.4} (43.9) & \textbf{51.9/51.9} (41.1) & \textbf{43.2/43.6} (26.9) \\
				& PGN & 91.3/91.4 (92.1) & 87.1/86.9 (88.1) & 82.8/83.2 (83.4) & \textbf{77.1/76.7} (66.2) & \textbf{75.9/76.1} (64.7) & \textbf{67.3/66.8} (49.7) \\
				& BSR & \textbf{94.3/94.6} (94.0) & \textbf{91.6/91.1} (89.5) & \textbf{86.1/85.9} (84.9) & \textbf{73.1/72.5} (55.3) & \textbf{70.7/71.4} (52.4) & \textbf{57.5/56.9} (33.5) \\
				& GI-FGSM & 65.7/64.6 (68.0) & 51.5/50.5 (55.3) & 48.1/48.8 (49.5) & \textbf{23.6/24.1} (16.2) & \textbf{25.3/24.1} (14.7) & \textbf{16.3/15.7} (7.7) \\
				& MultiANDA & \textbf{91.3/90.9} (89.5) & \textbf{85.2/86.2} (84.5) & \textbf{77.6/75.9} (75.5) & \textbf{66.4/66.2} (53.0) & \textbf{62.8/63.9} (52.2) & \textbf{53.4/54.1} (35.9) \\
				
				\specialrule{1.0pt}{0pt}{0pt}
				Model & Attack & Inc-v3 & Inc-v4 & Res-101 & Inc-v3$_{ens3}$ & Inc-v3$_{ens4}$ & IncRes-v2$_{ens}$ \\
				\midrule
				\multirow{6}{*}{IncRes-v2} & DIM & \textbf{81.9/83.0} (79.3) & \textbf{78.2/78.6} (74.2) & \textbf{69.8/69.1} (65.9) & \textbf{55.6/53.4} (40.5) & \textbf{49.9/49.6} (36) & \textbf{47.4/46.9} (28.8) \\ 
				& TIM & \textbf{58.4/58.5} (56.1) & \textbf{54.7/53.5} (50.6) & \textbf{47.0/46.1} (43.1) & \textbf{46.6/47.3} (43.3) & \textbf{47.4/46.5} (42.8) & \textbf{45.1/45.2} (41.8) \\
				& SIM & 85.3/84.4 (86.3) & \textbf{82.2/82.1} (81.9) & 76.3/75.1 (76.7) & \textbf{63.6/64.4} (52.6) & \textbf{58.5/58.0} (47.3) & \textbf{55.0/55.4} (38.4) \\
				& PGN & 92.7/92.7 (93.8) & \textbf{93.0}/92.2 (92.5) & 88.3/\textbf{89.0} (88.7) & \textbf{85.0/83.5} (76.4) & \textbf{81.6/80.7} (70.6) & \textbf{80.4/79.7} (66.9) \\
				& BSR & \textbf{95.3}/94.3 (95.2) & \textbf{94.7/95.2} (93.9) & \textbf{91.6/91.4} (89.9) & \textbf{84.1/84.2} (71.2) & \textbf{80.9/80.5} (65.2) & \textbf{76.6/75.2} (51.6) \\
				& GI-FGSM & 66.7/67.4 (68.6) & \textbf{61.6}/59.5 (60.8) & \textbf{50.3/49.4} (48.5) & \textbf{25.1/24.3} (15.5) & \textbf{26.5/24.8} (15.3) & \textbf{18.6/18.1} (9.4) \\
				& MultiANDA & \textbf{93.0/93.0} (92.3) & \textbf{91.2/91.6} (90.4) & \textbf{86.3/85.5} (84.8) & \textbf{75.6/75.3} (62.0) & \textbf{70.1/72.3} (58.7) & \textbf{69.1/68.6} (47.2) \\
				
				\specialrule{1.0pt}{0pt}{0pt}
				Model & Attack & Inc-v3 & Inc-v4 & IncRes-v2 & Inc-v3$_{ens3}$ & Inc-v3$_{ens4}$ & IncRes-v2$_{ens}$ \\
				\midrule
				\multirow{6}{*}{Res-101} & DIM & \textbf{83.5}/82.6 (82.9) & \textbf{79.3/79.5} (76.9) & \textbf{80.1/79.2} (77.7) & \textbf{65.4/64.2} (50.2) & \textbf{59.6/58.9} (45.0) & \textbf{50.4/50.2} (31.7) \\ 
				& TIM & \textbf{53.1/51.8} (48.6) & \textbf{44.7/46.2} (42.0) & \textbf{41.2/40.6} (35.3) & \textbf{43.2/43.7} (40.5) & \textbf{45.0/47.8} (39.8) & \textbf{37.7/38.4} (33.3) \\
				& SIM & 78.2/76.7 (78.5) & 71.1/\textbf{72.0} (71.9) & 71.1/71.5 (72.0) & \textbf{54.9/53.9} (43.9) & \textbf{51.8/52.8} (38.8) & \textbf{40.6/41.1} (25.3) \\
				& PGN & \textbf{88.2/88.5} (87.2) & \textbf{84.2}/82.8 (83.9) & \textbf{84.2/84.3} (83.7) & \textbf{80.3/77.9} (71.7) & \textbf{77.5/78.6} (69.2) & \textbf{69.7/69.7} (58.7) \\
				& BSR & \textbf{95.7/95.5} (95.4) & \textbf{94.3/94.3} (93.9) & \textbf{94.0/94.2} (93.9) & \textbf{88.1/88.5} (75.6) & \textbf{86.4/85.3} (71.1) & \textbf{78.3/79.3} (52.8) \\
				& GI-FGSM & 67.9/66.8 (69.7) & 61.5/61.7 (61.9) & 59.3/59.9 (60.4) & \textbf{34.7/34.6} (23.6) & \textbf{34.1/33.8} (21.3) & \textbf{22.9/22.4} (11.5) \\
				& MultiANDA & \textbf{95.7/95.6} (95.3) & \textbf{94.2/93.8} (93.2) & \textbf{93.4/94.2} (93.2) & \textbf{84.3/84.0} (72.2) & \textbf{81.5/79.8} (67.1) & \textbf{73.2/71.9} (52.6) \\
				
				\specialrule{1.0pt}{0pt}{0pt}
			\end{tabular}}
		\caption{Success rates (\%) of transferable attack (with v.s. without $1^\circ$ rotation) on seven models under the single model setting (NIPS'17). A value before or after `/' without any parenthese denotes the result with a $1^\circ$ left or $1^\circ$ right rotation to the input, respectively. We bold a value that exceeds the baseline to highlight the improvement.}
		\label{tab:rotations:nips17}
	\end{center}
\end{table*}

\begin{table}
	\begin{center}
		\resizebox{0.48\textwidth}{!}{
		\begin{tabular}{c>{\rowmac}c>{\rowmac}c>{\rowmac}c}
			\specialrule{1.0pt}{0pt}{0pt}
			Attack & Inc-v3$_{ens3}$ & Inc-v3$_{ens4}$ & IncRes-v2$_{ens}$ \\
			\midrule
			DIM & \textbf{91.3/90.1} (79.0) & \textbf{87.0/87.9} (74.6) & \textbf{83.1/82.2} (59.7) \\ 
			TIM & \textbf{89.6/89.7} (87.1) & \textbf{88.5/89.3} (86.6) & \textbf{87.4/87.3} (83.0) \\
			SIM & \textbf{91.5/91.5} (83.4) & \textbf{89.4/88.7} (79.0) & \textbf{82.6/82.8} (63.1) \\
			PGN & \textbf{92.6/92.3} (89.8) & \textbf{91.8/92.0} (88.4) & \textbf{88.4/88.8} (83.4) \\
			BSR & \textbf{97.5/97.6} (91.8) & \textbf{96.0/95.6} (89.2) & \textbf{93.8/93.6} (76.5) \\
			GI-FGSM & \textbf{45.6/44.9} (28.6) & \textbf{44.3/45.3} (26.5) & \textbf{30.5/30.3} (14.0) \\
			MultiANDA & \textbf{90.0/89.2} (77.7) & \textbf{87.2/86.1} (73.8) & \textbf{81.9/80.9} (58.1) \\
			\specialrule{1.0pt}{0pt}{0pt}
		\end{tabular}}
		\caption{Success rates (\%) of transferable attacks (with v.s. without $1^\circ$ rotation) on seven models under the ensemble setting (NIPS'17). Again, AEs are crafted on an ensemble of four white-box models (Inc-v3, Inc-v4, IncRes-v2, and Res-101).}
		\label{tab:rotation_ensembles:nips17}
	\end{center}
\end{table}

\subsection{Why does a $1^\circ$ rotation improve the transferability?}
\label{sec:further_study:why_1_rotation_work}
The improvement from our transpose operation is relatively intuitive, as it alters the input patterns in a way that may disrupt the internal mechanisms of a DNN. However, the transferability gain observed from a $1^\circ$ rotation appears less intuitive, as such a subtle change should not lead to noticeable visual alterations. To investigate this phenomenon, we hypothesize that even a minor angle adjustment could trigger meaningful fluctuations starting from an intermediate layer within the DNN. To test this hypothesis, we conduct an experiment to identify the feature fluctuations that might influence the DNN's prediction.

Using the single-model setting from Section \ref{sec:meth_expts:transpose:single}, we select a specific black-box model and gather feature maps from all its hidden layers for AEs that fail to deceive the model. Then, we repeat this process with AEs that successfully deceive the model after a $1^\circ$ rotation is applied. This process provides two types of feature maps from each hidden layer: one with and one without the $1^\circ$ input rotation. Next, we compute the absolute differences between these two types through a layer-by-layer subtraction, then sort the differences within each layer in a channel-wise descending order, selecting the top-K indices, where K is empirically set to 16 for this experiment. These indices are then used to extract the corresponding channels from each layer’s feature maps for illustration.

\begin{figure*}[!h]
	\centering
	\begin{tabular}{@{}c@{}c@{}c@{}c@{}c@{}}
		\includegraphics[height=0.19\textwidth]{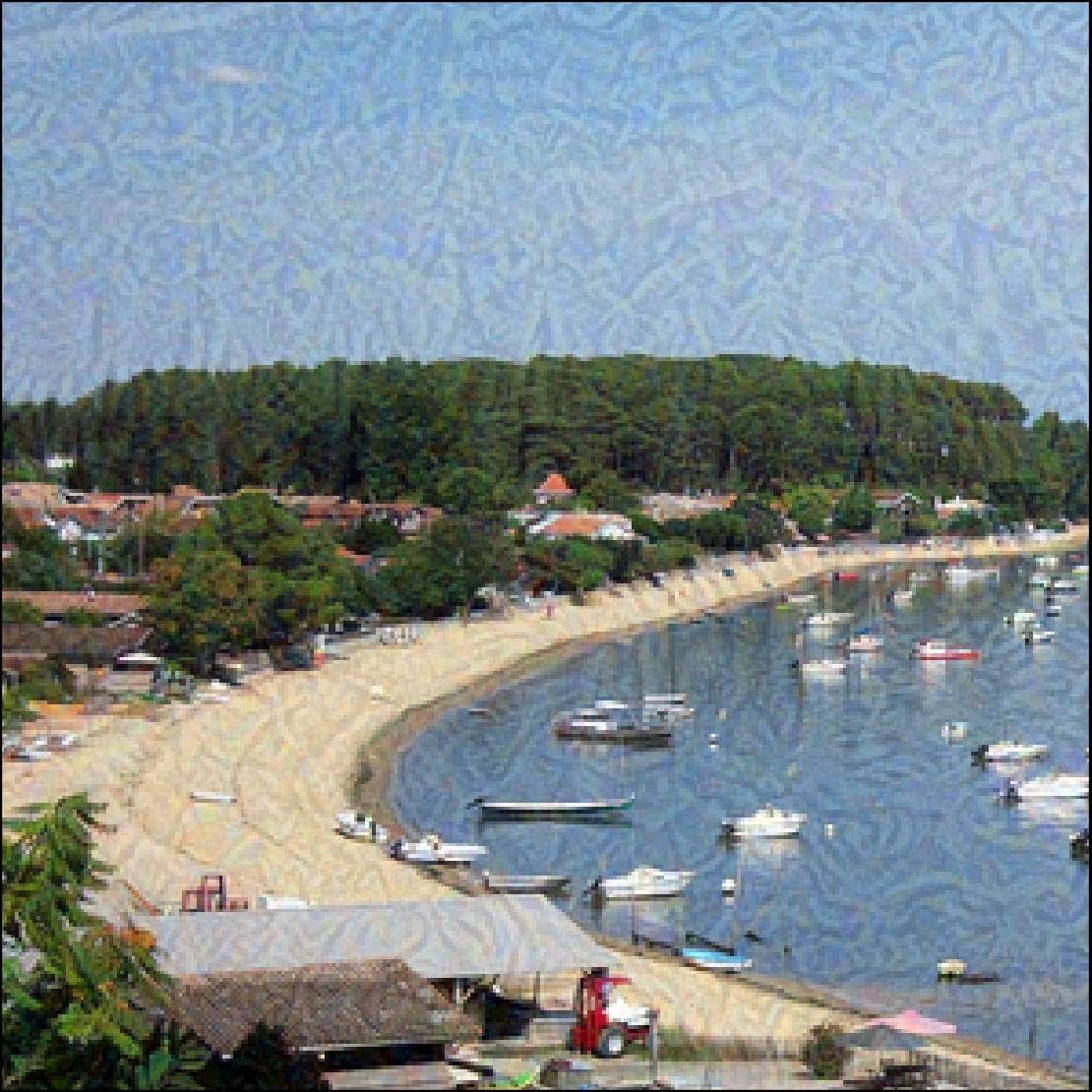} \quad & 
		\includegraphics[height=0.19\textwidth]{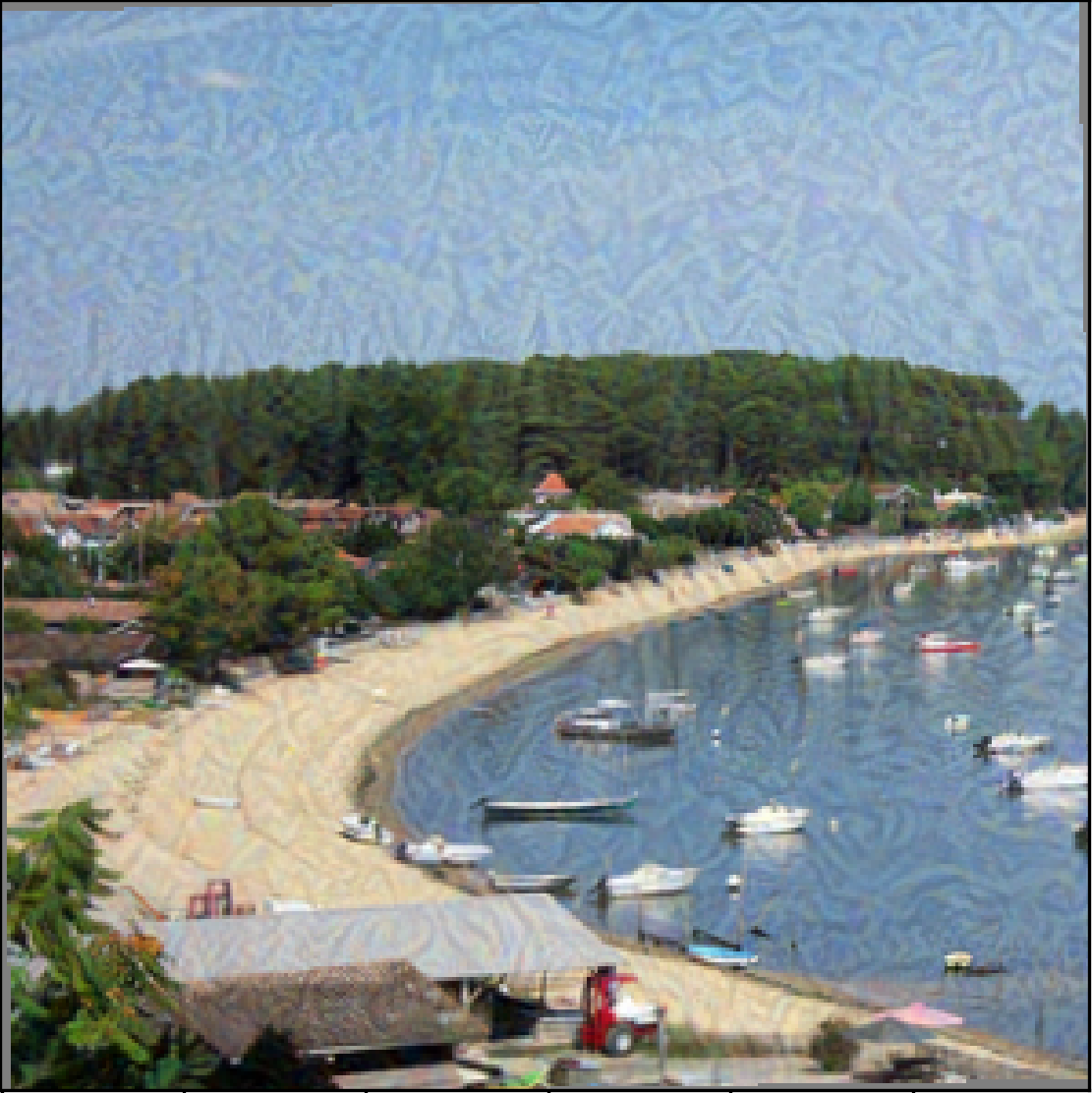} \quad &
		\includegraphics[height=0.19\textwidth]{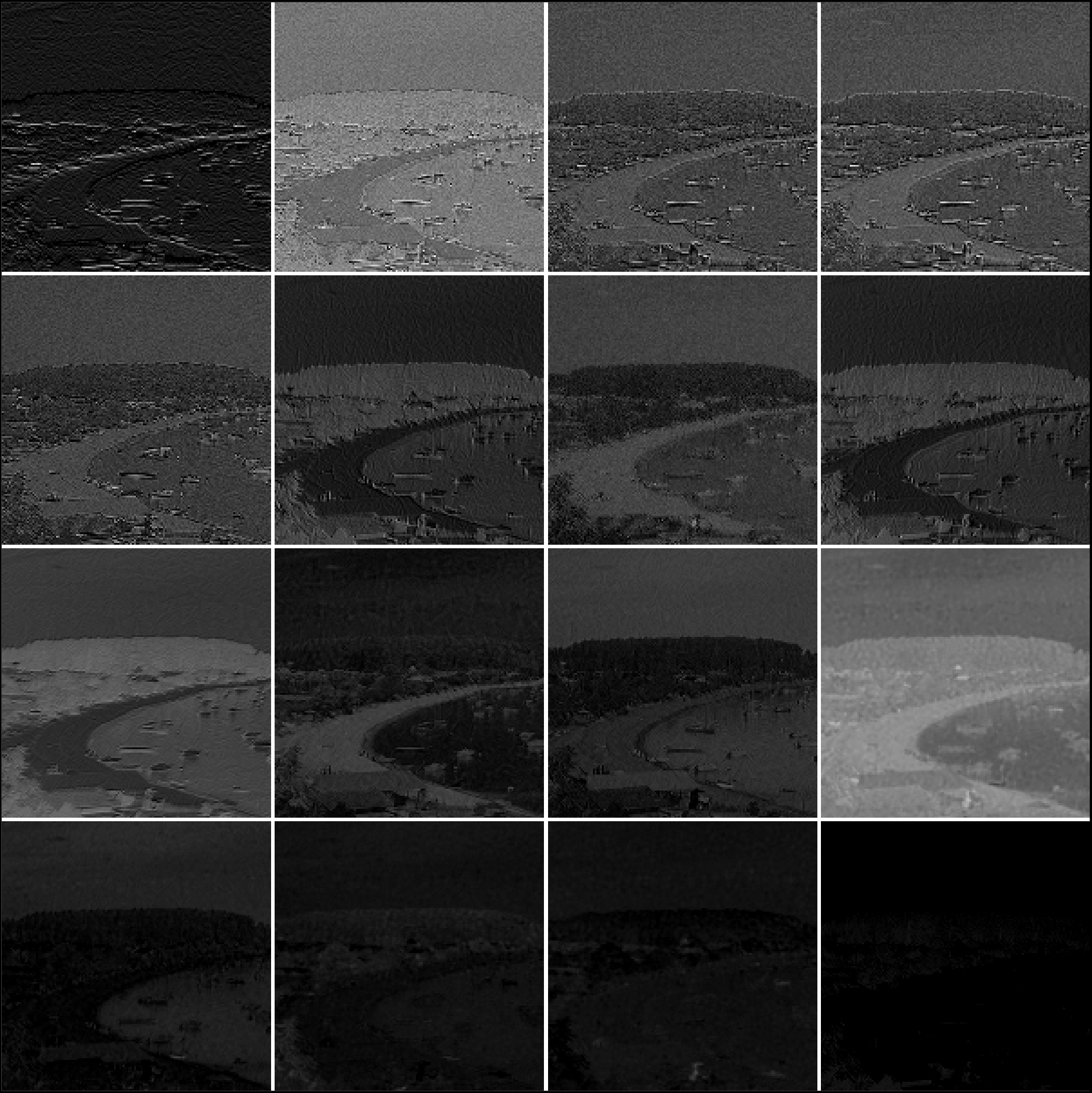} \quad &
		\includegraphics[height=0.19\textwidth]{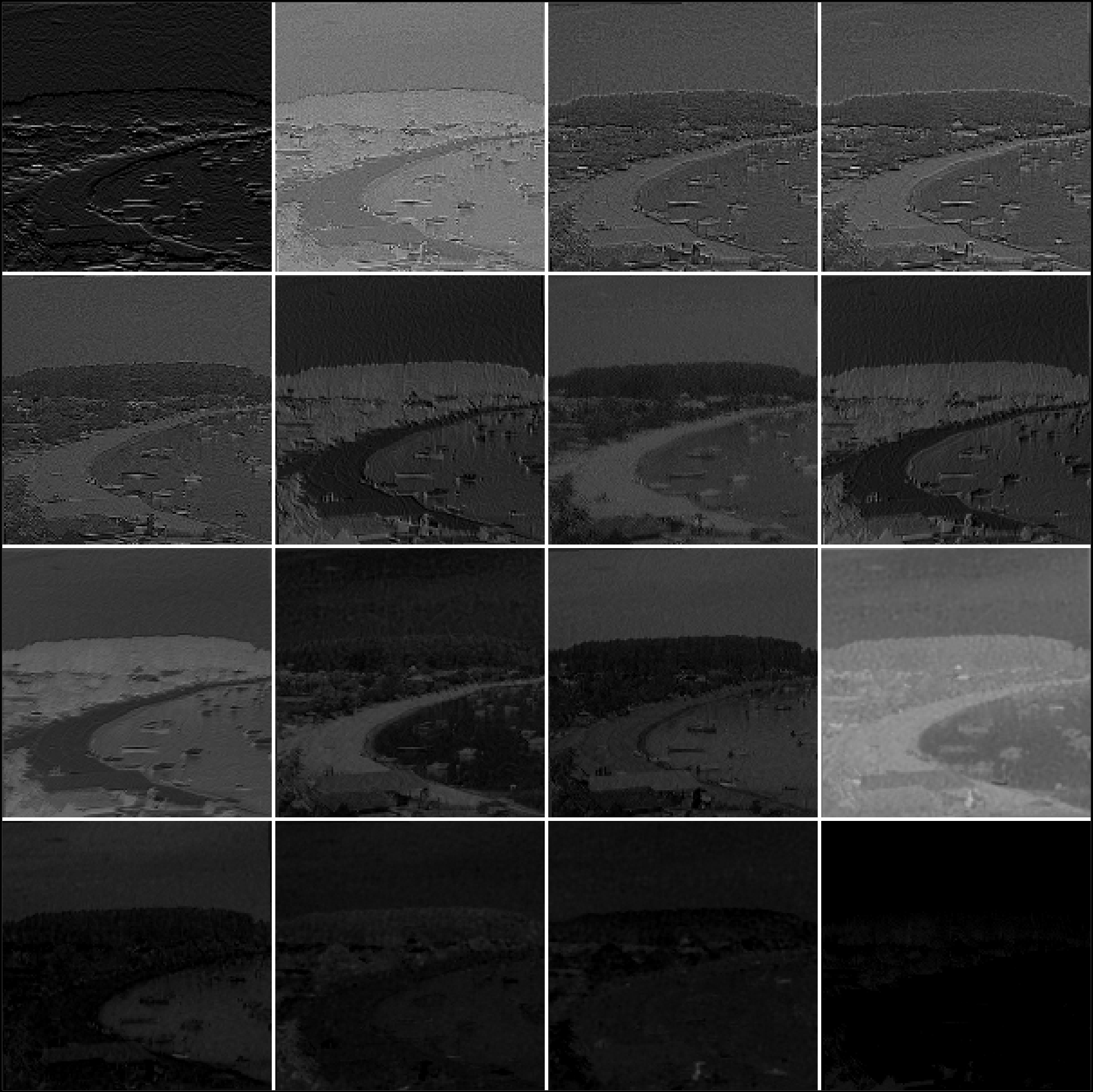} \quad &
		\includegraphics[height=0.19\textwidth]{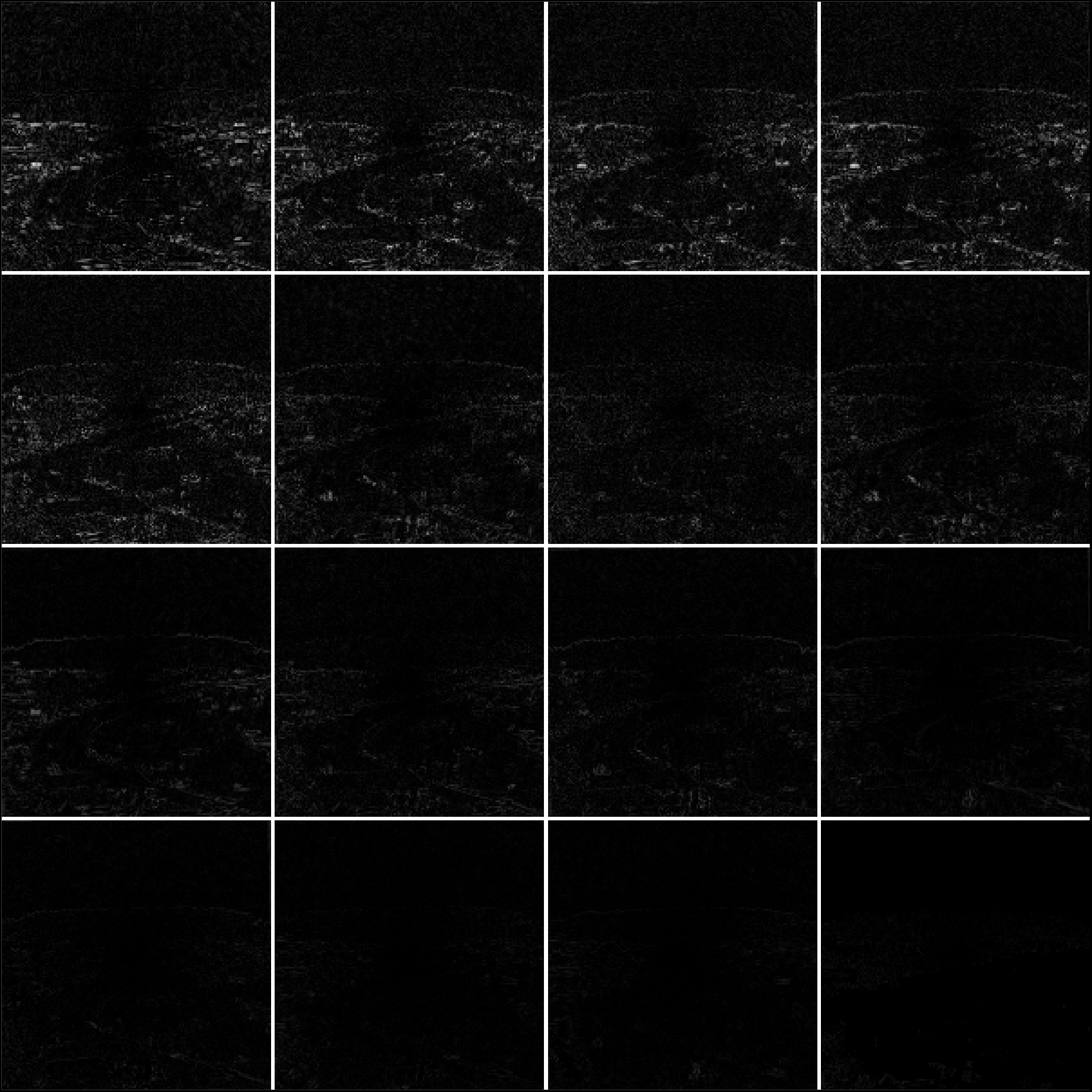} \\
		$(1)$ Advr & $(2)$ $1^\circ$ Left & $(3)$ FMs wo $1^\circ$ & $(4)$ FMs w $1^\circ$ & $(5)$ Diffs between $(3)$ \& $(4)$ \\
	\end{tabular}
	\caption{Identify major feature fluctuations that potentially influence the $\text{Inc-v3}_{ens3}$'s decision under the single model setting. We use the IncRes-v2 as the white-box model and the DIM as the baseline attack. We illustrate five typical subplots: $(1)$ An AE that fails to deceive the $\text{Inc-v3}_{ens3}$. $(2)$ The $1^\circ$ left rotation of $(1)$ that successfully deceives the $\text{Inc-v3}_{ens3}$. $(3)$ Feature maps from $16$ selected channels in the Conv2d-1a layer of the $\text{Inc-v3}_{ens3}$ with $(1)$ as the input. $(4)$ Feature maps of the same $16$ channels with $(2)$ as the input. $(5)$ Absolute differences between $(3)$ and $(4)$, magnified for clarity. Note: Channels in $(3)$ and $(4)$ are selected based on the top 16 indices from $(5)$, ranked by descending mean absolute differences between feature maps with $(1)$ and $(2)$ as the input across all channels in the Conv2d-1a.}
	\label{fig:featureanalysis}
\end{figure*}

We illustrate selected subplots in Figure \ref{fig:featureanalysis}, using IncRes-v2 as the white-box model and DIM as the baseline attack on NIPS’17. In Figure \ref{fig:featureanalysis}, subplots $(1)$ and $(2)$ show an AE without and with a $1^\circ$ left rotation, respectively. Subplots $(3)$ and $(4)$ depict 16-channel feature maps selected from the Conv2d-1a layer (the lowest convolutional layer) of the $\text{Inc-v3}_{ens3}$ model, using $(1)$ and $(2)$ as the model's inputs, respectively. The selection of these channels is based on the top-16 indices retrieved from the subplot $(5)$, where the largest and smallest differences between $(3)$ and $(4)$ appear in the top-left and bottom-right corners, respectively.

Results indicate that $(5)$ exhibits visible patterns in its first five channels, stemming from contour differences between $(3)$ and $(4)$. This observation suggests that our initial hypothesis may have underestimated the impact of a $1^\circ$ rotation, which, in fact, generates non-negligible feature fluctuations even at the low-level layers.

% Exp Feature analysis figs

\subsection{Does a rotation angle have its optimal?}
\label{sec:further_study:optimal_rotation}
Our findings show that a transpose operation can substantially enhance the transferability of baseline attacks on both NIPS’17 and CIFAR-10. Even when reduced to a mere $1^\circ$ rotation, this adjustment still provides transferability improvements for most attacks on NIPS’17. However, this trend does not extend to CIFAR-10, despite the similarities in evaluation settings. This discrepancy suggests that the attack success rate might achieve an optimal level at specific rotation angles, and that this optimal could vary depending on the model or dataset characteristics. In other words, the current level of transferability achieved with a transpose may not yet be fully optimized on these datasets.

To investigate further, we first assume that black-box queries are always permitted and then conduct an experiment to empirically identify the optimal rotation angle for transferability. Using the single-model setting, we rotate AEs crafted with a baseline attack in $10^\circ$ counter-clockwise increments, recording each (angle, attack success rate) pair across six black-box models. This yields a success rate curve for each black-box model over a complete 360° rotation.

\begin{figure}[t]
	\centering
	\includegraphics[width=0.9\linewidth]{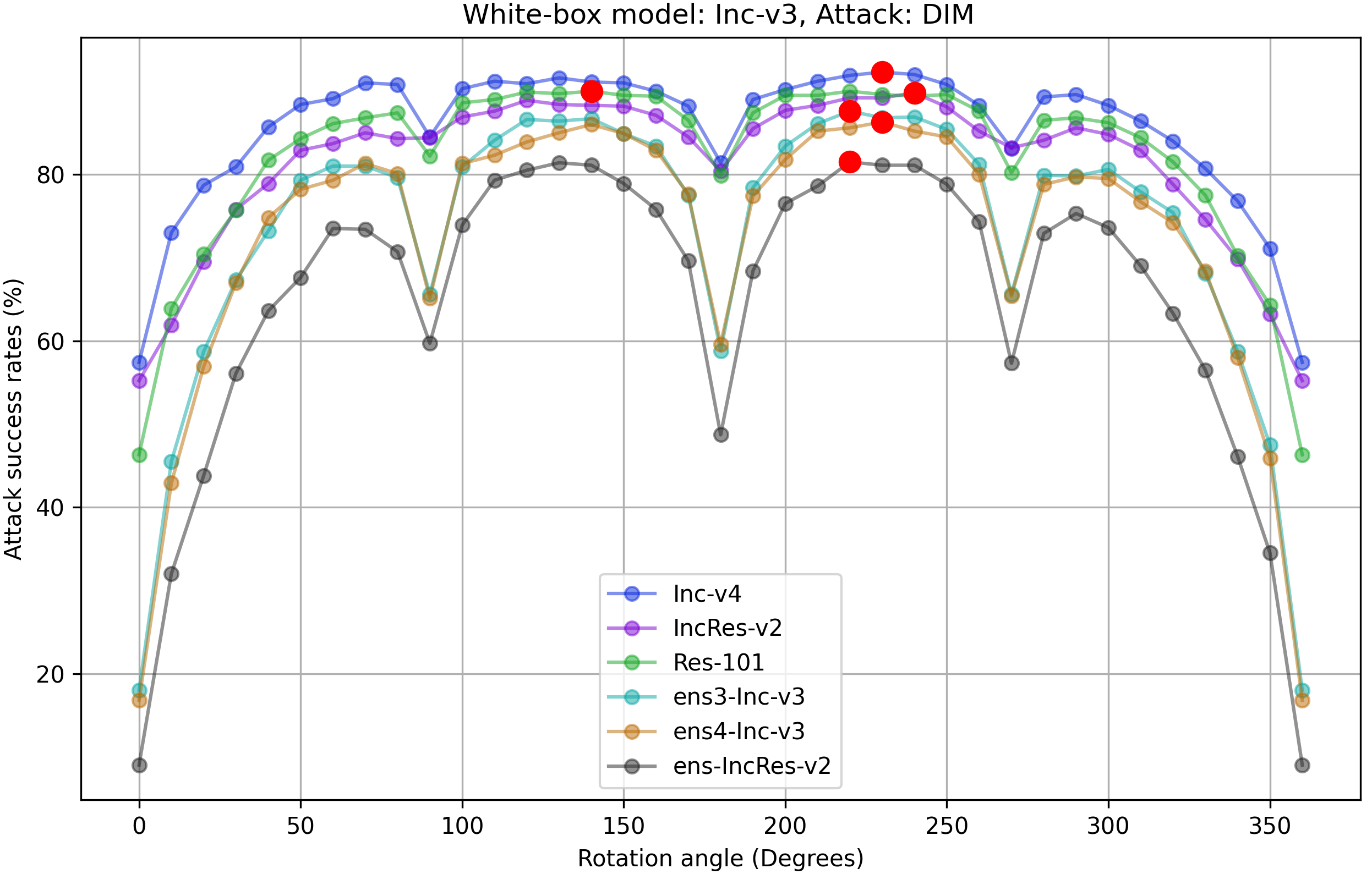}
	\caption{Identify the optimal rotation angle for maximizing transferability across six black-box models (NIPS'17). This figure is generated under the single-model setting with the Inc-v3 as the white-box model and DIM as the baseline attack. The x-axis represents the counter-clockwise rotation angle from $0^\circ$ to $360^\circ$, and the y-axis indicates the transferable attack success rates (\%). The six curves depict the fluctuations in attack success rates for each black-box model as the rotation angle varies.}
	\label{fig:optimal_angle:NIPS17}
\end{figure}

\begin{figure}[t]
	\centering
	\begin{tabular}{@{}c@{}}
		\includegraphics[width=0.9\linewidth]{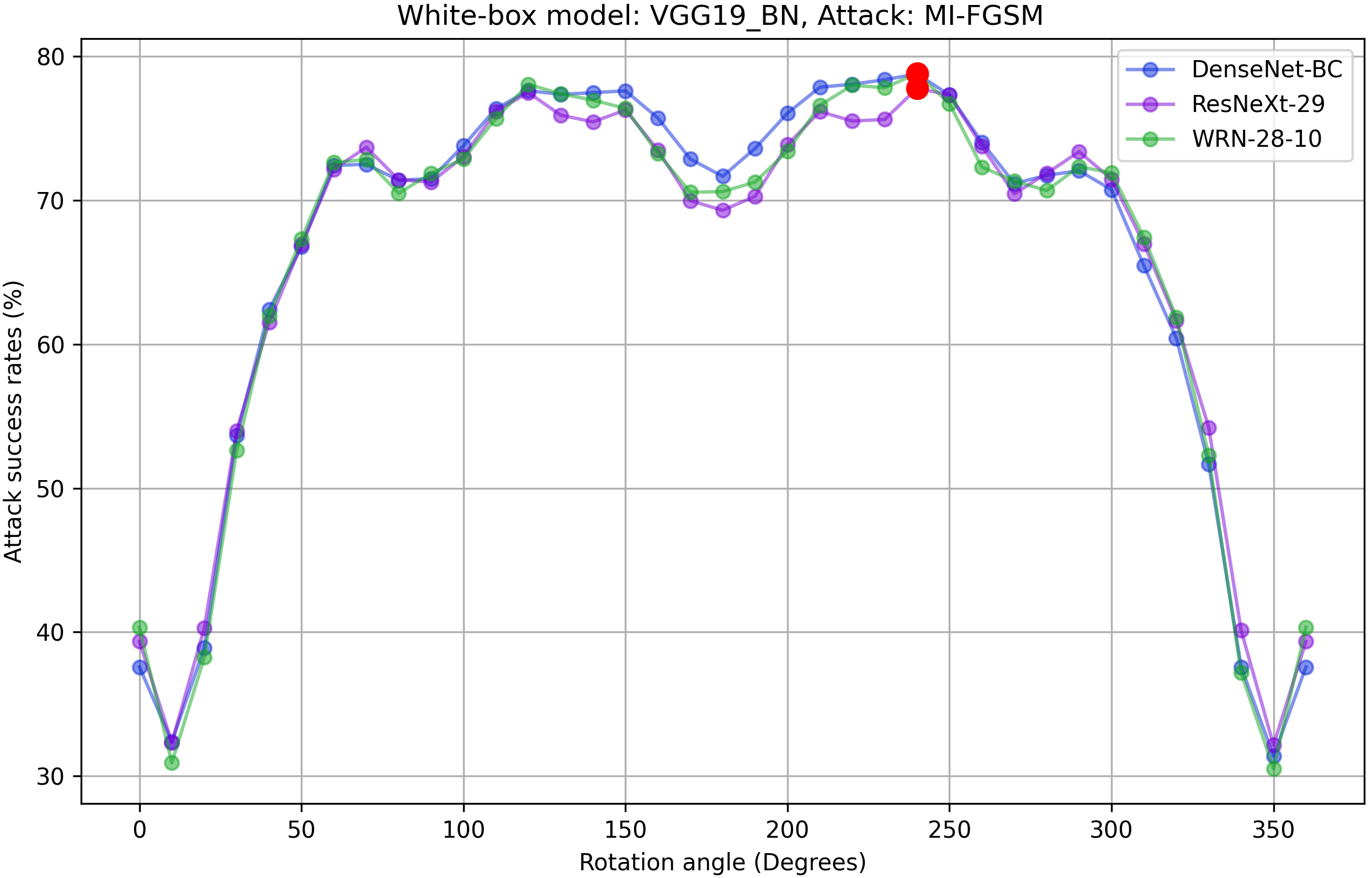} \\
		\includegraphics[width=0.9\linewidth]{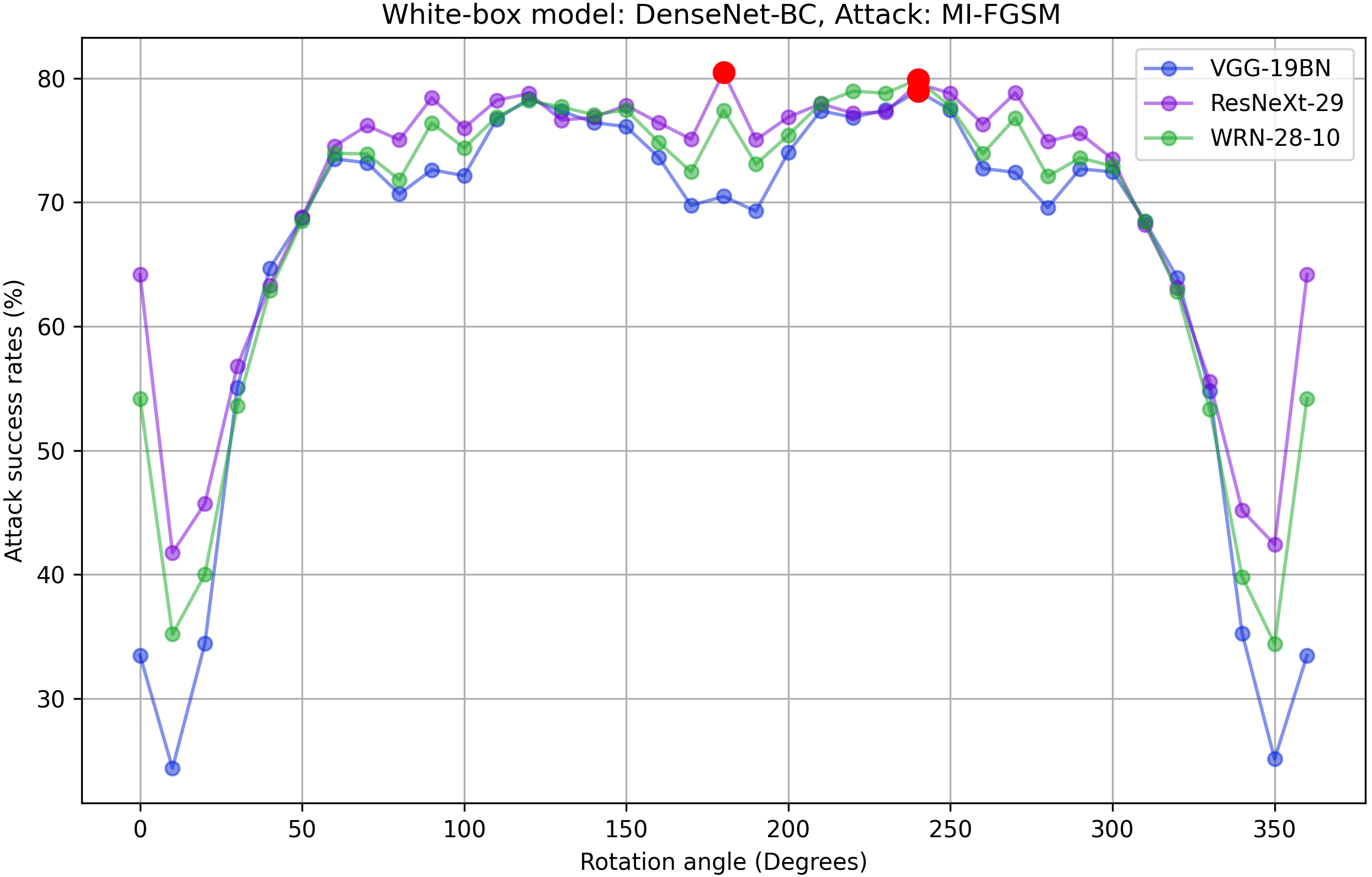}
	\end{tabular}
	\caption{Identify the optimal rotation angle for maximizing transferability across three black-box models (CIFAR-10). The white-box model is the VGG-19BN (upper) or DenseNet-BC (lower), respectively. Please refer to other details in Figure \ref{fig:optimal_angle:NIPS17}.}
	\label{fig:optimal_angle:CIFAR10}
\end{figure}

Figures \ref{fig:optimal_angle:NIPS17} and \ref{fig:optimal_angle:CIFAR10} illustrate three representative subplots from these experiments. Figure \ref{fig:optimal_angle:NIPS17}, using Inc-v3 as the white-box model and DIM as the baseline attack on NIPS’17, displays rotation angles from $0^\circ$ to $360^\circ$ on the x-axis and attack success rates on the y-axis. The results indicate that the six curves from different black-box models follow similar overall patterns, with local maxima and minima aligning across rotation angles, suggesting consistent underlying trends. However, variations in peak values indicate distinct optimal rotation angles across models for maximum transferability.

Figure \ref{fig:optimal_angle:CIFAR10} presents two additional subplots, using VGG-19BN and DenseNet-BC as white-box models, with MI-FGSM as the attack method on CIFAR-10. Here, the curves also exhibit a plateau between $70\%$ and $80\%$ success rates but lack the alignment of local maxima and minima seen in Figure \ref{fig:optimal_angle:NIPS17}, suggesting that the response to rotation angles varies across models in this dataset.

Across all subplots, we observe that rotation angles between $210^\circ$ and $240^\circ$ tend to yield higher attack success rates, supporting our assumption that the transpose operation may not yet be fully optimized. Adjusting AEs within this rotation range could potentially improve transferability beyond current levels.

% Exp Optimal angle figs

\section{Conclusion}
\label{sec:conclusion}
This paper presents a novel perspective on transferable attacks by introducing an input transpose method. This approach significantly enhances the cross-model generalization of AEs simply through transposition, with minimal additional labor or computational cost. Our further exploration reveals that, on specific datasets, even a minor input rotation--such as a $1^\circ$ shift left or right--can improve the transferability of many existing mainstream adversarial attacks. Our further analysis reveals that such a subtle adjustment could trigger non-negligible pattern fluctuations even at the DNN’s low-level layers. Additionally, our investigation finds that the transferability of adversarial examples varies with rotation angles, with an optimal range between $210^\circ$ and $240^\circ$ for maximizing cross-model success rates, provided unlimited black-box model queries are available. We believe this research contributes valuable insights into optimizing transferable attacks while keeping algorithmic complexity and computational costs low.

\newpage
{	
    \small
    \bibliographystyle{ieeenat_fullname}
    \bibliography{main}
}

\clearpage
\setcounter{page}{1}
\maketitlesupplementary

\section{Why does a $1^\circ$ rotation improve the transferability? (More evaluations)}
\label{supp:sec:why_1_rotation_work}
\begin{figure*}[!h]
	\centering
	\begin{tabular}{@{}c@{}c@{}c@{}c@{}c@{}}
		\includegraphics[height=0.19\textwidth]{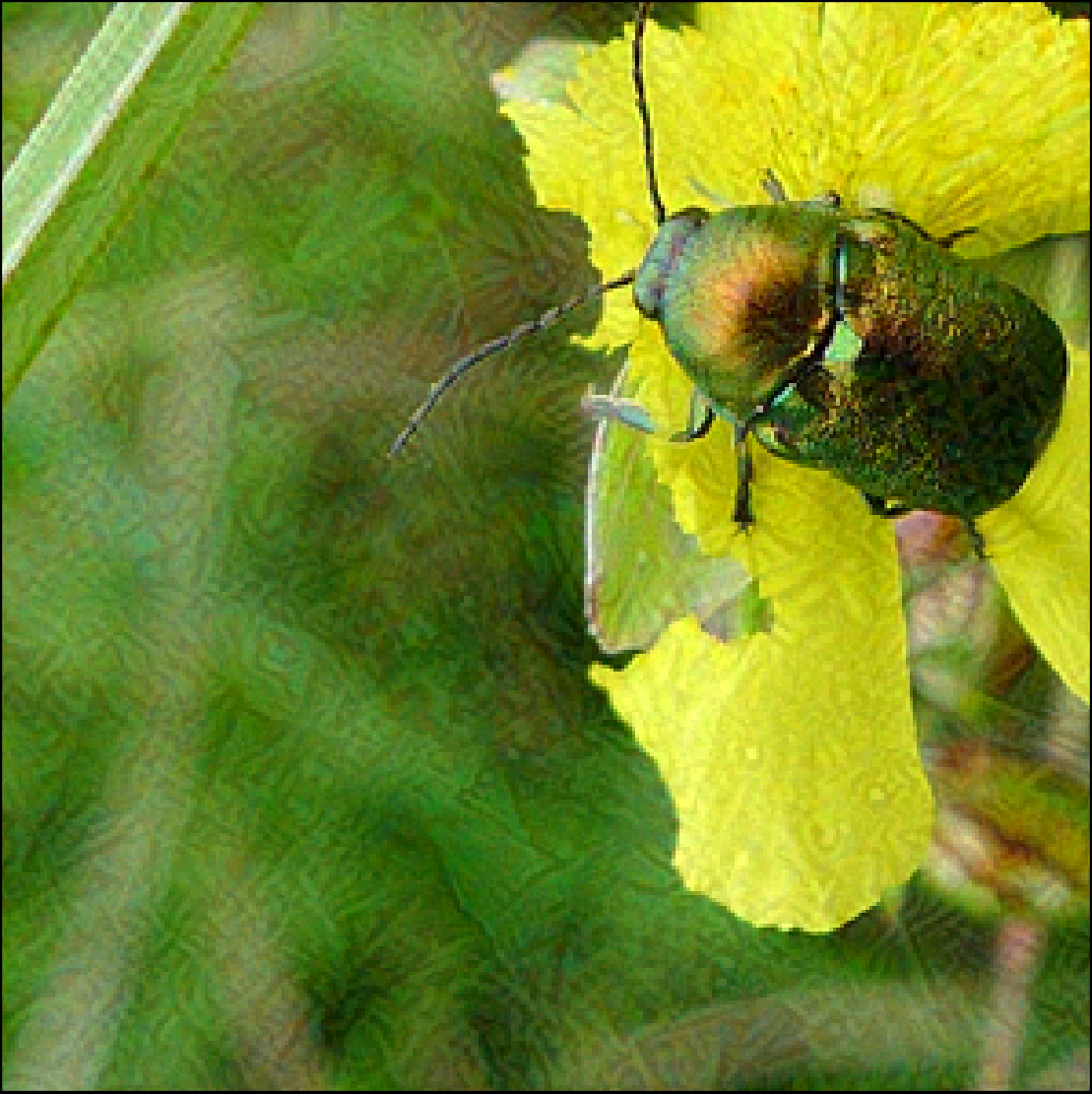} \quad & 
		\includegraphics[height=0.19\textwidth]{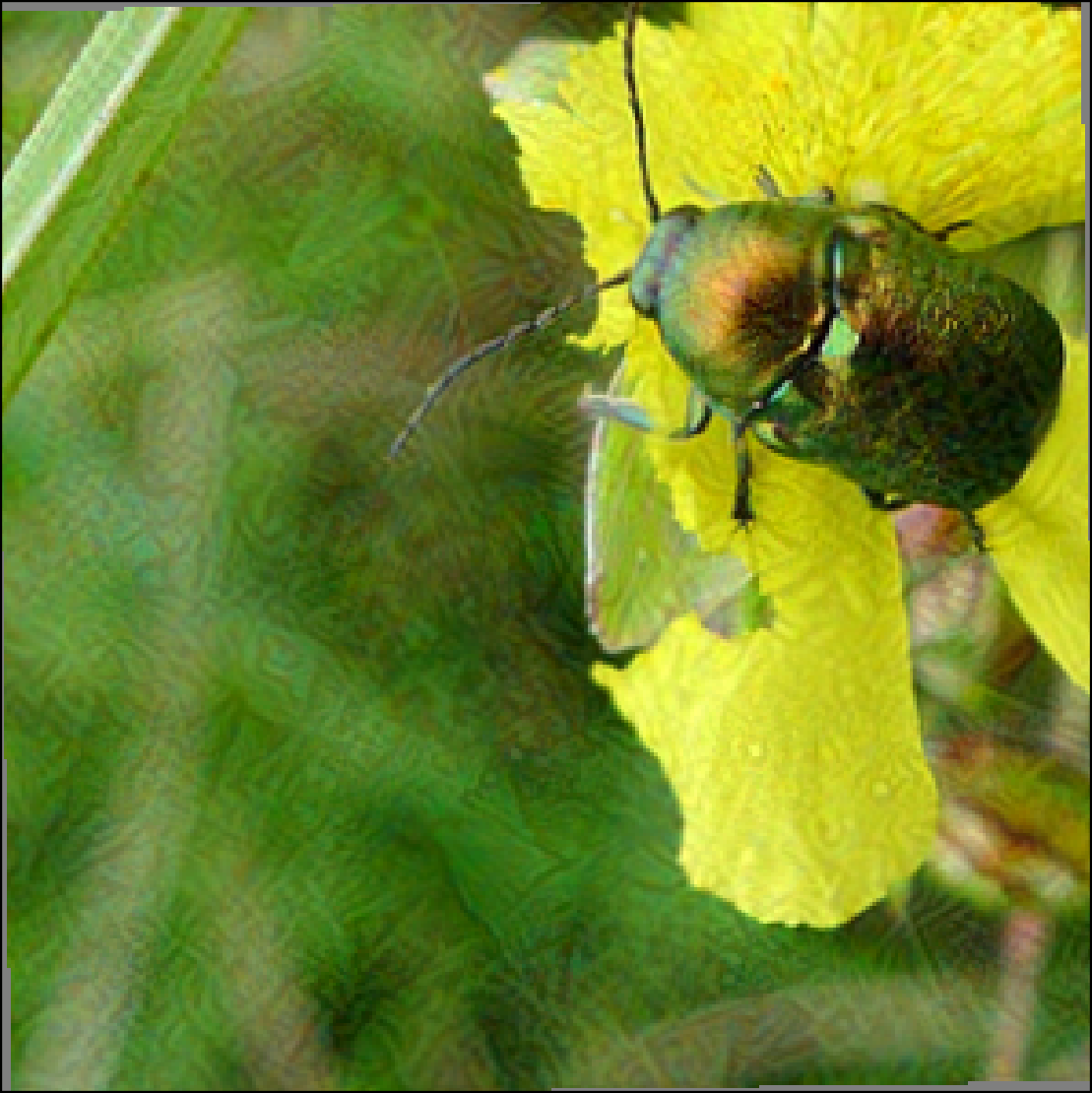} \quad &
		\includegraphics[height=0.19\textwidth]{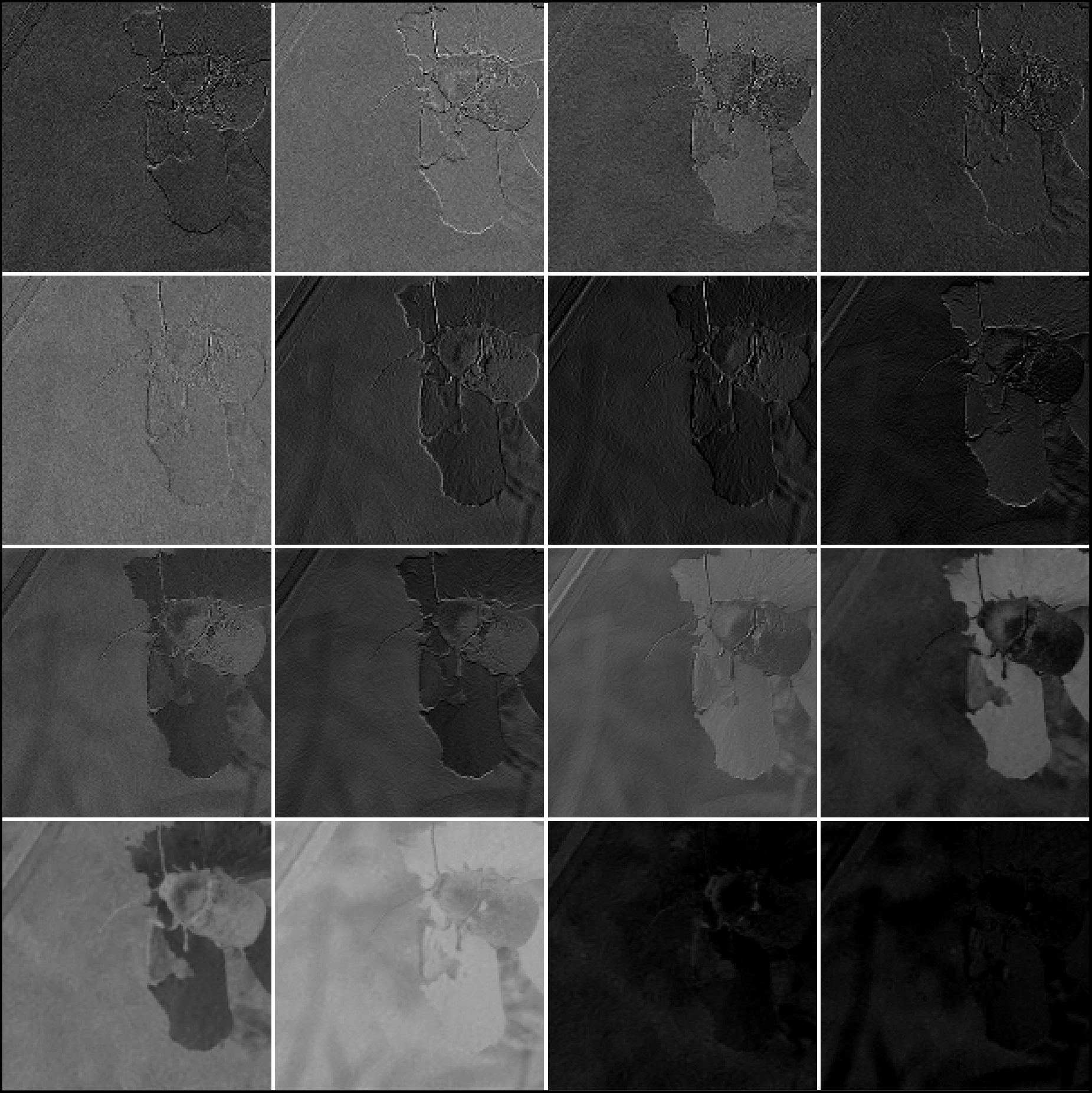} \quad &
		\includegraphics[height=0.19\textwidth]{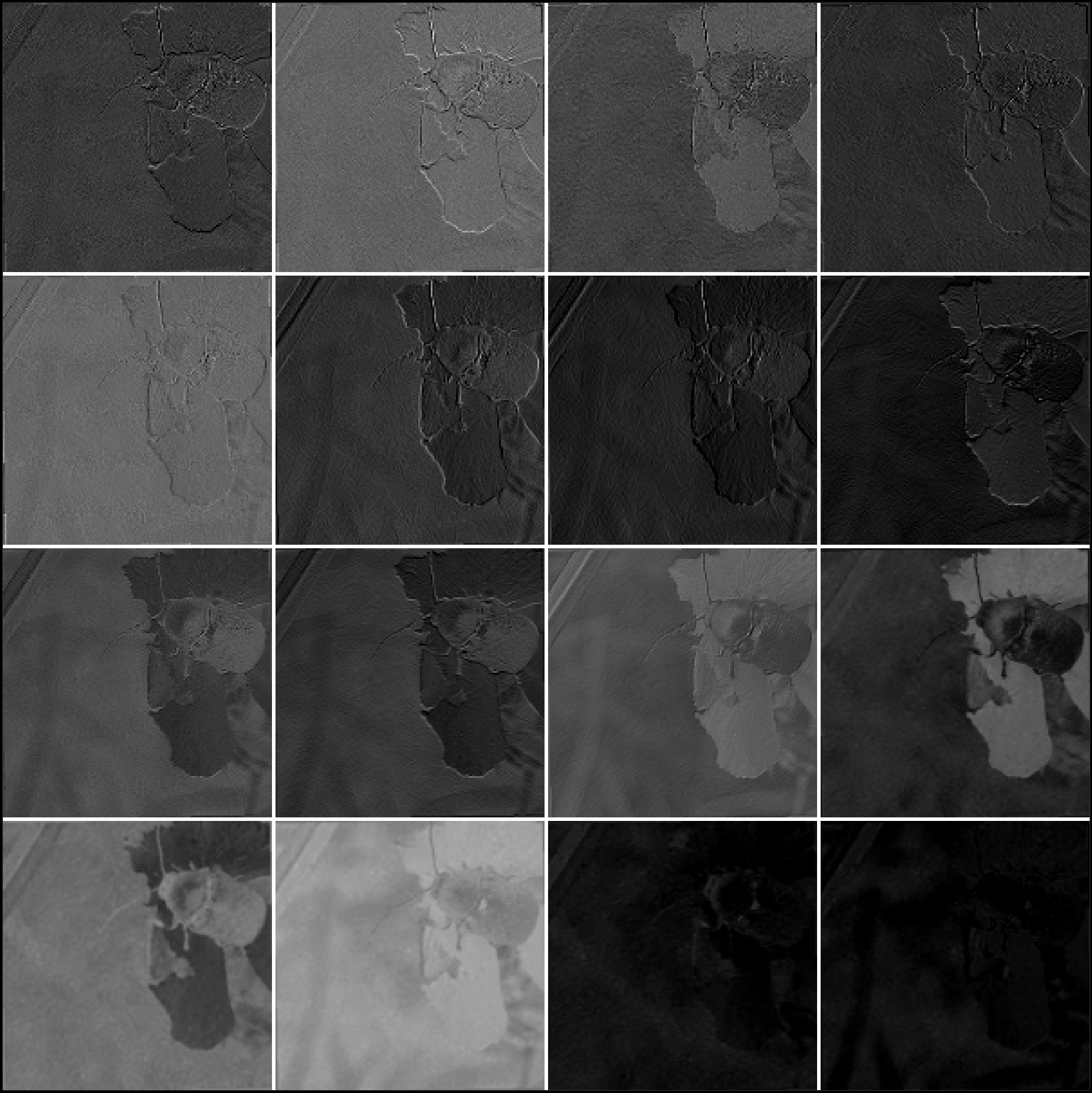} \quad &
		\includegraphics[height=0.19\textwidth]{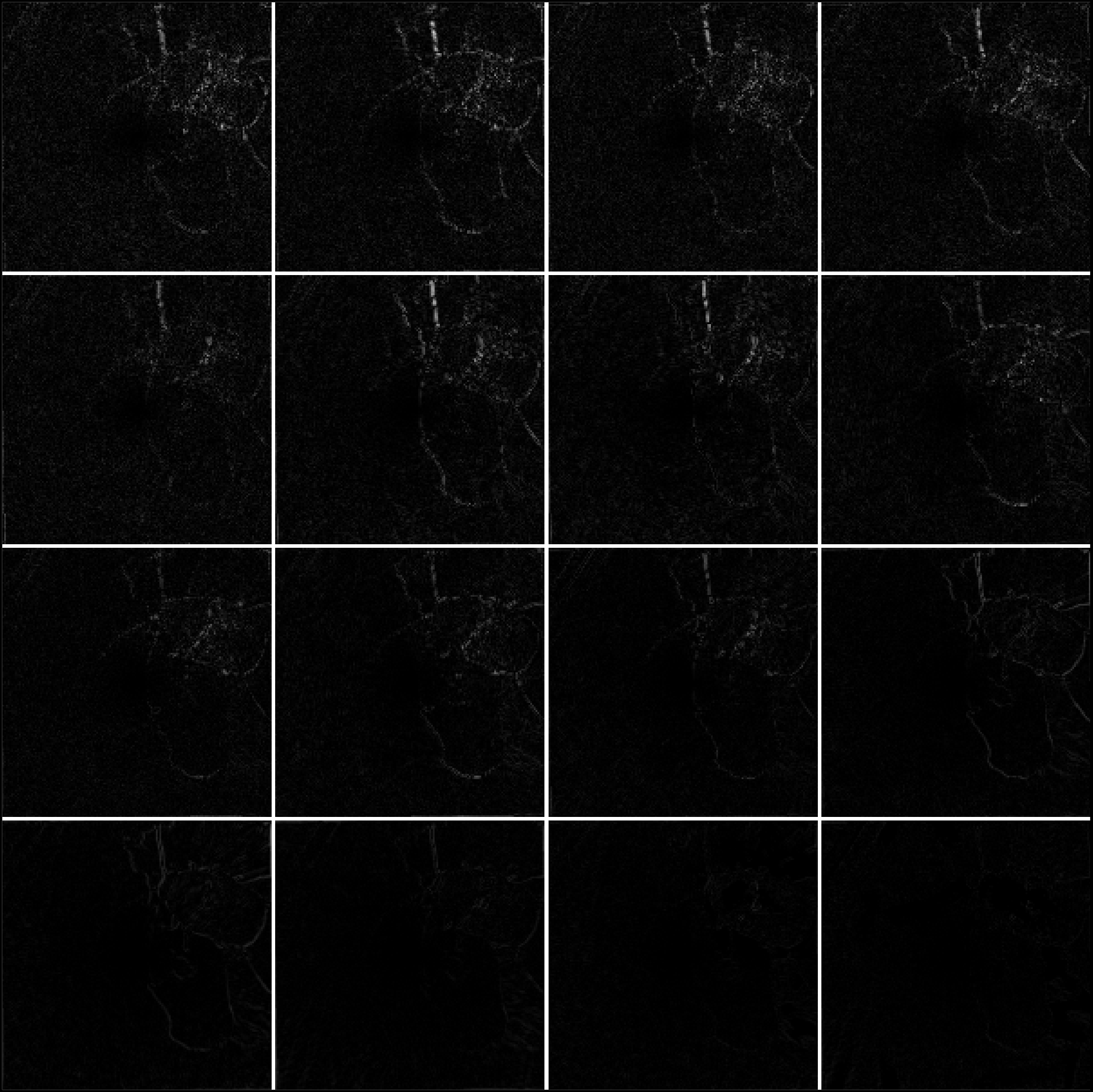} \\
		$(1)$ Advr & $(2)$ $1^\circ$ Left & $(3)$ FMs wo $1^\circ$ & $(4)$ FMs w $1^\circ$ & $(5)$ Diffs between $(3)$ \& $(4)$ \\
	\end{tabular}
	\caption{Identify major feature fluctuations that potentially influence the $\text{Inc-v3}_{ens4}$'s decision under the single model setting. The white-box model and the baseline are consistent with those in Figure \ref{fig:featureanalysis}. We analyze the black-box model: $\text{Inc-v3}_{ens4}$. $(1)$ An AE that fails to deceive the $\text{Inc-v3}_{ens4}$. $(2)$ The $1^\circ$ left rotation of $(1)$ that successfully deceives the $\text{Inc-v3}_{ens4}$. $(3)$ Feature maps from $16$ selected channels in the Conv2d-1a layer in $\text{Inc-v3}_{ens4}$ with $(1)$ as the input. Similar as Figure \ref{fig:featureanalysis}, the $16$ channels are determined by $(5)$. $(4)$ Feature maps of the same $16$ channels with $(2)$ as the input; $(5)$ Absolute differences between $(3)$ and $(4)$ (Zoom in to see details).}
	\label{supp:fig:featureanalysis:dim:left_1:1}
\end{figure*}

\begin{figure*}[!h]
	\centering
	\begin{tabular}{@{}c@{}c@{}c@{}c@{}c@{}}
		\includegraphics[height=0.19\textwidth]{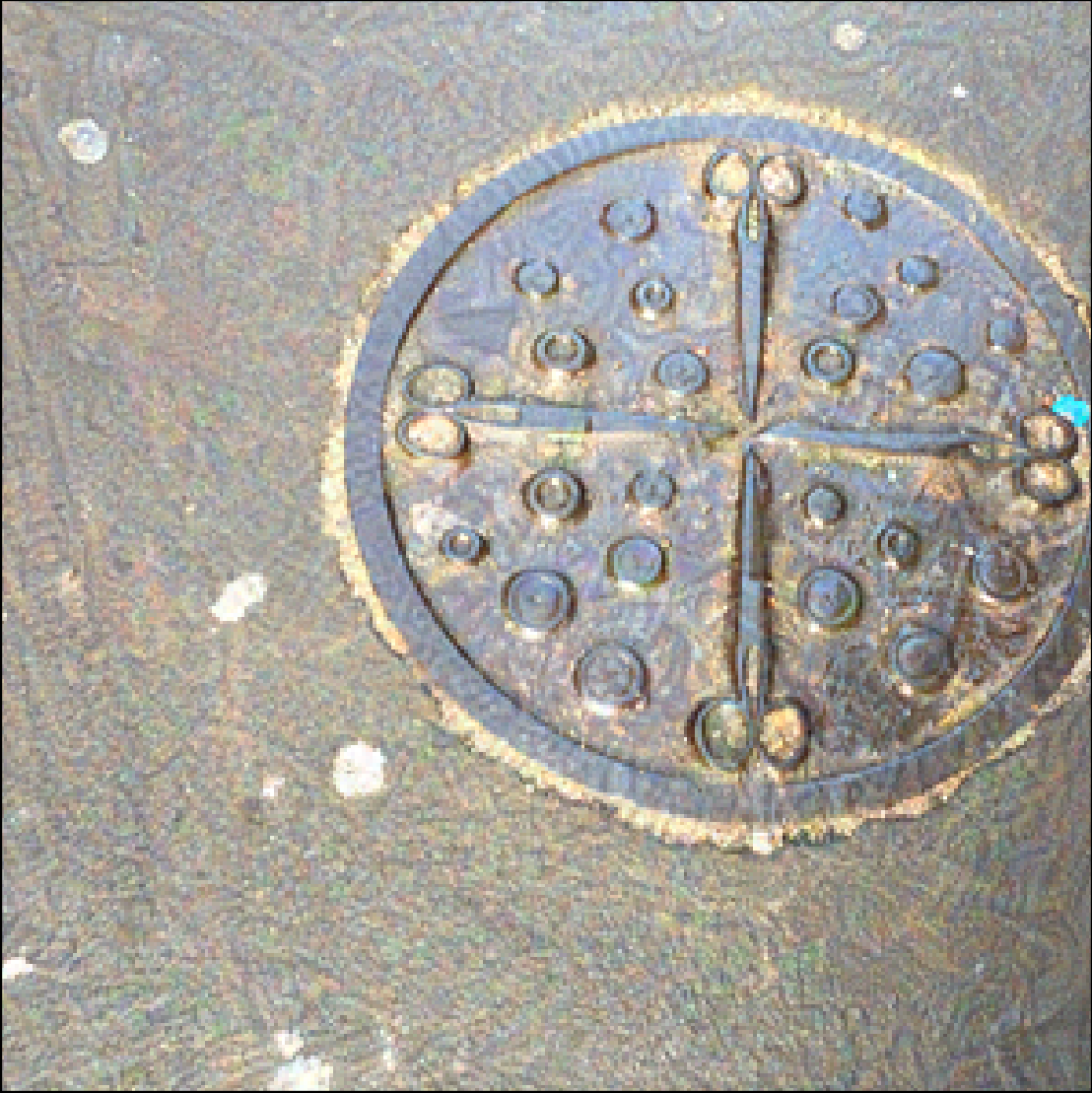} \quad & 
		\includegraphics[height=0.19\textwidth]{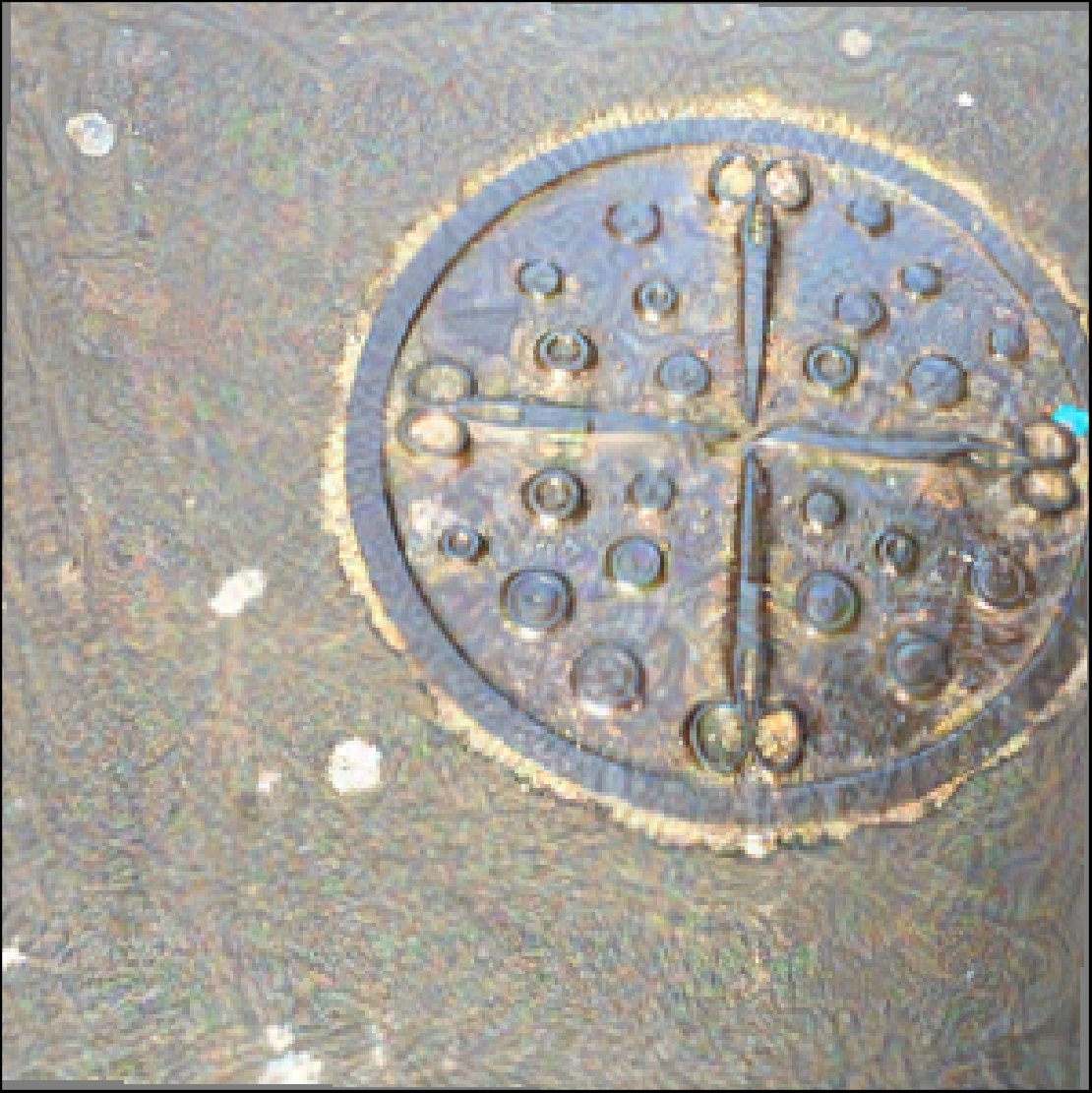} \quad &
		\includegraphics[height=0.19\textwidth]{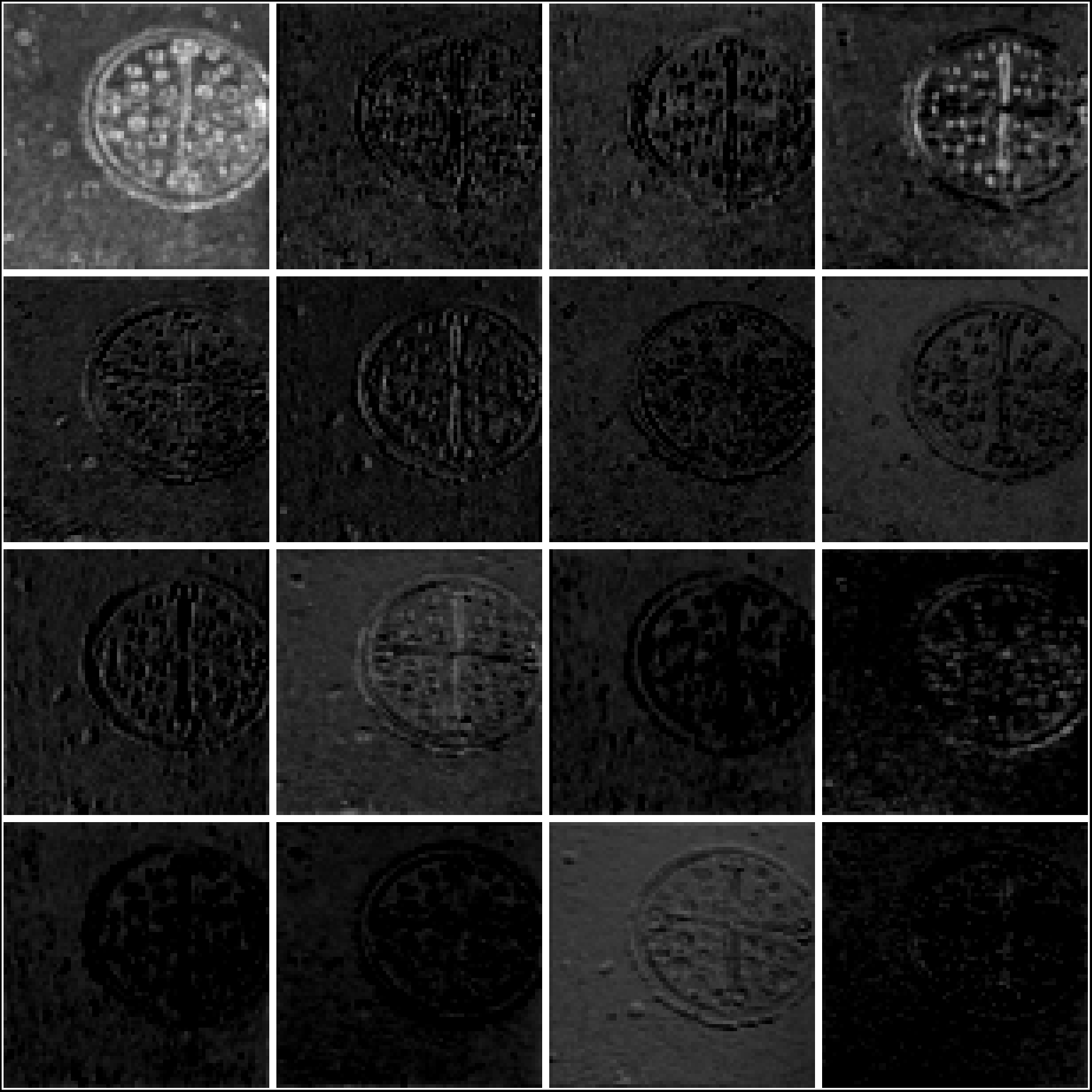} \quad &
		\includegraphics[height=0.19\textwidth]{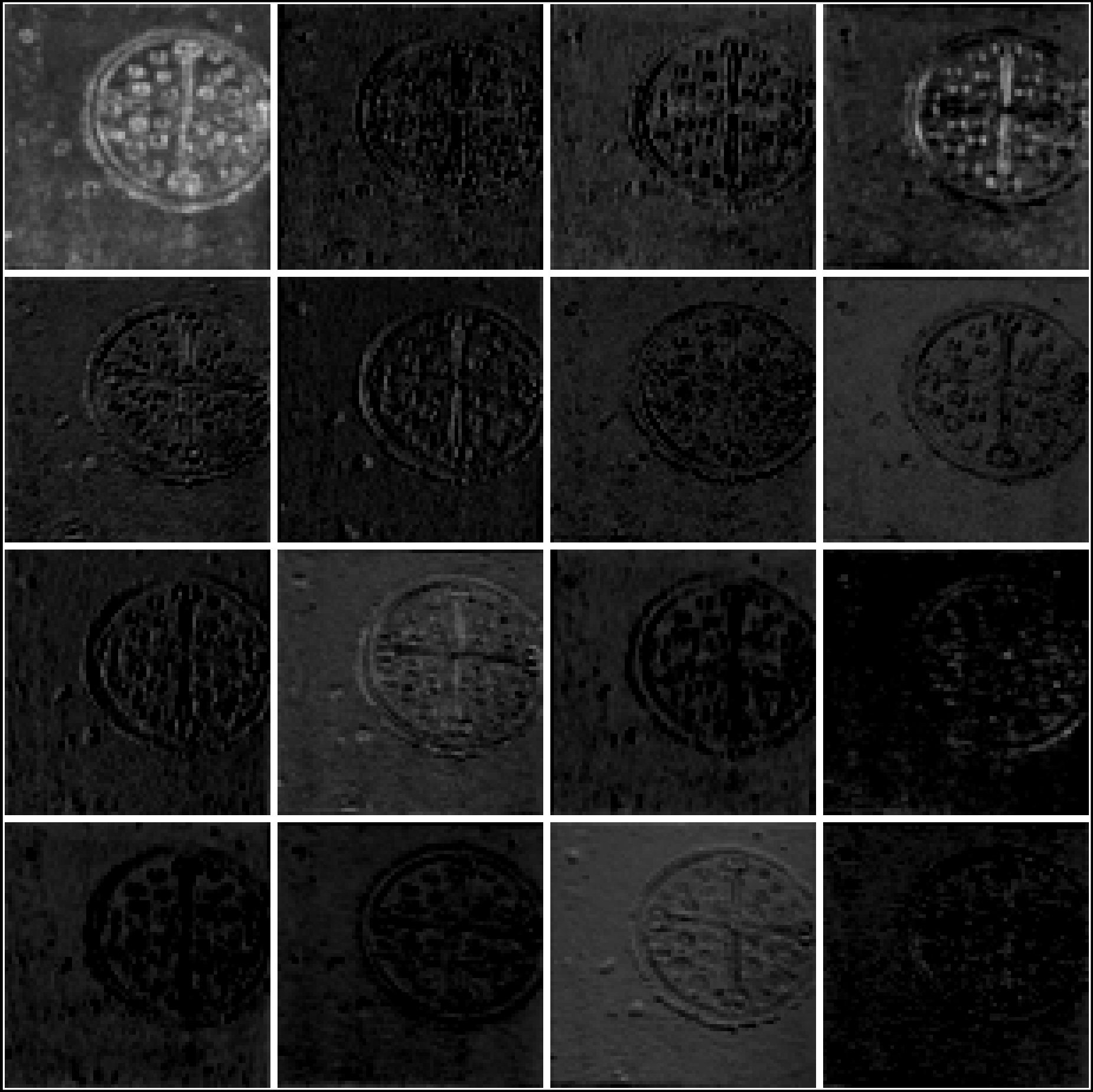} \quad &
		\includegraphics[height=0.19\textwidth]{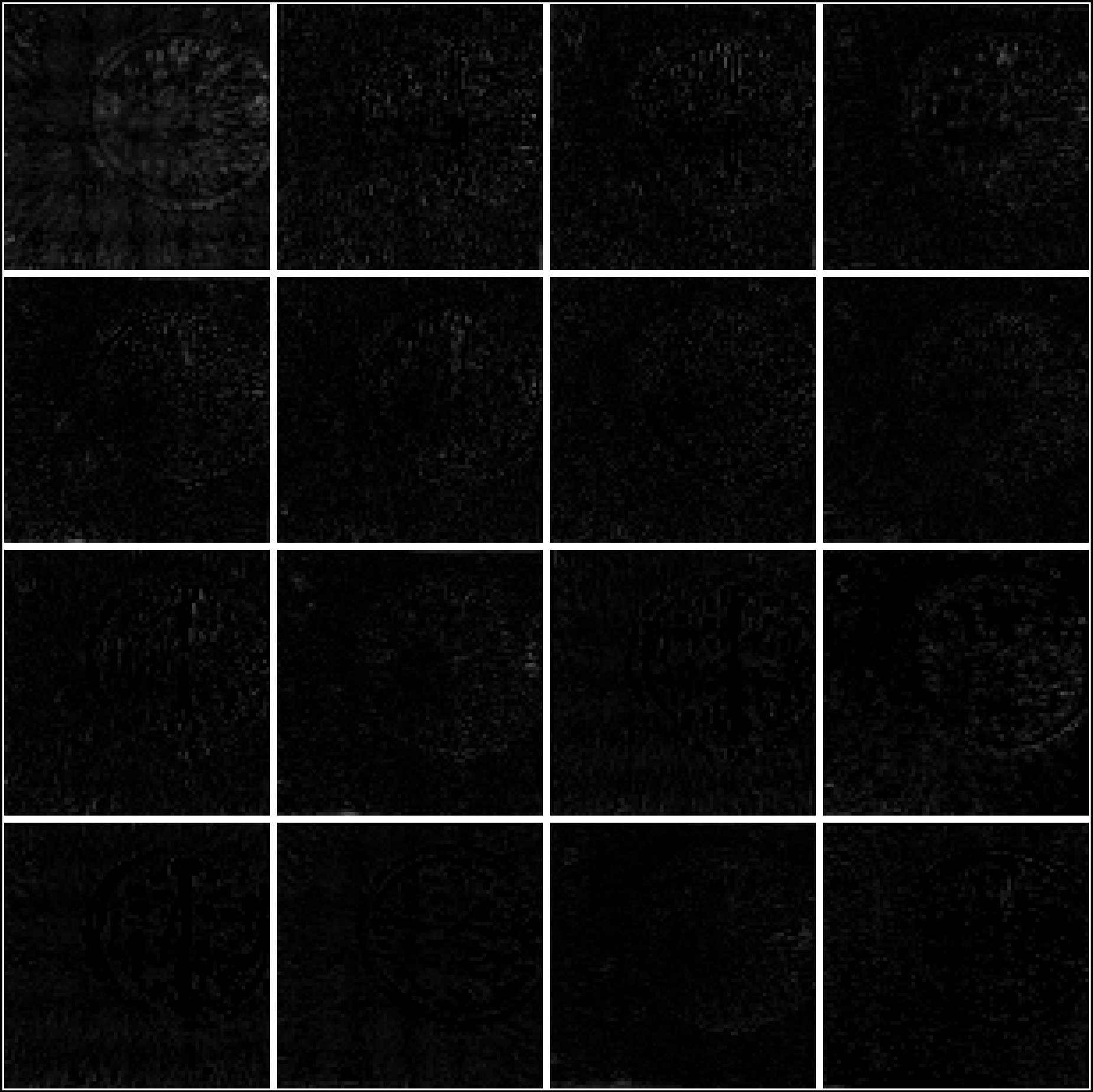} \\
		$(1)$ Advr & $(2)$ $1^\circ$ Right & $(3)$ FMs wo $1^\circ$ & $(4)$ FMs w $1^\circ$ & $(5)$ Diffs between $(3)$ \& $(4)$ \\
		\includegraphics[height=0.19\textwidth]{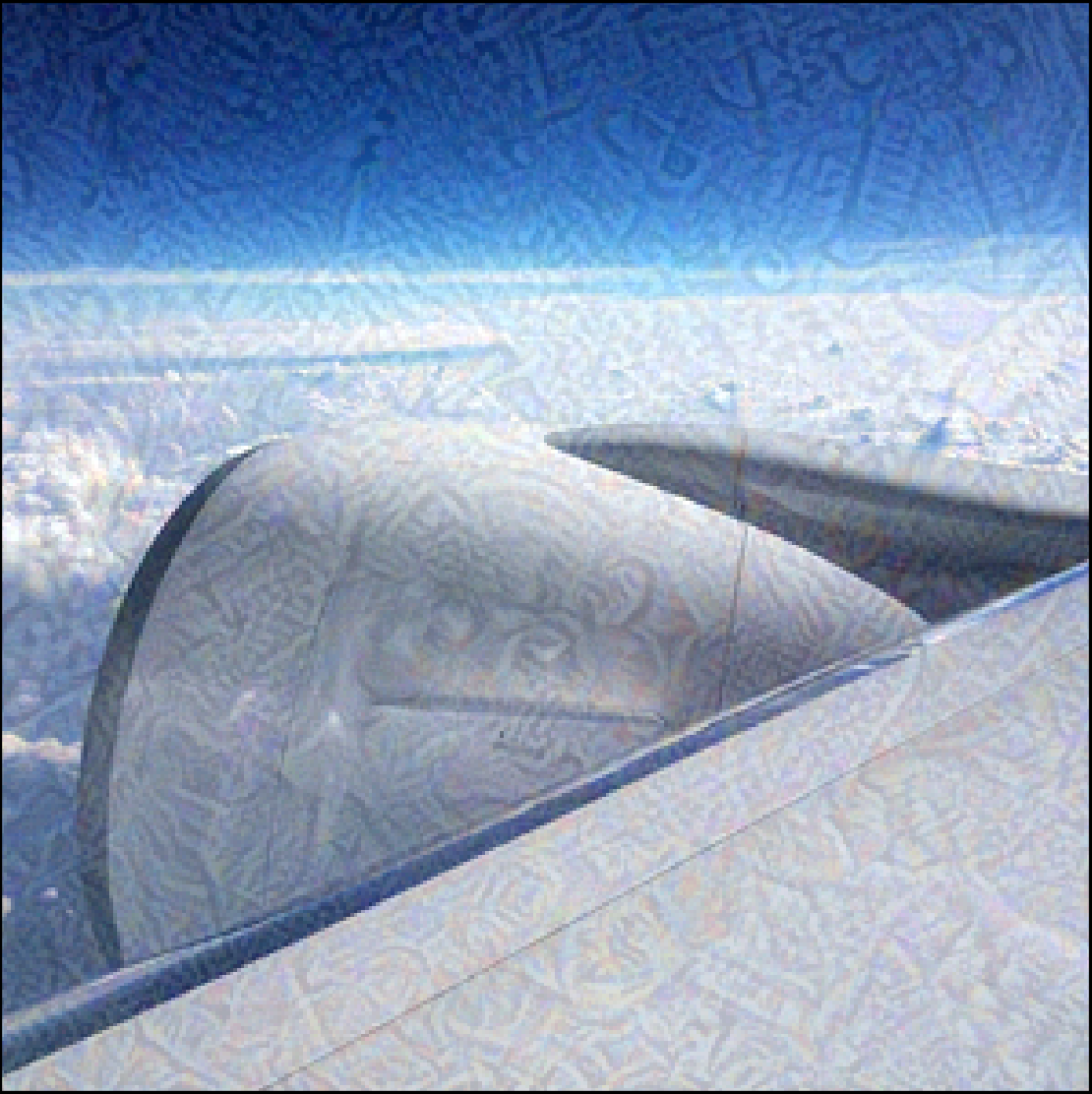} \quad & 
		\includegraphics[height=0.19\textwidth]{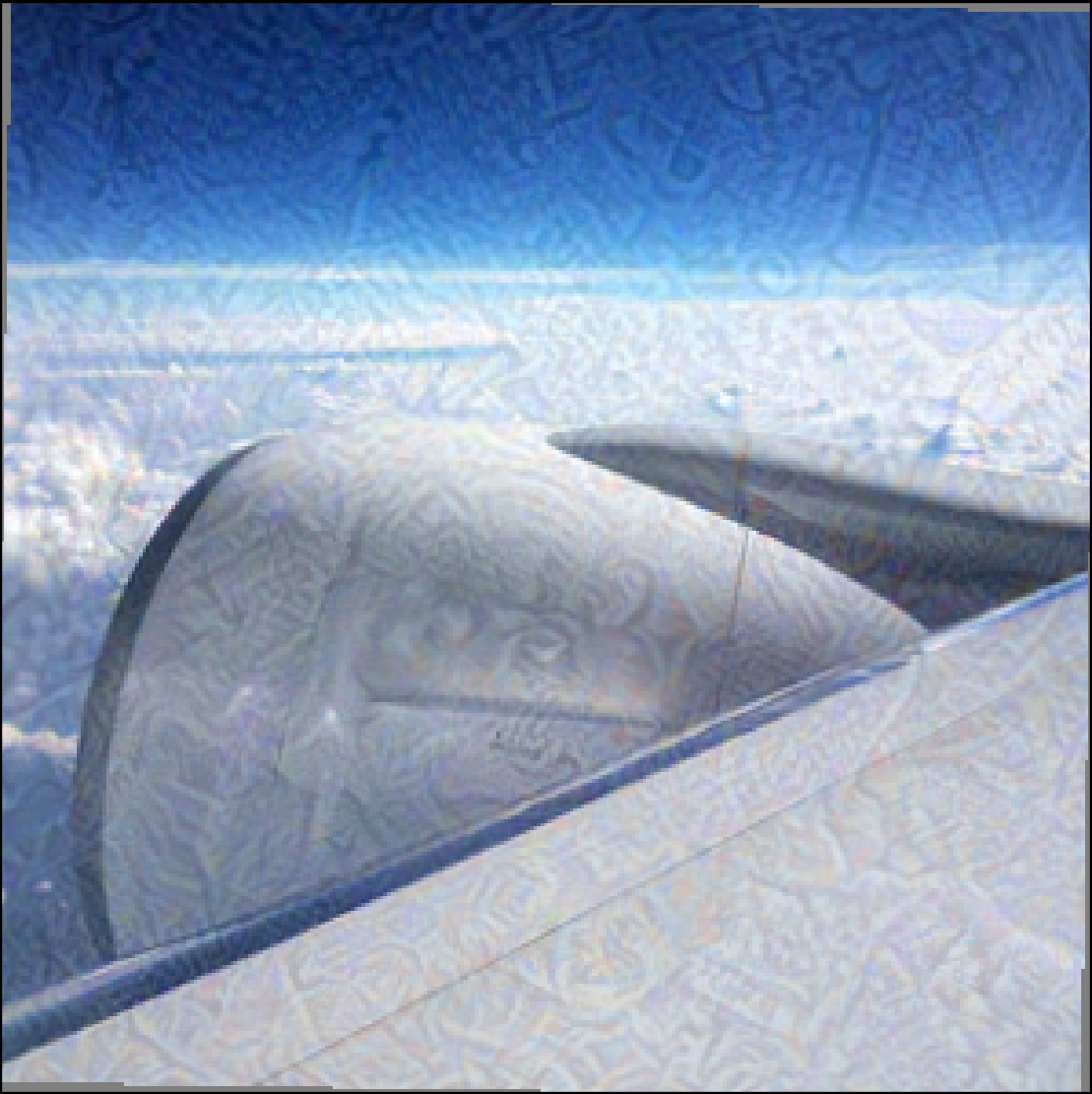} \quad &
		\includegraphics[height=0.19\textwidth]{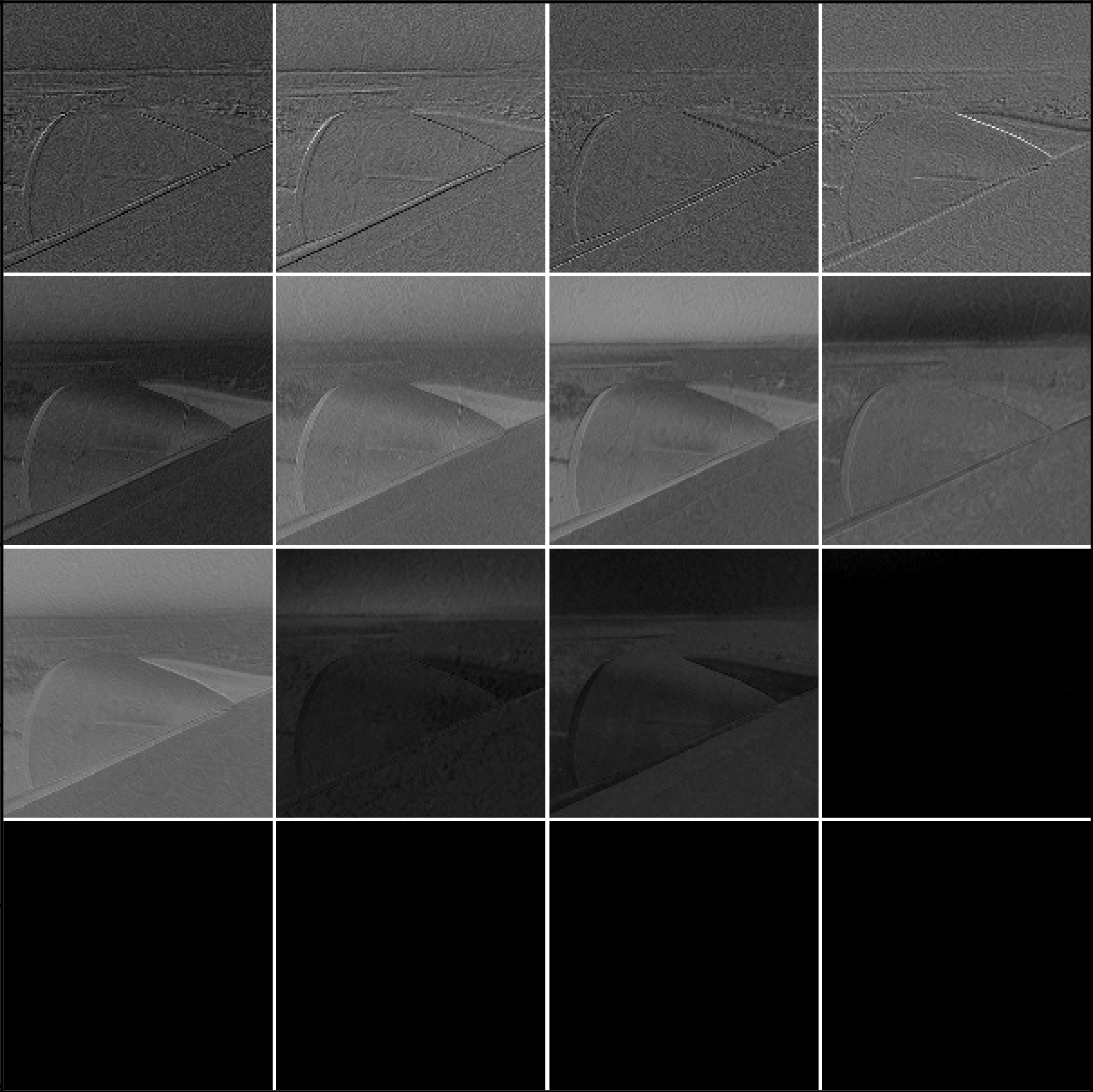} \quad &
		\includegraphics[height=0.19\textwidth]{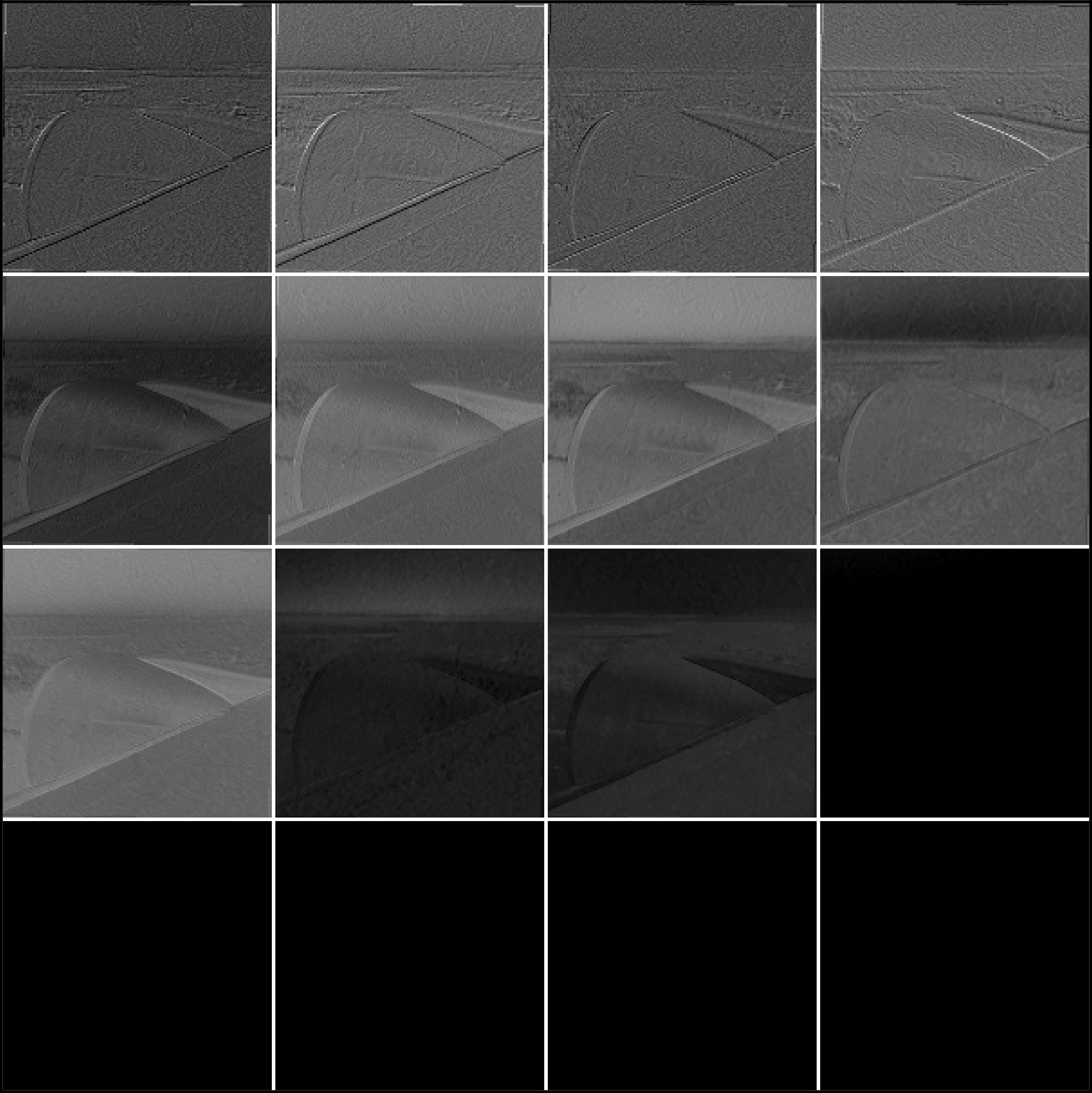} \quad &
		\includegraphics[height=0.19\textwidth]{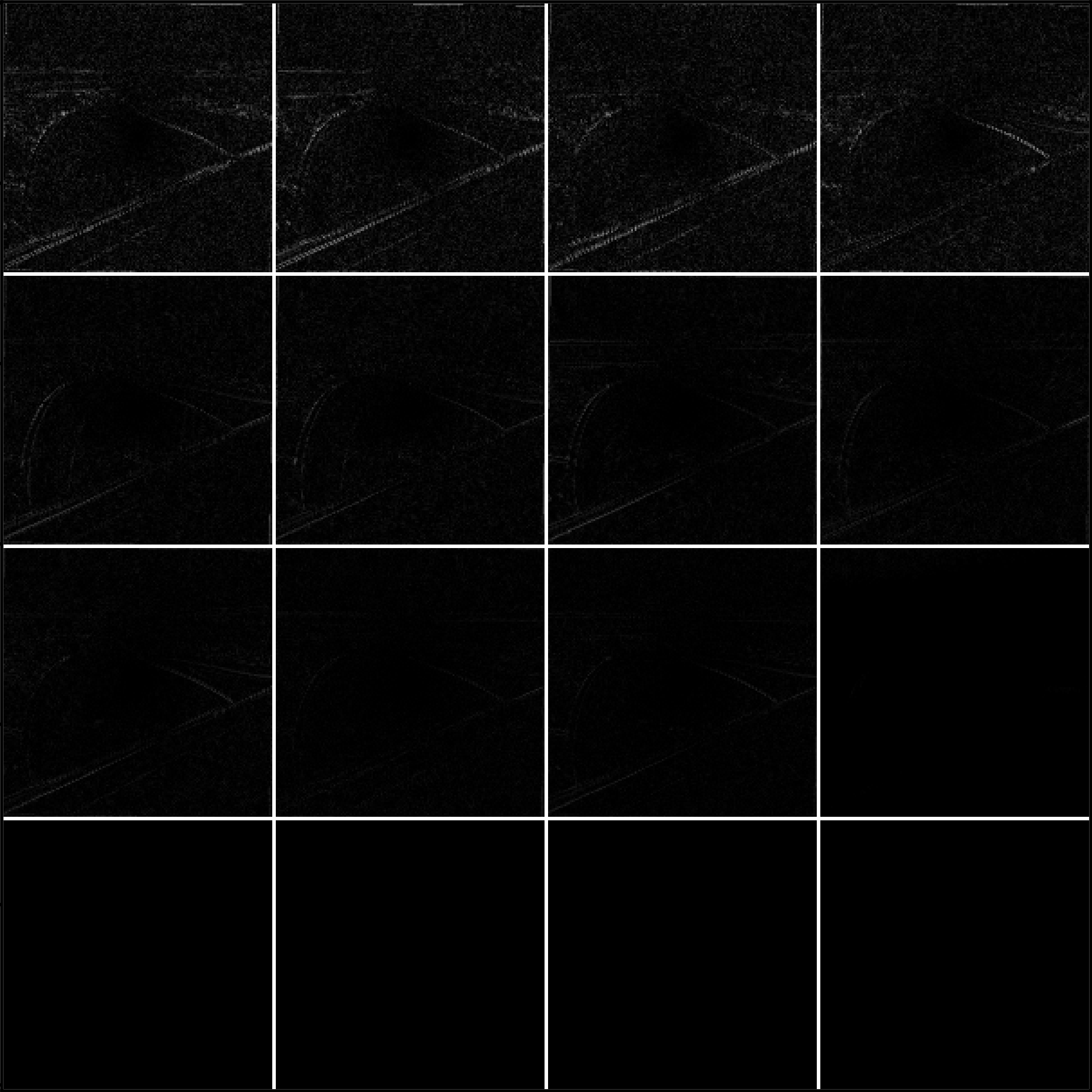} \\
	\end{tabular}
	\caption{Identify major feature fluctuations that potentially influence the decisions of Res-101 (upper row) and IncRes-v2$_{ens}$ (lower row) under the single model setting. The white-box model and the baseline are consistent with those in Figure \ref{fig:featureanalysis}. $(2)$ The $1^\circ$ right rotation of $(1)$ that successfully deceives the corresponding black-box model. $(3)$ The upper feature maps are selected from the Conv1 layer of the Block 1 in Res-101, while the lower feature maps are from the Conv2d-1a layer in IncRes-v2$_{ens}$. Additional details can be found in Figure \ref{supp:fig:featureanalysis:dim:left_1:1}.}
	\label{supp:fig:featureanalysis:dim:right_1}
\end{figure*}

\begin{figure*}[!h]
	\centering
	\begin{tabular}{@{}c@{}c@{}c@{}c@{}c@{}}
		\includegraphics[height=0.19\textwidth]{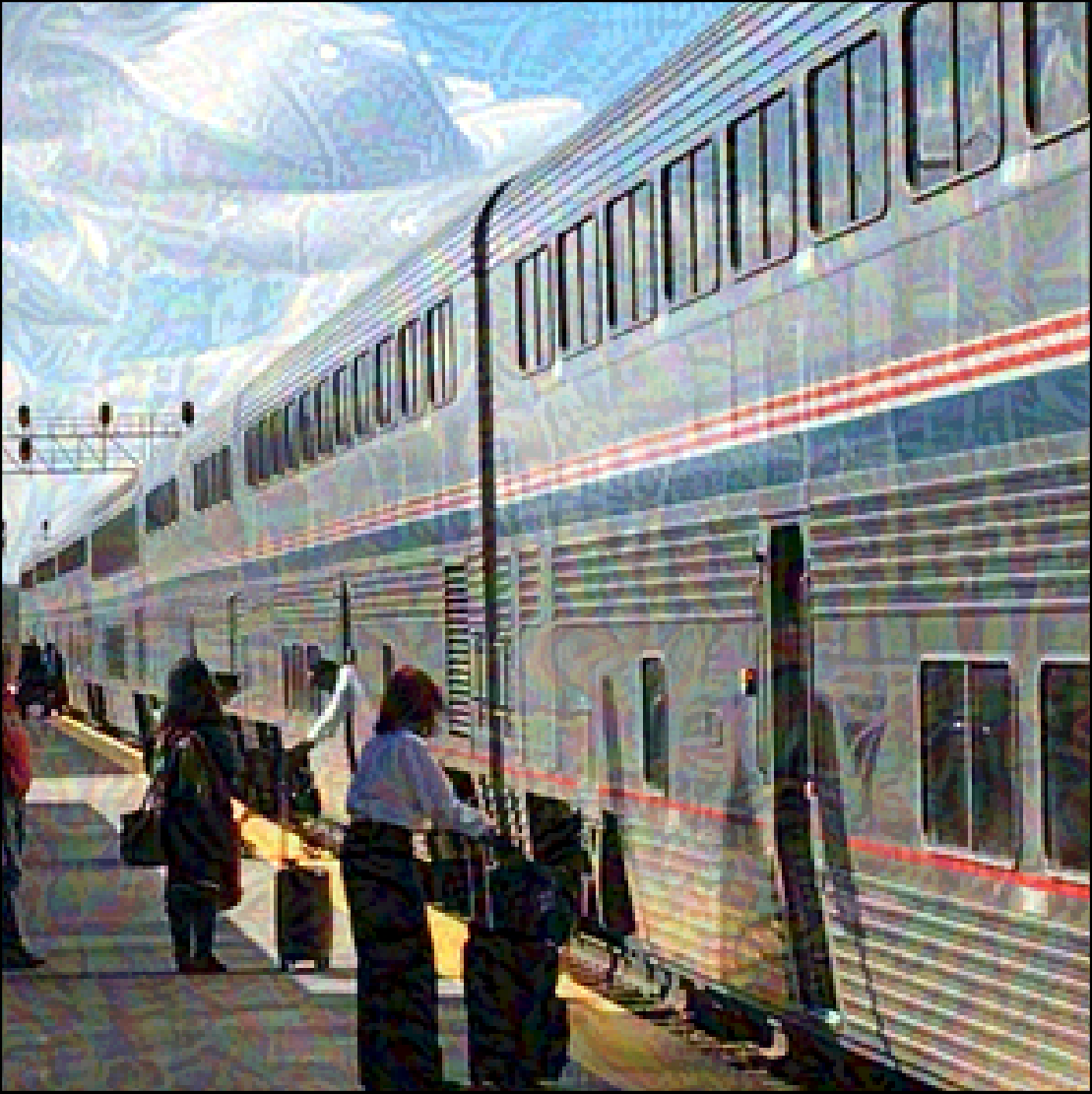} \quad & 
		\includegraphics[height=0.19\textwidth]{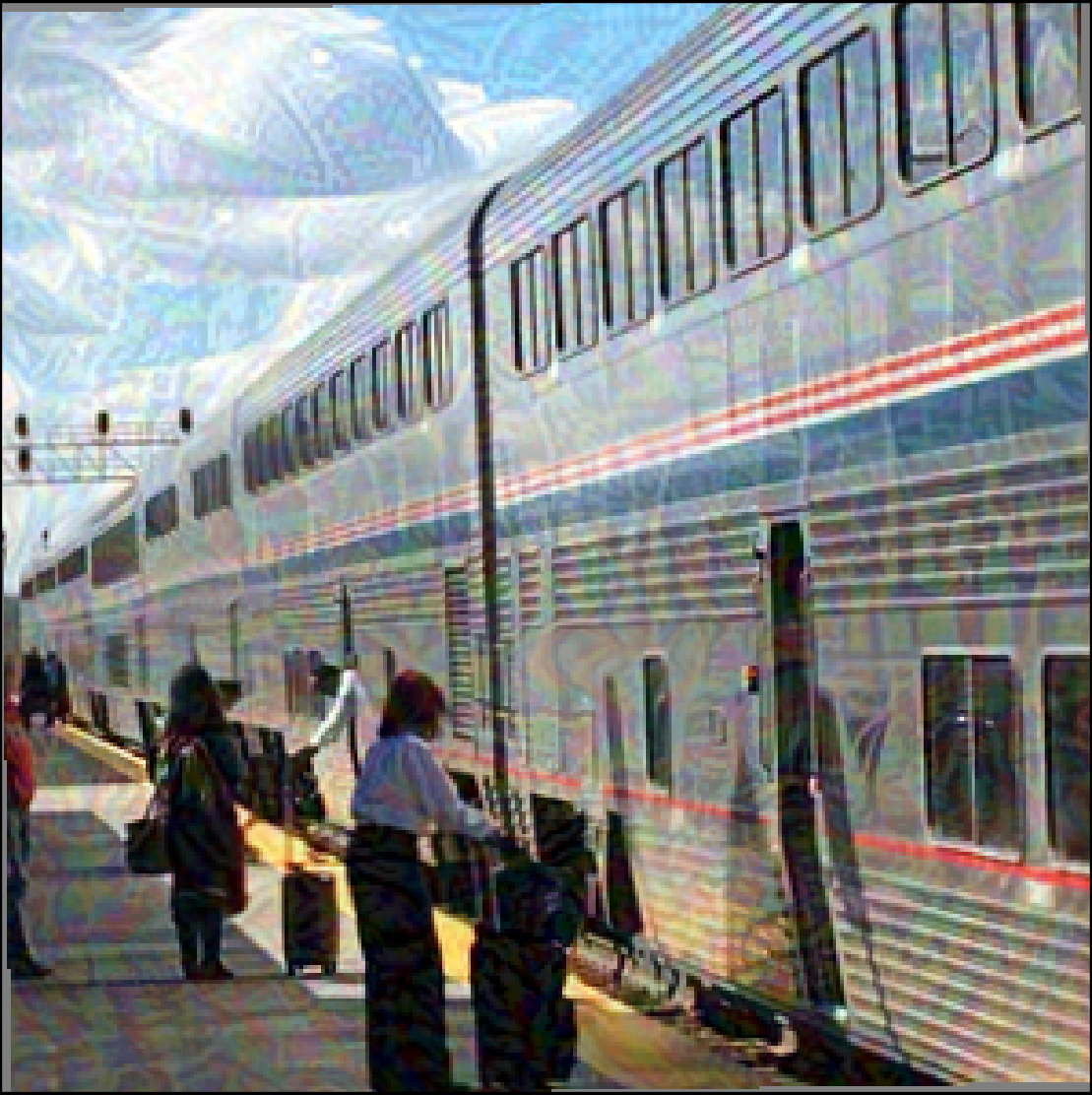} \quad &
		\includegraphics[height=0.19\textwidth]{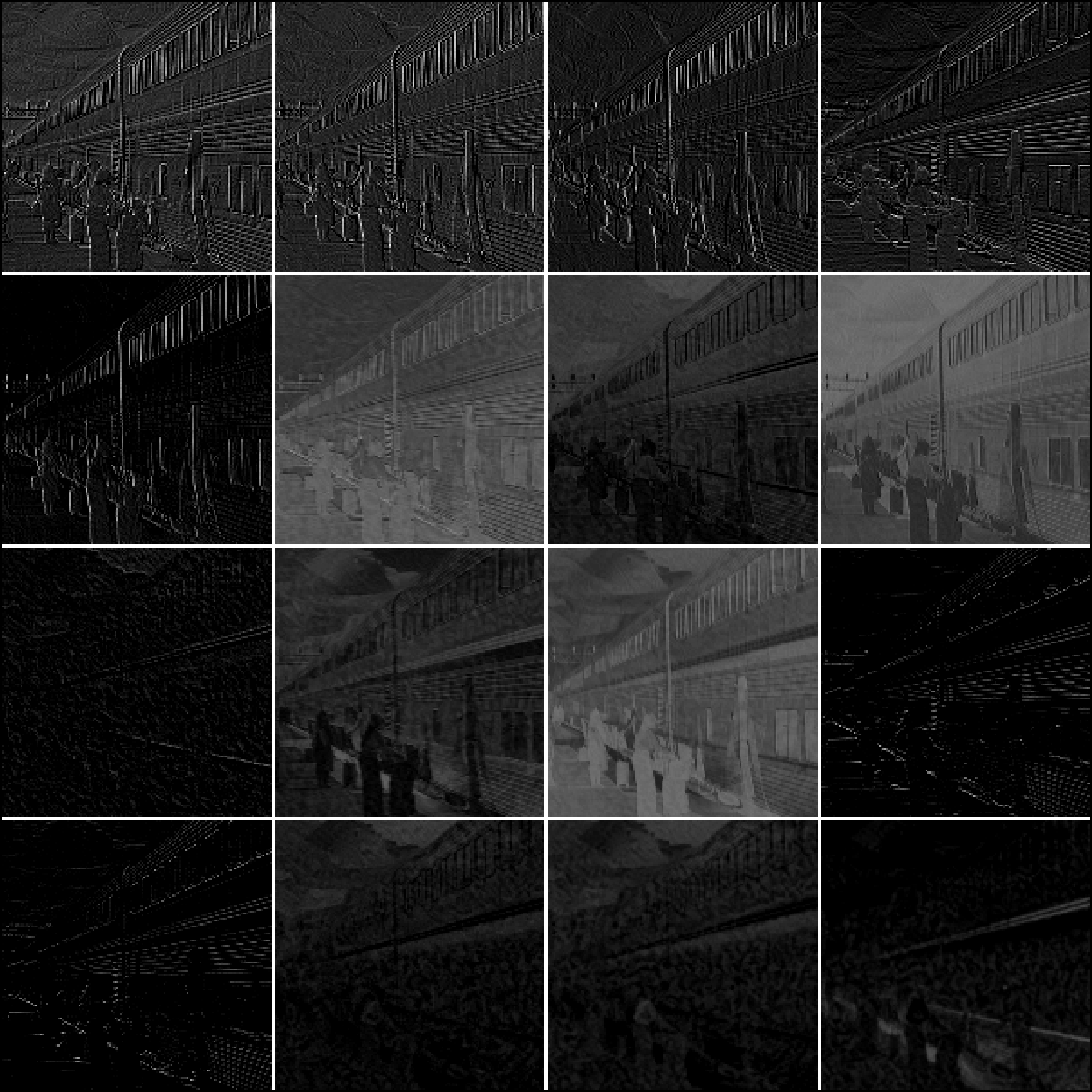} \quad &
		\includegraphics[height=0.19\textwidth]{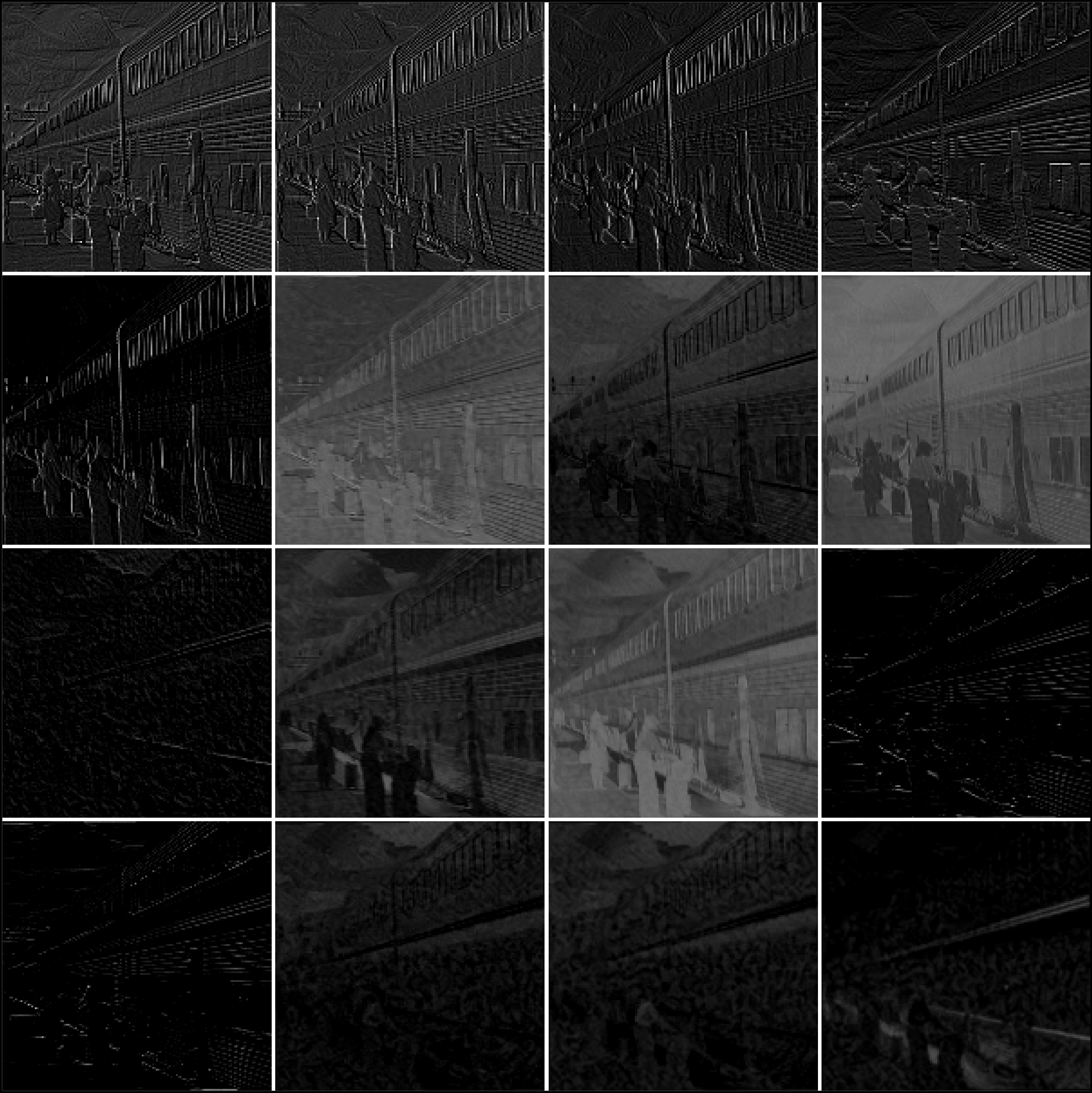} \quad &
		\includegraphics[height=0.19\textwidth]{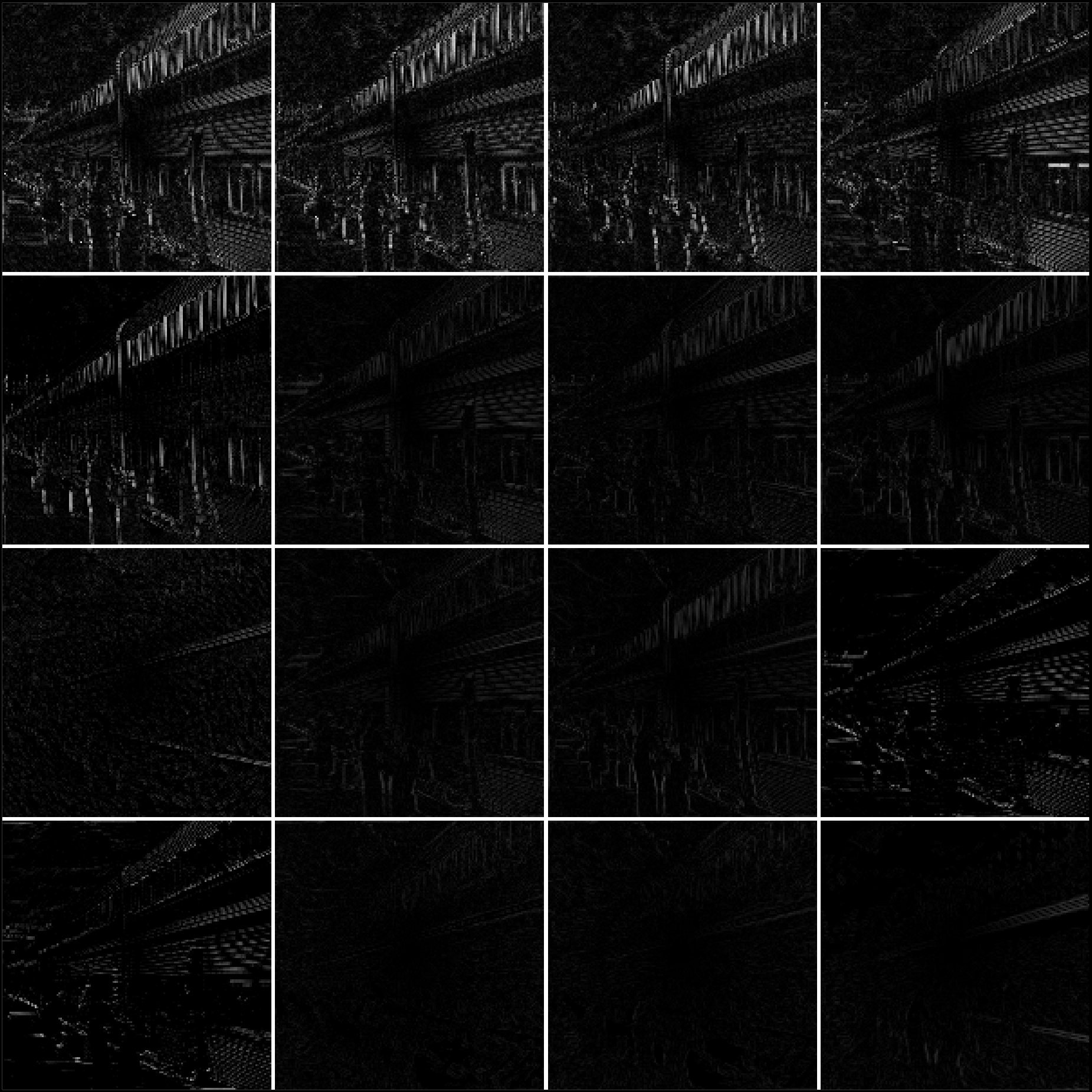} \\
		$(1)$ Advr & $(2)$ $1^\circ$ Left & $(3)$ FMs wo $1^\circ$ & $(4)$ FMs w $1^\circ$ & $(5)$ Diffs between $(3)$ \& $(4)$ \\
		\includegraphics[height=0.19\textwidth]{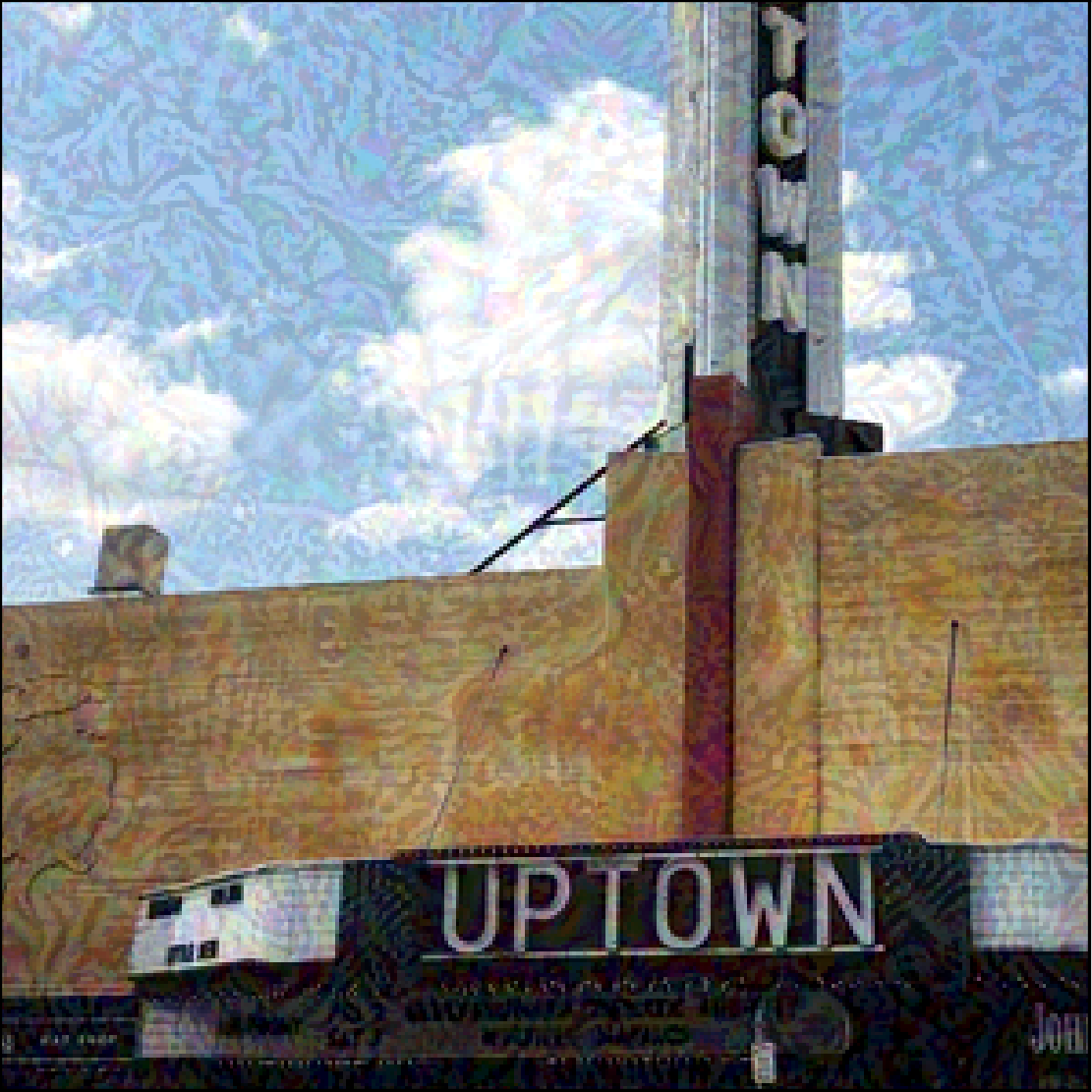} \quad & 
		\includegraphics[height=0.19\textwidth]{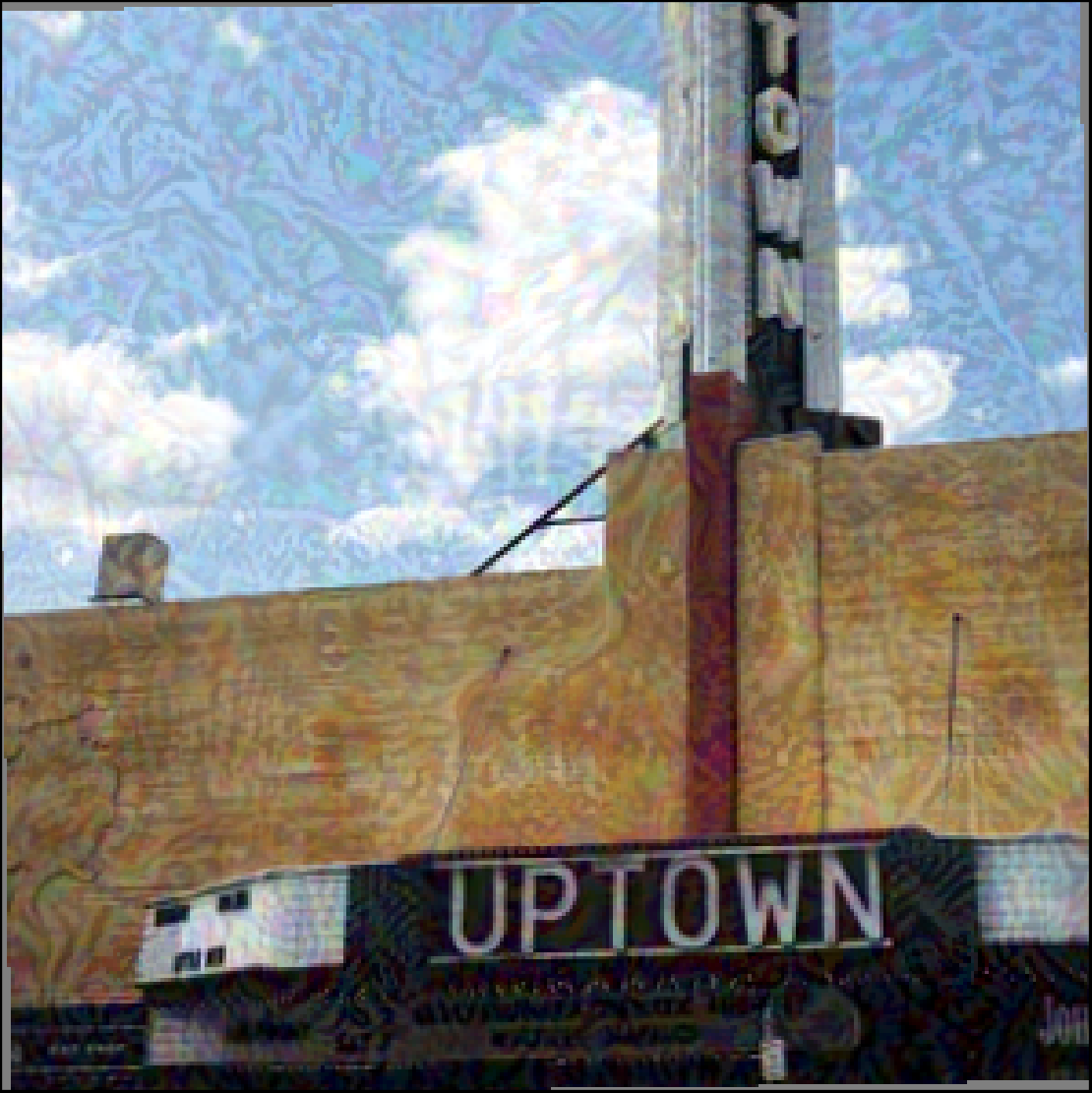} \quad &
		\includegraphics[height=0.19\textwidth]{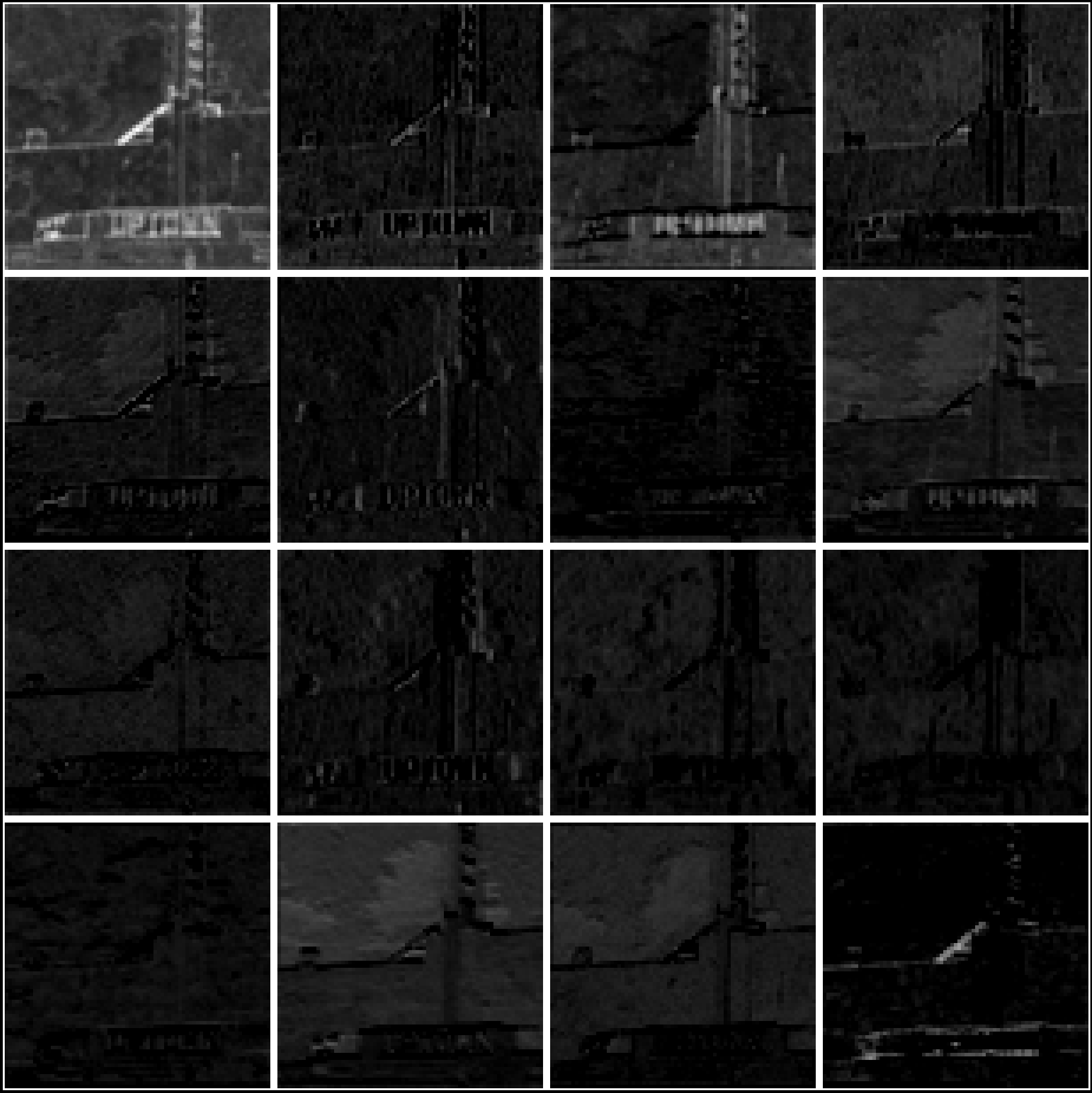} \quad &
		\includegraphics[height=0.19\textwidth]{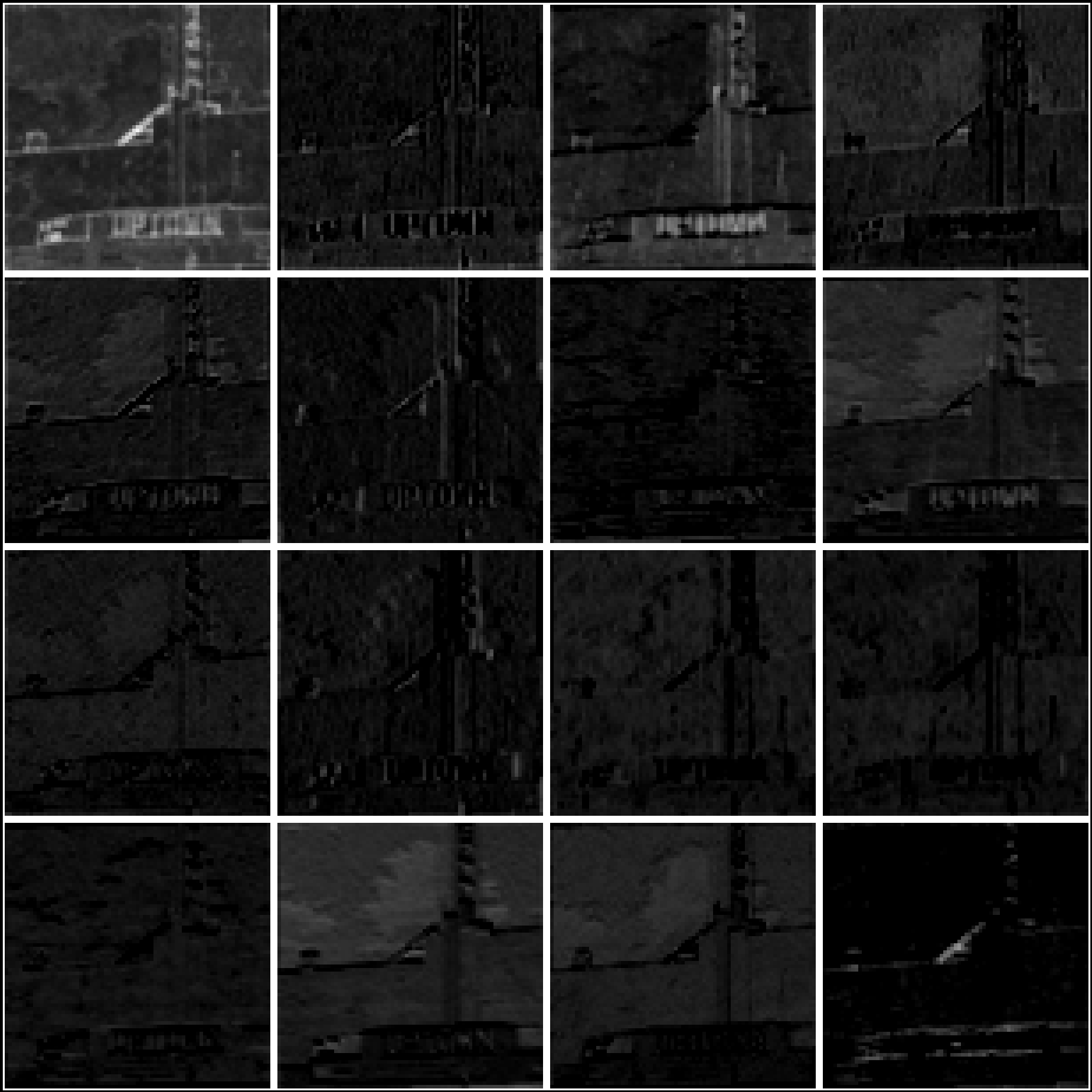} \quad &
		\includegraphics[height=0.19\textwidth]{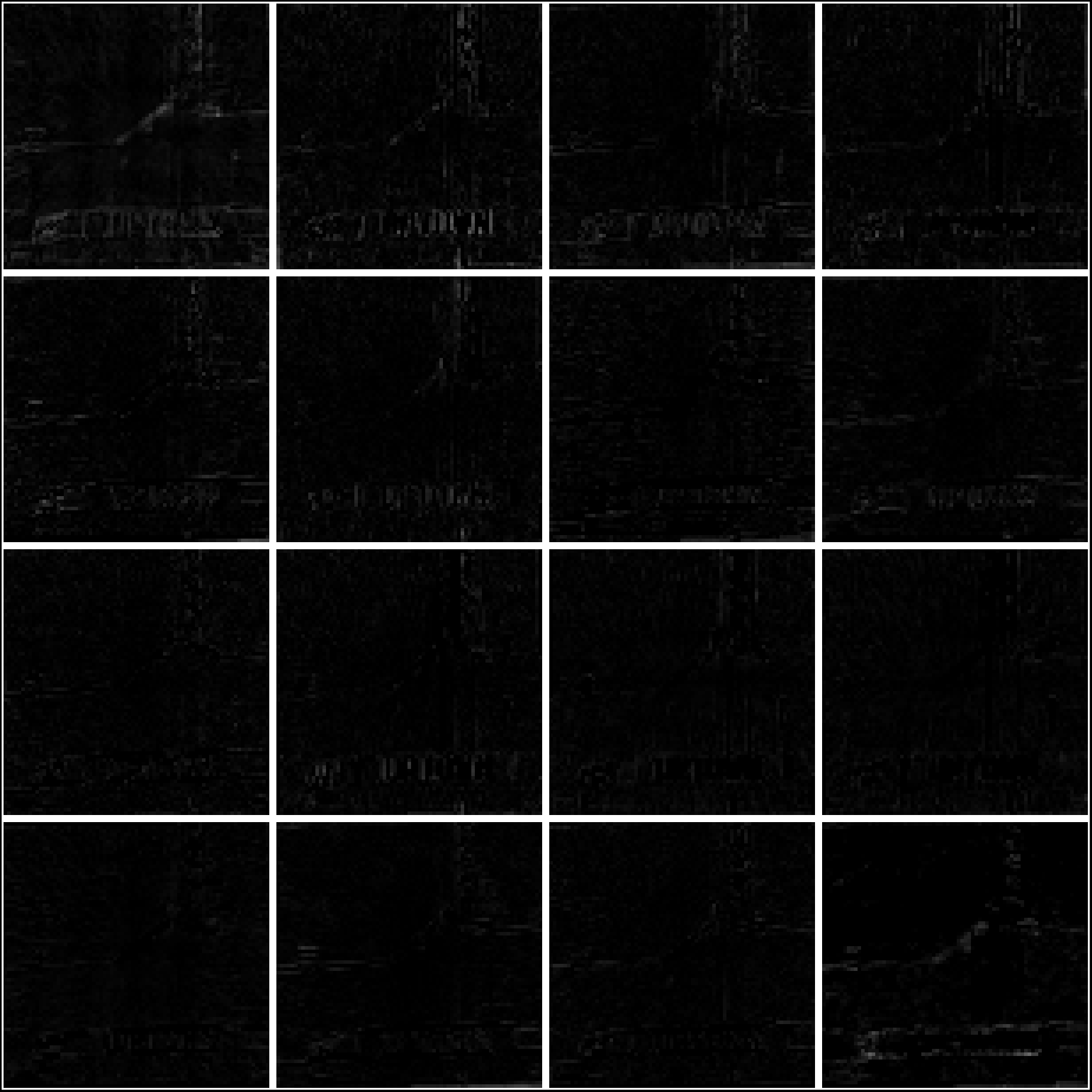} \\
	\end{tabular}
	\caption{Identify major feature fluctuations that potentially influence the decisions of Inc-v3's (upper row) and Res-101 (lower row) under the single model setting. The white-box model and the baseline are Inc-v4 and BSR, respectively. $(3)$ The upper feature maps are selected from the Conv2d-1a layer in Inc-v3, while the lower feature maps are from the Conv1 layer of the Block 1 in Res-101. Additional details can be found in Figure \ref{supp:fig:featureanalysis:dim:left_1:1}.}
	\label{supp:fig:featureanalysis:bsr:left_1}
\end{figure*}

\begin{figure*}[!h]
	\centering
	\begin{tabular}{@{}c@{}c@{}c@{}c@{}c@{}}
		\includegraphics[height=0.19\textwidth]{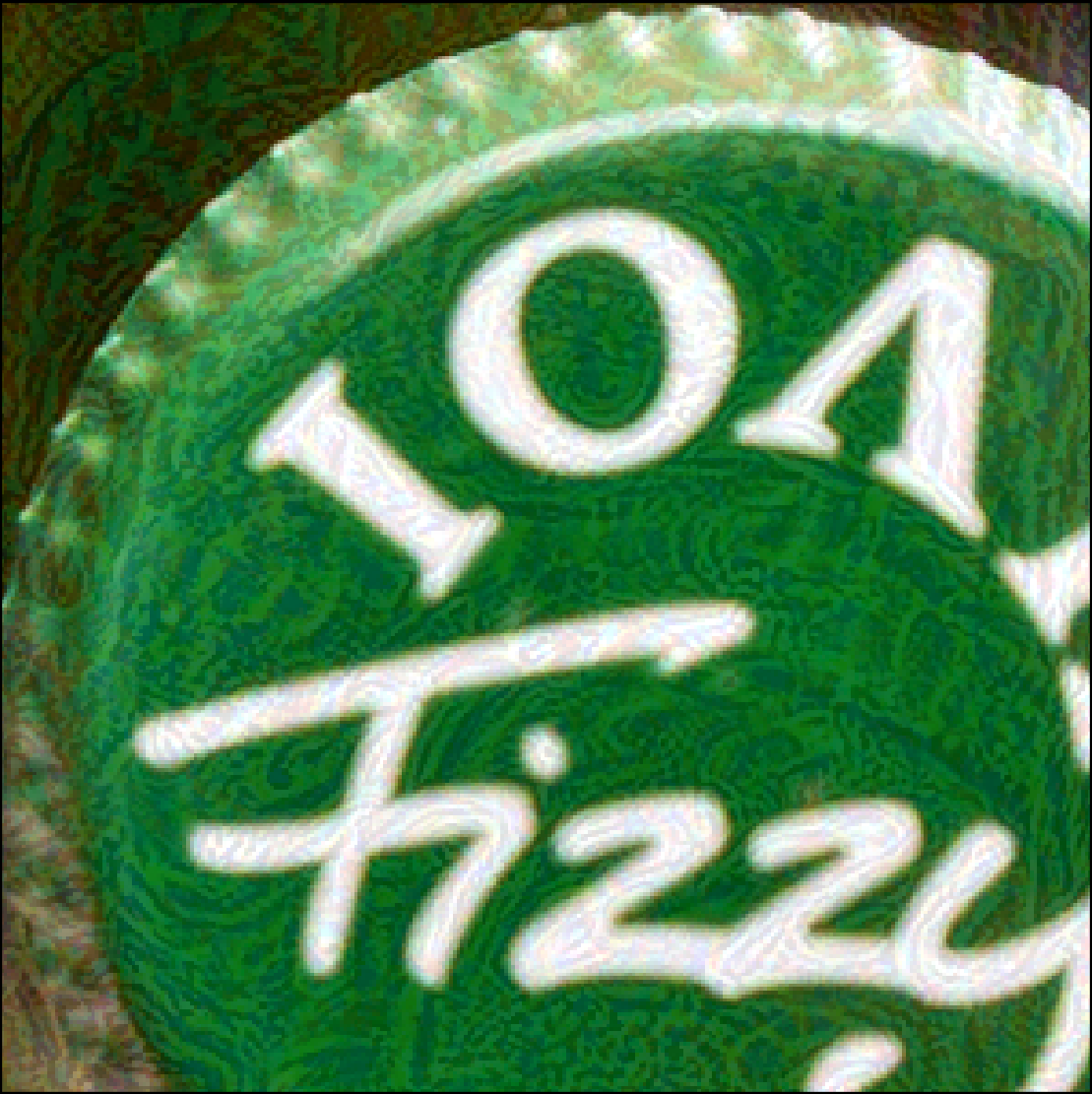} \quad & 
		\includegraphics[height=0.19\textwidth]{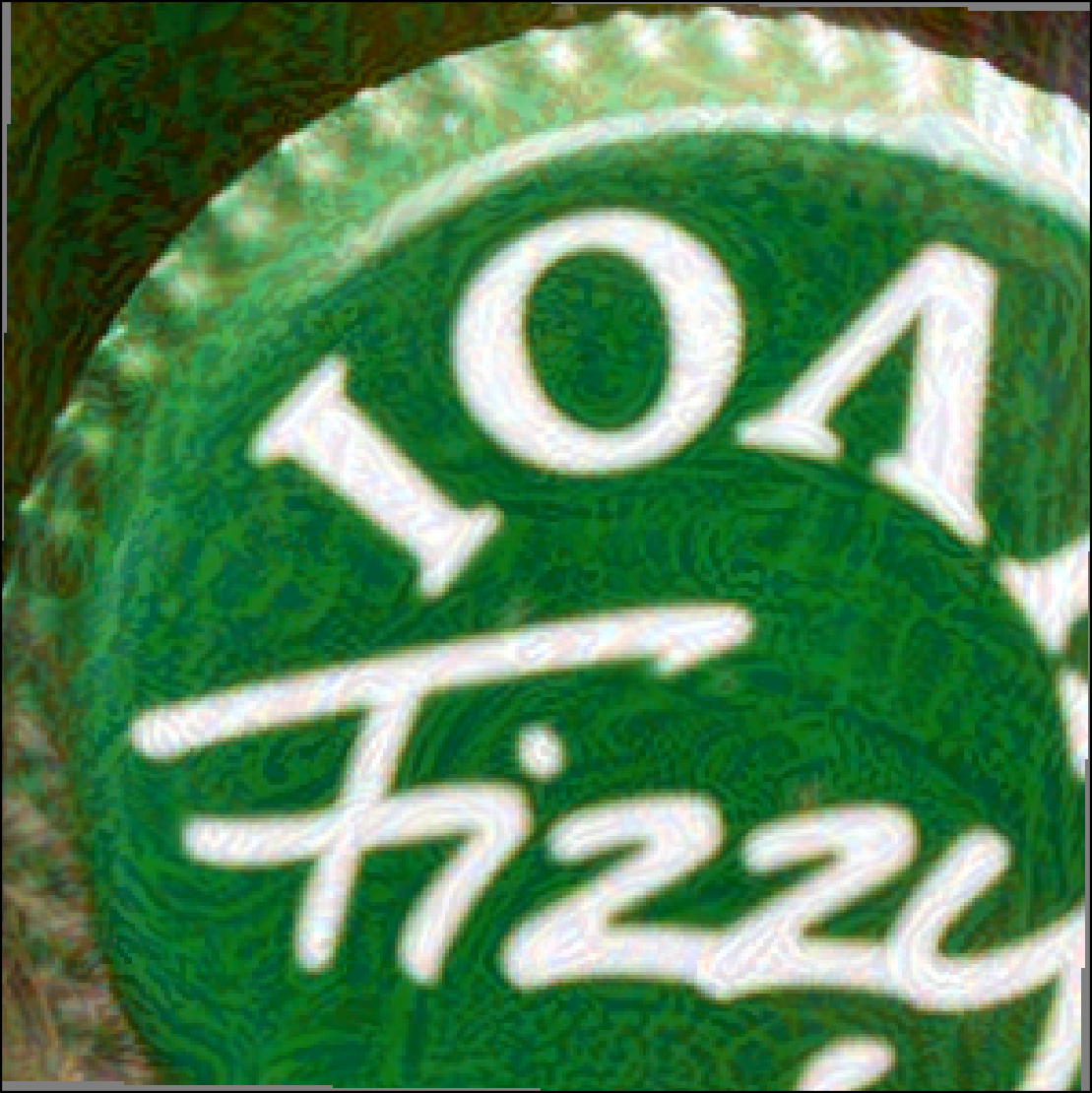} \quad &
		\includegraphics[height=0.19\textwidth]{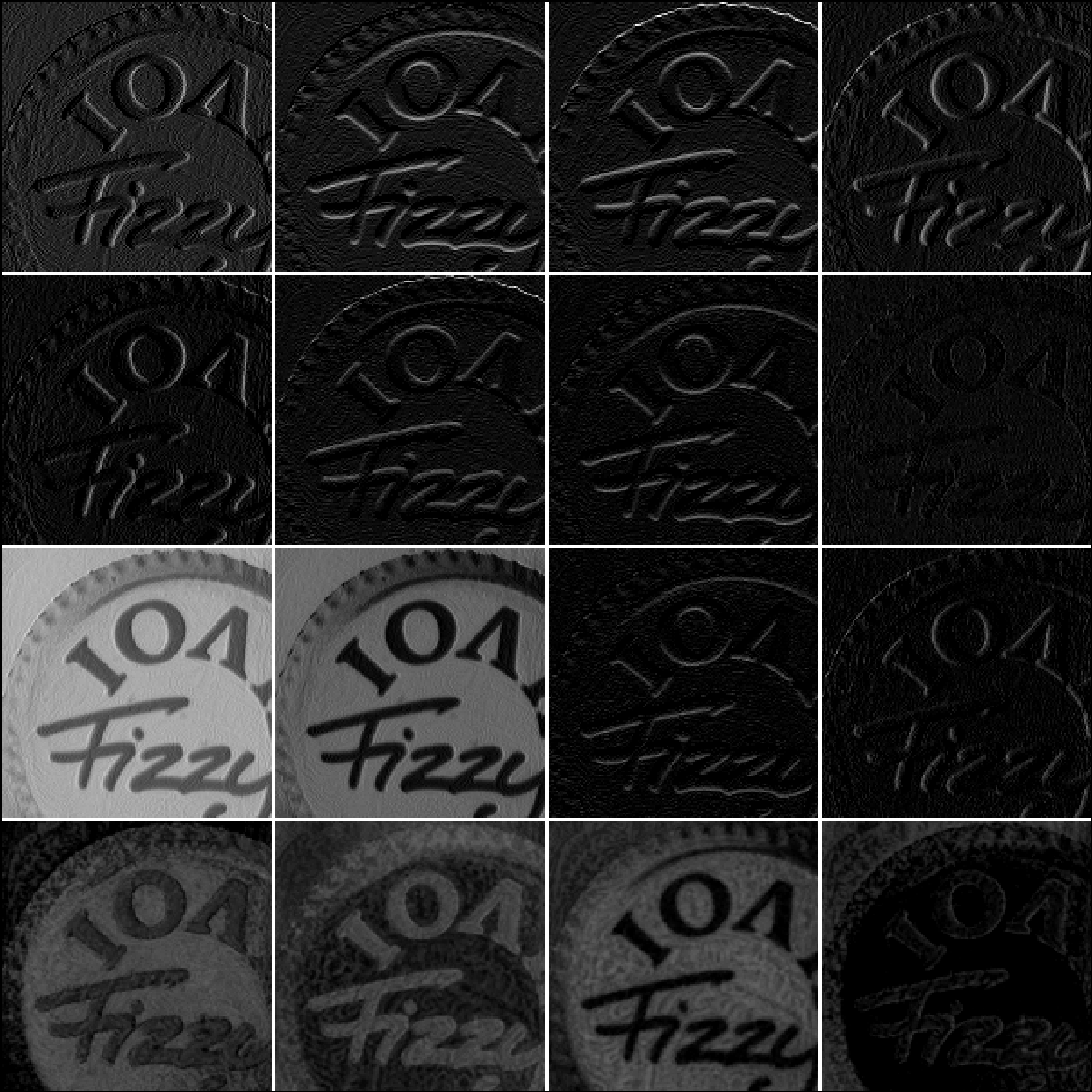} \quad &
		\includegraphics[height=0.19\textwidth]{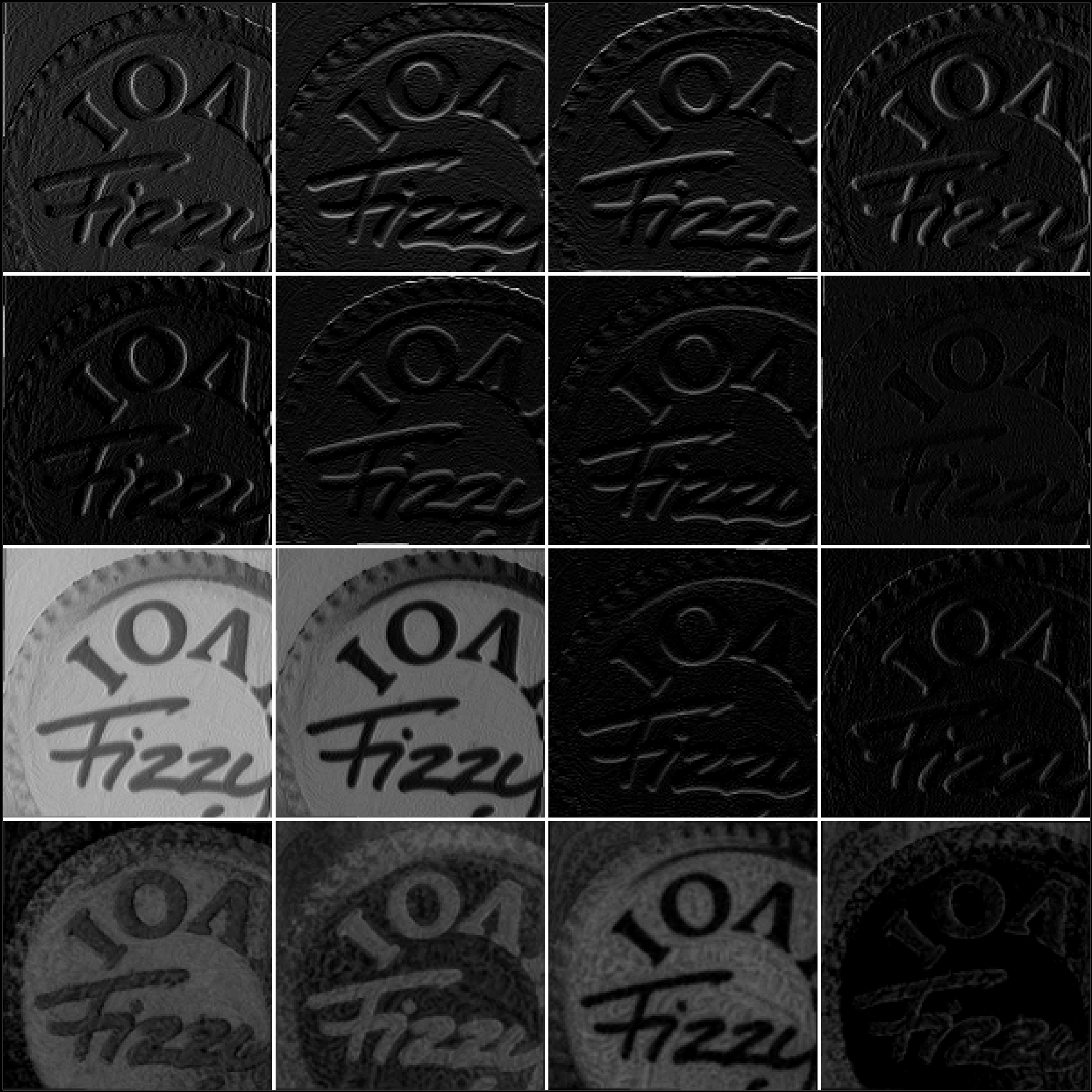} \quad &
		\includegraphics[height=0.19\textwidth]{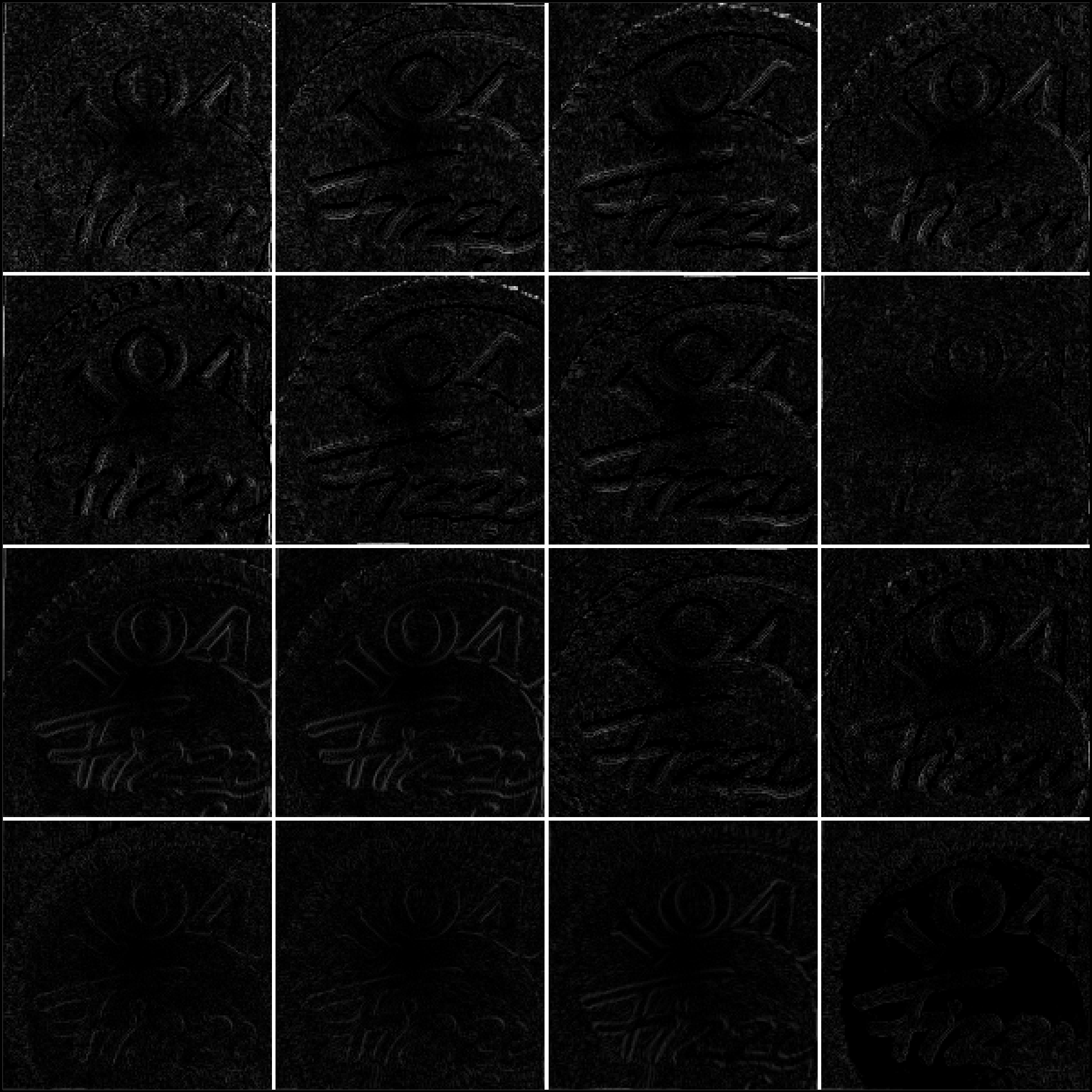} \\
		$(1)$ Advr & $(2)$ $1^\circ$ Right & $(3)$ FMs wo $1^\circ$ & $(4)$ FMs w $1^\circ$ & $(5)$ Diffs between $(3)$ \& $(4)$ \\
		\includegraphics[height=0.19\textwidth]{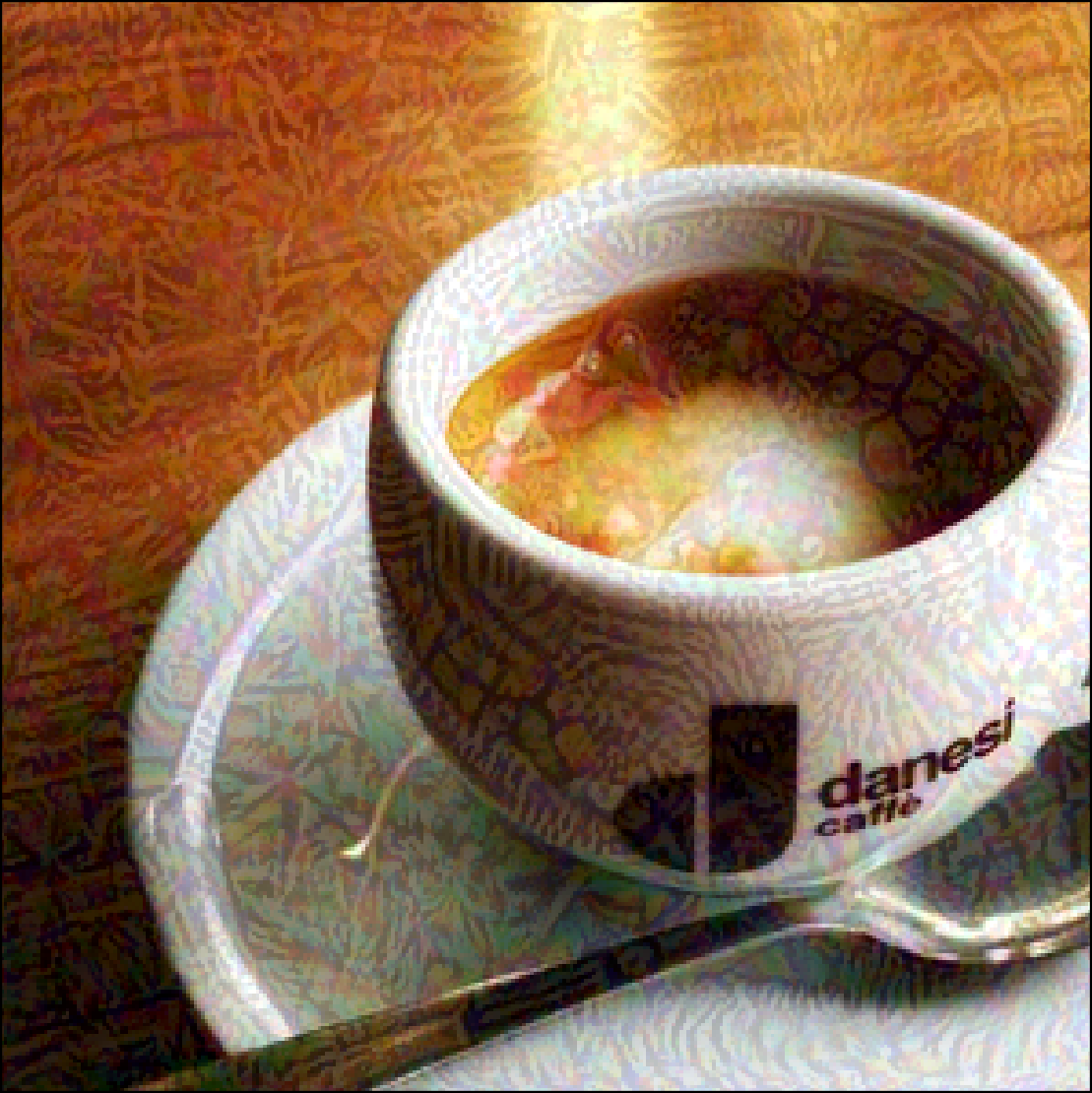} \quad & 
		\includegraphics[height=0.19\textwidth]{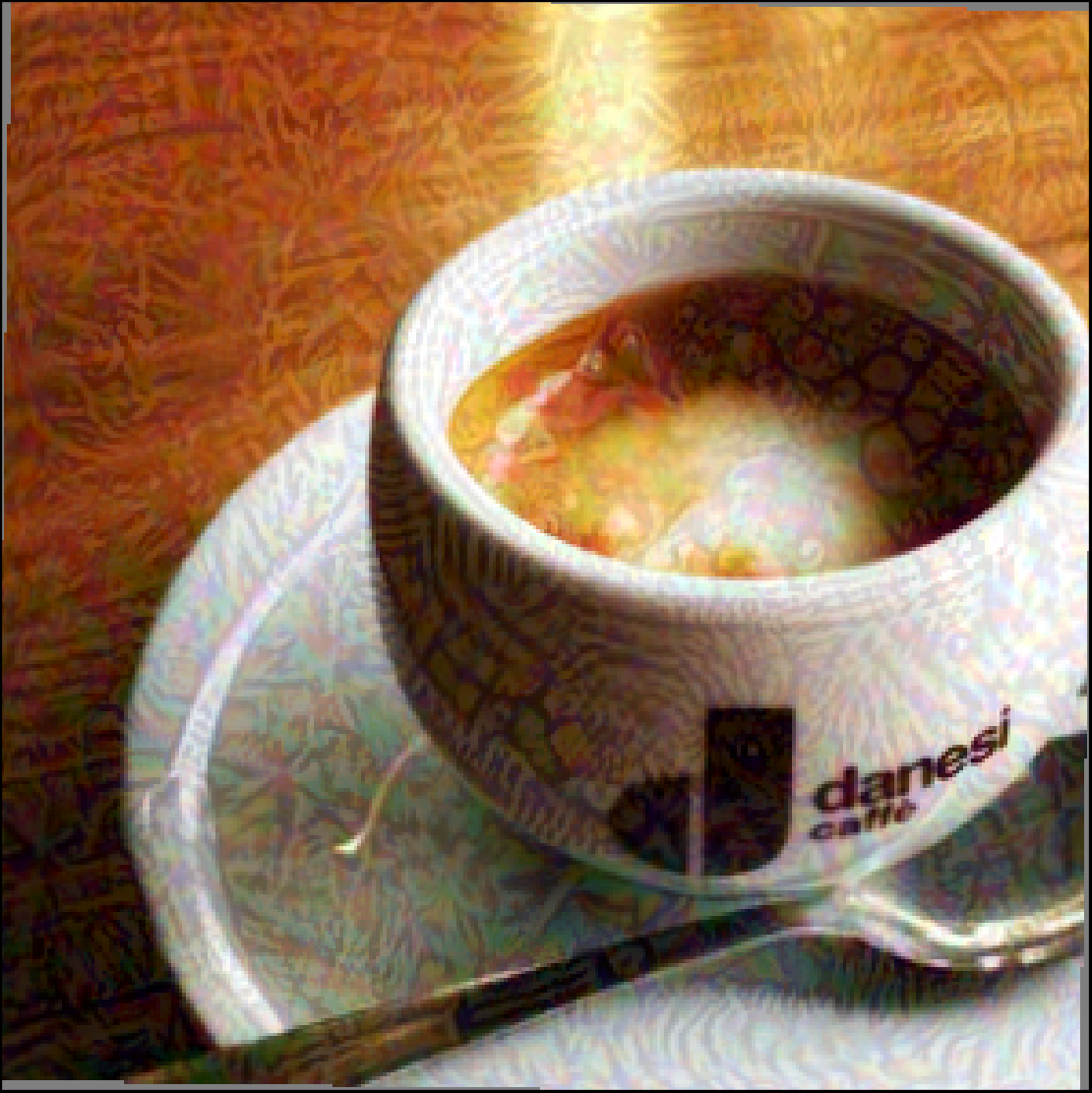} \quad &
		\includegraphics[height=0.19\textwidth]{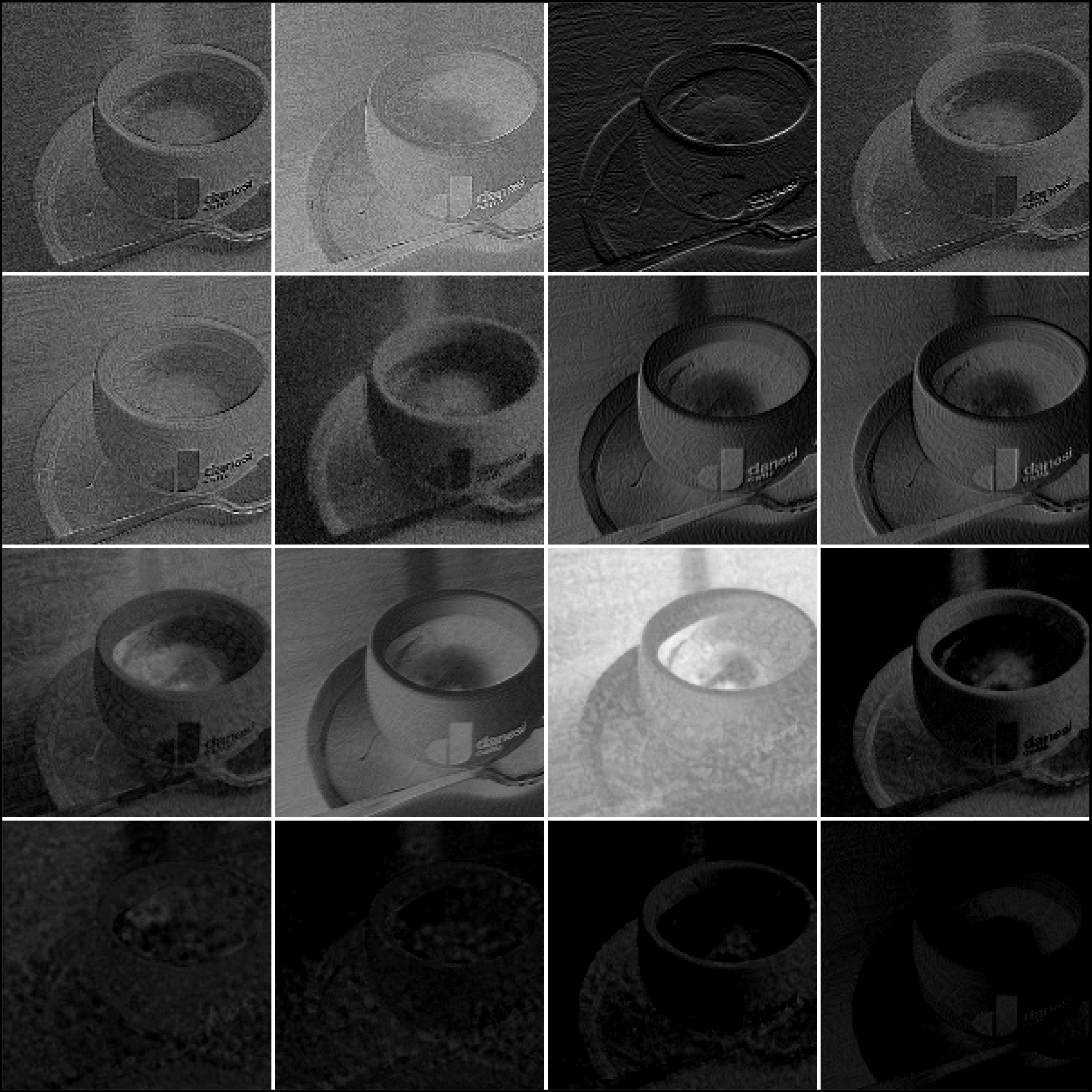} \quad &
		\includegraphics[height=0.19\textwidth]{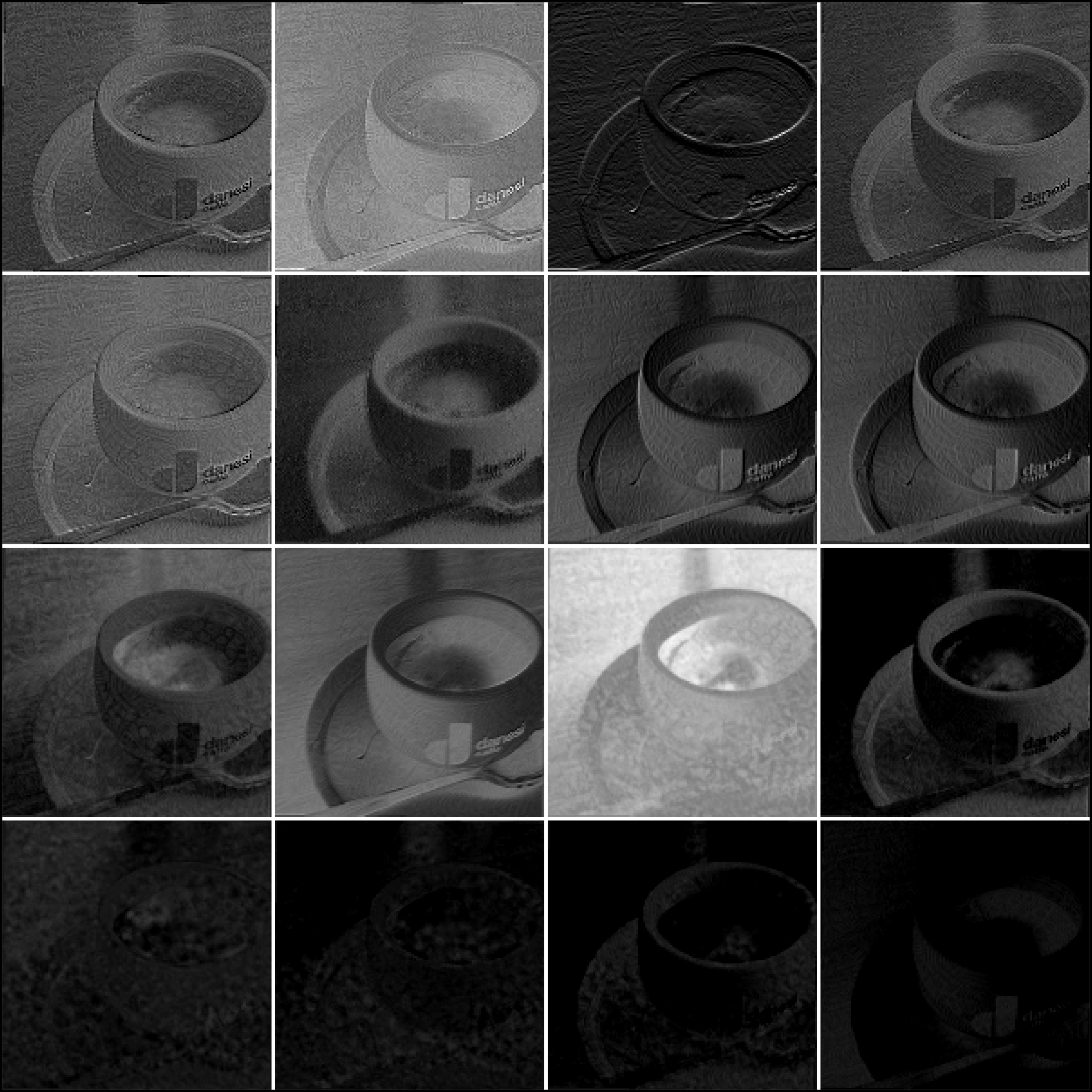} \quad &
		\includegraphics[height=0.19\textwidth]{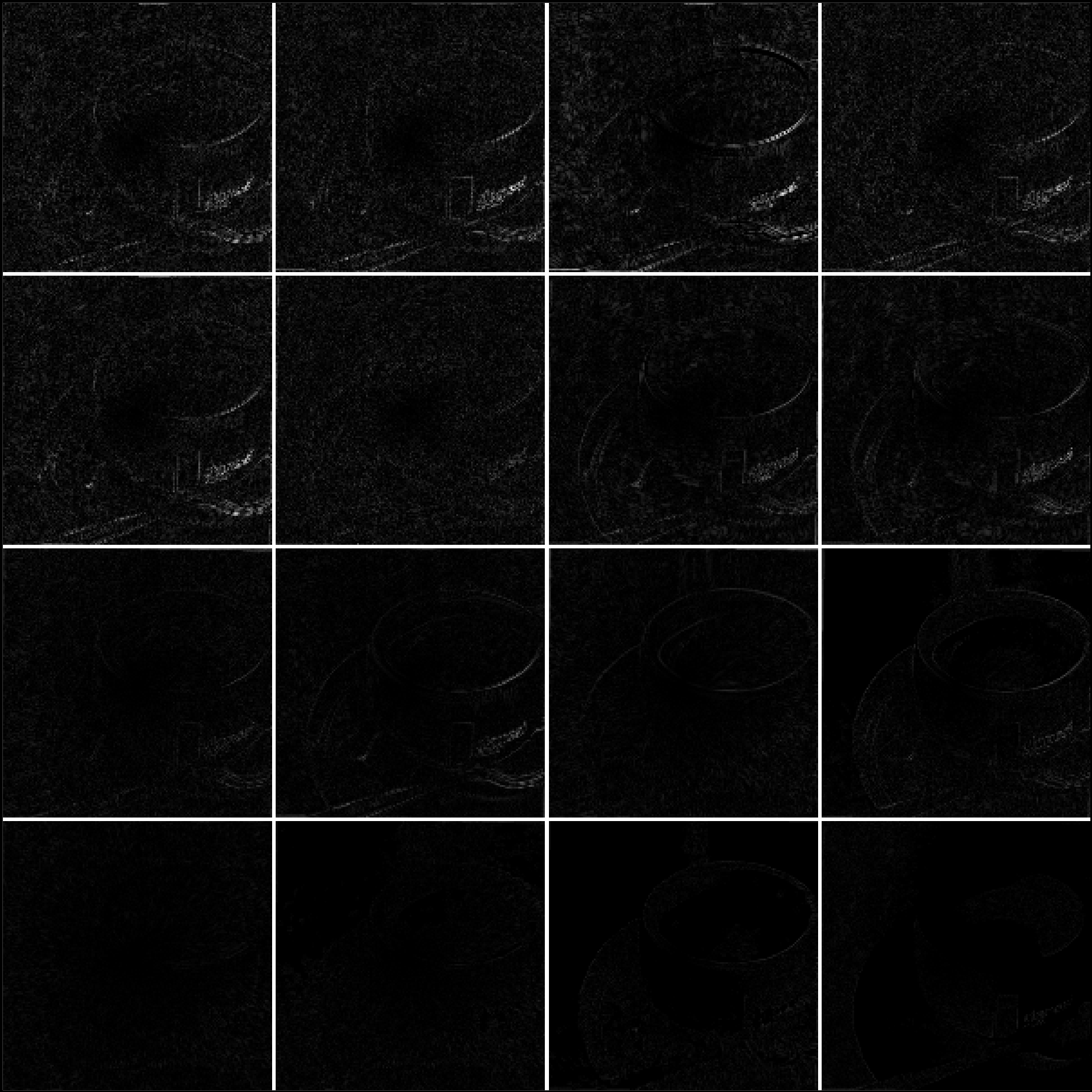} \\
	\end{tabular}
	\caption{Identify major feature fluctuations that potentially influence the decisions of IncRes-v2 (upper row) and Inc-v3$_{ens3}$ (lower row) under the single model setting. The white-box model and the baseline are consistent with those in Figure \ref{supp:fig:featureanalysis:bsr:left_1}. $(2)$ The $1^\circ$ right rotation of $(1)$. $(3)$ The upper feature maps are selected from the Conv2d-1a layer in Inc-v3$_{ens3}$, and the lower feature maps are from the Conv2d-1a layer in Inc-v3$_{ens3}$. Additional details can be found in Figure \ref{supp:fig:featureanalysis:dim:left_1:1}.}
	\label{supp:fig:featureanalysis:bsr:right_1}
\end{figure*}

Section \ref{sec:further_study:why_1_rotation_work} represented a typical set of subplots highlighting feature fluctuations that may influence the prediction of $\text{Inc-v3}_{ens3}$. Building on these findings, this section extends the analysis by following the experimental setup from Section \ref{sec:further_study:why_1_rotation_work} and visualizing additional subplots (Figures \ref{supp:fig:featureanalysis:dim:left_1:1}, \ref{supp:fig:featureanalysis:dim:right_1}, \ref{supp:fig:featureanalysis:bsr:left_1}, and \ref{supp:fig:featureanalysis:bsr:right_1}). These subplots are derived from experiments conducted under various white-box models and baseline attacks, providing a broader perspective on feature-level dynamics.

These figures demonstrate that subtle angle adjustments to an input typically lead to feature differences in a few channels of a DNN model. However, these differences are significant, as they originate from low-level layers and ultimately alter the model's predictions.

\section{Does a rotation angle have its optimal? (More evaluations)}
\label{supp:sec:optimal_rotation}
\begin{figure*}[!h]
	\centering
	\includegraphics[width=0.5\linewidth]{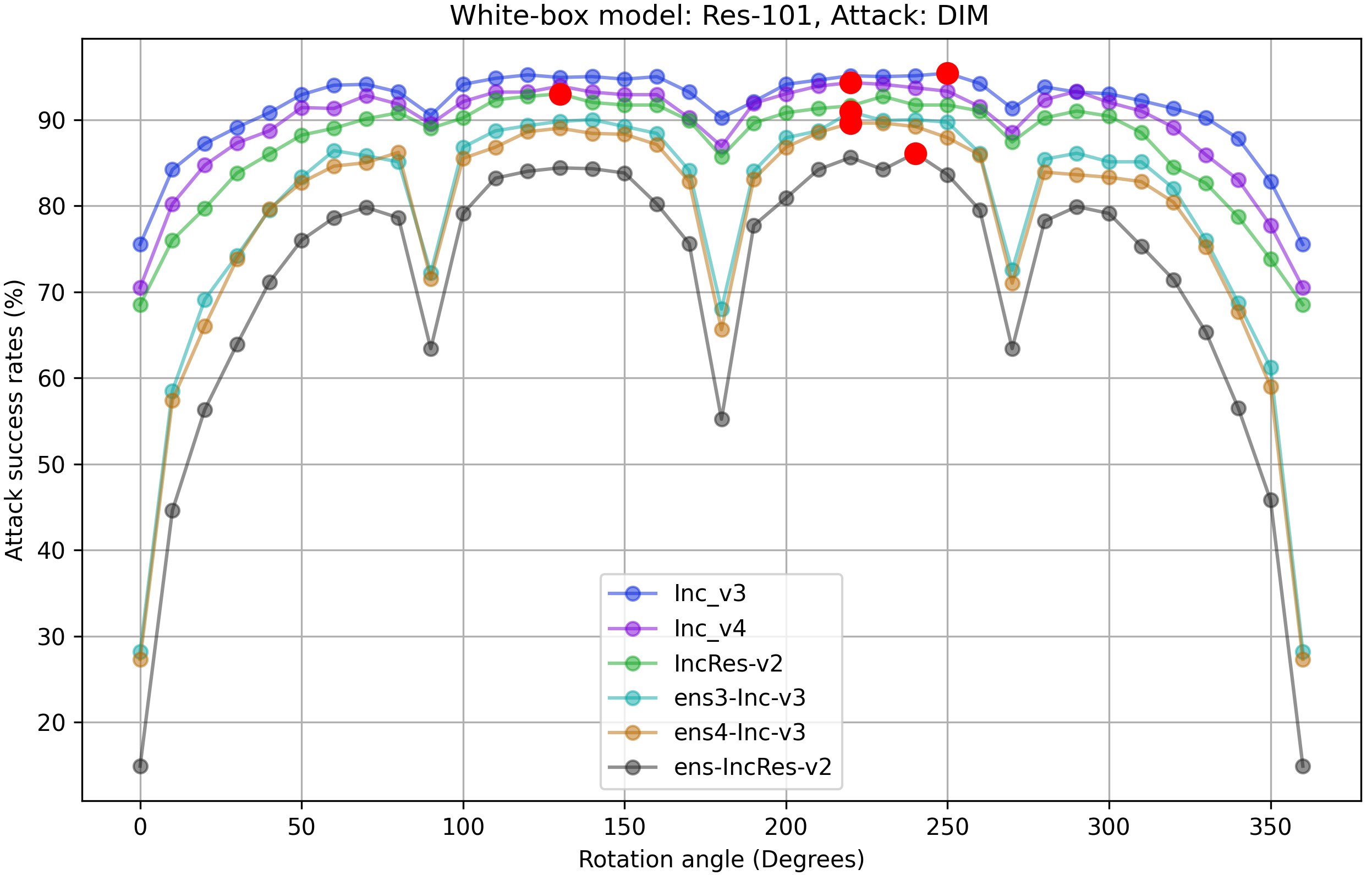}
	\caption{Identify the optimal rotation angle for maximizing transferability across six black-box models (NIPS'17). This figure is generated under the single-model setting with the Res-101 as the white-box model and DIM as the baseline attack. The x-axis represents the counter-clockwise rotation angle from $0^\circ$ to $360^\circ$, and the y-axis indicates the transferable attack success rates (\%). The six curves depict the fluctuations in attack success rates for each black-box model as the rotation angle varies. A red dot indicates the global maximum transferability on that curve.}
	\label{supp:fig:optimal_angle:NIPS17:Inc-v3_DIM}
\end{figure*}

\begin{figure*}[!h]
	\centering
	\begin{tabular}{@{}c@{}c@{}}
		\includegraphics[width=0.45\linewidth]{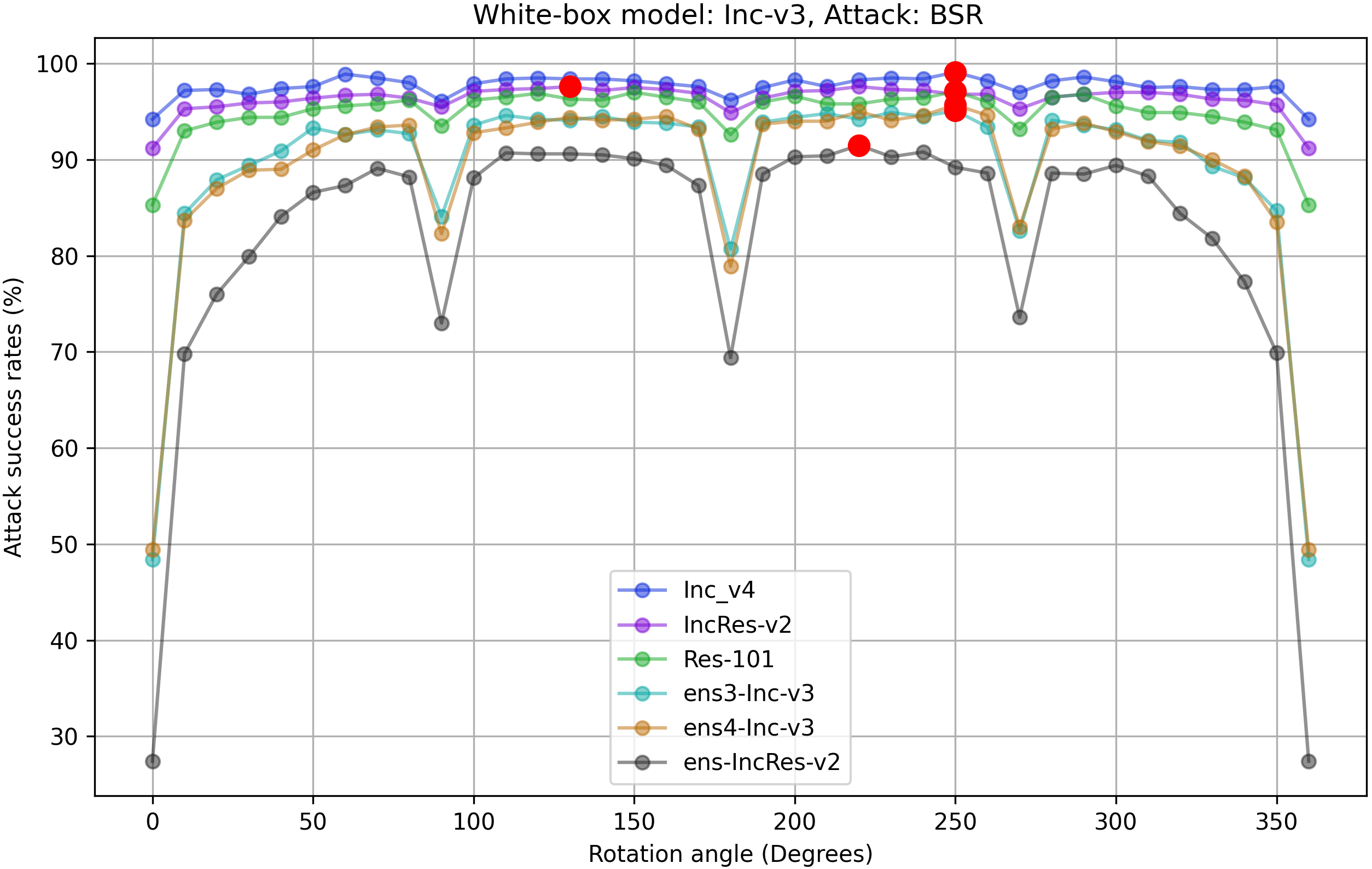} \quad \quad & \includegraphics[width=0.45\linewidth]{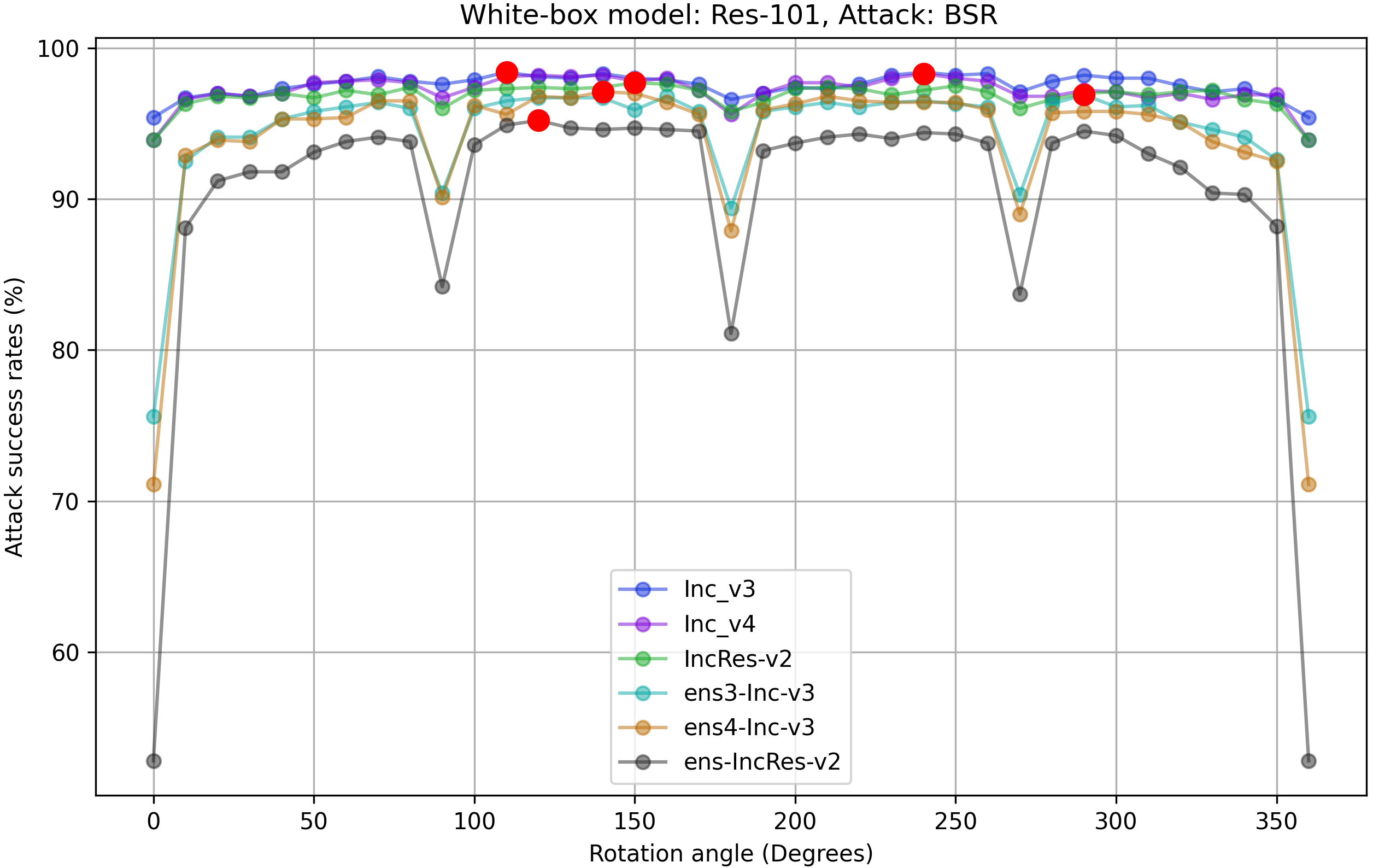} \\
		$(1)$ Inc-v3, BSR & $(2)$ Res-101, BSR \\ 
	\end{tabular}
	\caption{Identify the optimal rotation angle for maximizing transferability across six black-box models (NIPS'17). The left and right figures use the Inc-v3 and Res-101 as the white-box model, respectively. Both use BSR as the baseline attack. Additional details can be found in Figure \ref{supp:fig:optimal_angle:NIPS17:Inc-v3_DIM}.}
	\label{supp:fig:optimal_angle:NIPS17:Inc-v3_Res-101_BSR}
\end{figure*}

\begin{figure*}[!h]
	\centering
	\begin{tabular}{@{}c@{}c@{}}
		\includegraphics[width=0.45\linewidth]{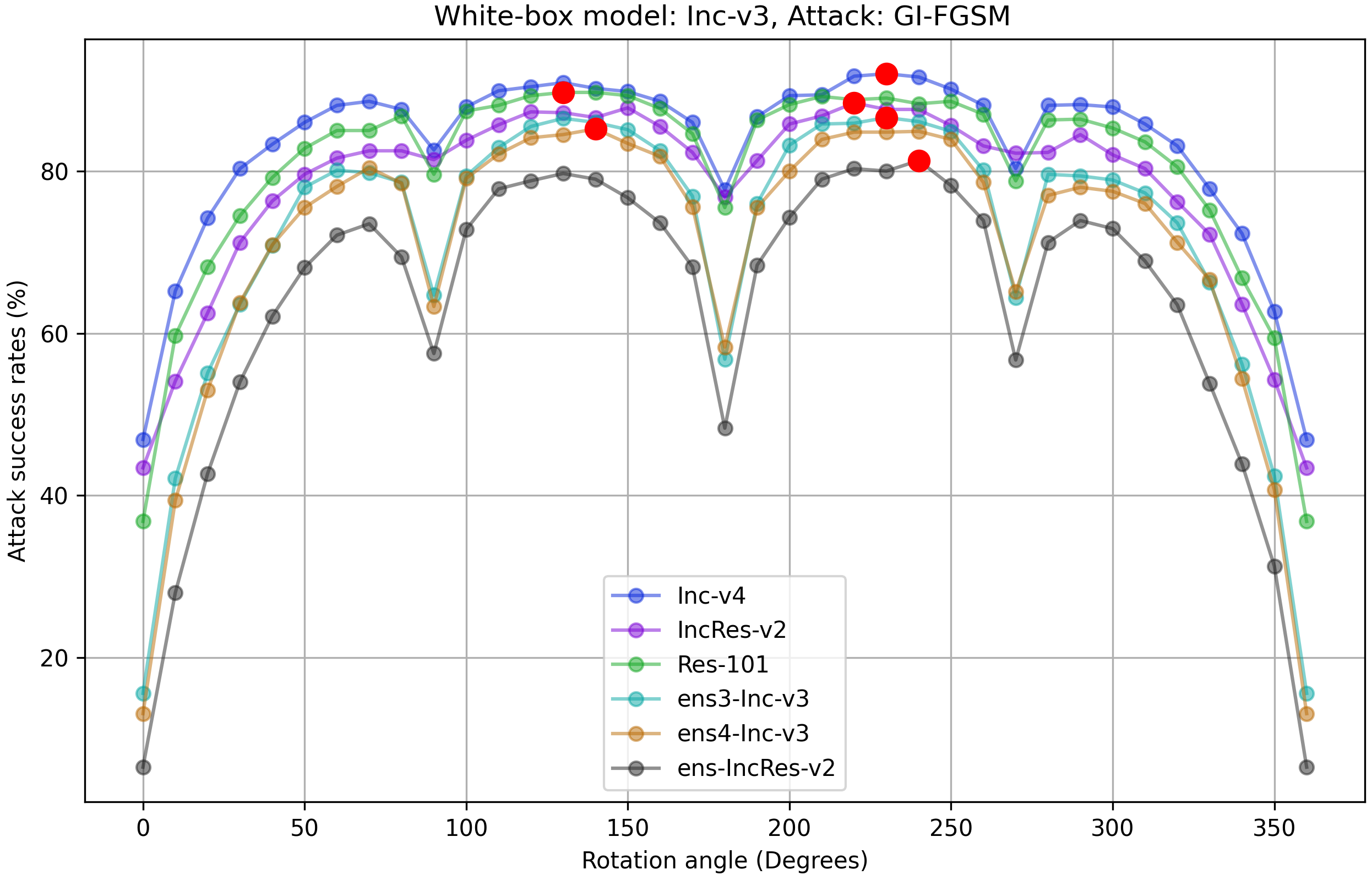} \quad \quad & \includegraphics[width=0.45\linewidth]{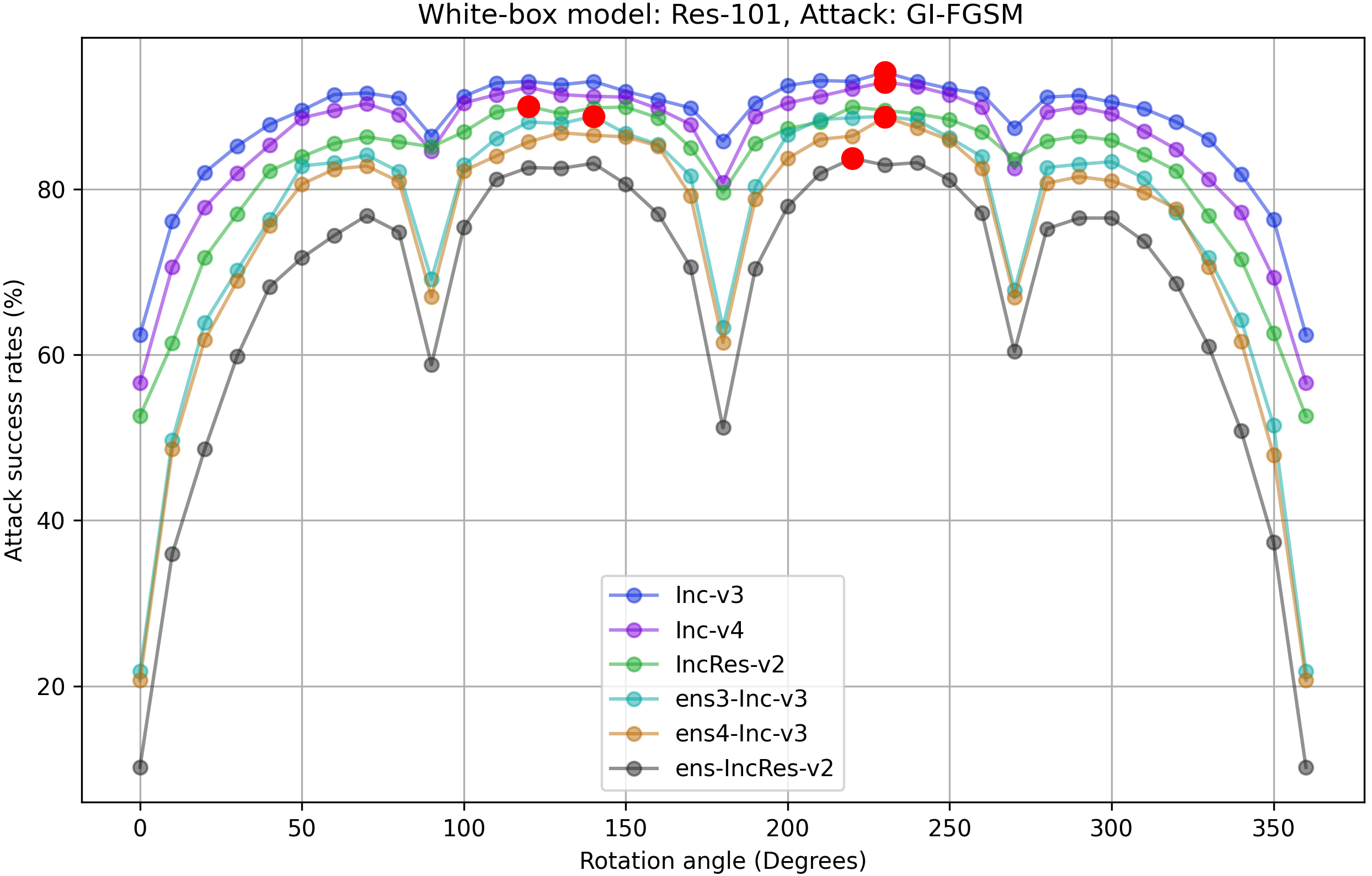} \\
		$(1)$ Inc-v3, GI-FGSM & $(2)$ Res-101, GI-FGSM \\ 
	\end{tabular}
	\caption{Identify the optimal rotation angle for maximizing transferability across six black-box models (NIPS'17). The left and right figures use the Inc-v3 and Res-101 as the white-box model, respectively. Both use GI-FGSM as the baseline attack. Additional details can be found in Figure \ref{supp:fig:optimal_angle:NIPS17:Inc-v3_DIM}.}
	\label{supp:fig:optimal_angle:NIPS17:Inc-v3_Res-101_GI-FGSM}
\end{figure*}

\begin{figure*}[!h]
	\centering
	\begin{tabular}{@{}c@{}c@{}}
		\includegraphics[width=0.45\linewidth]{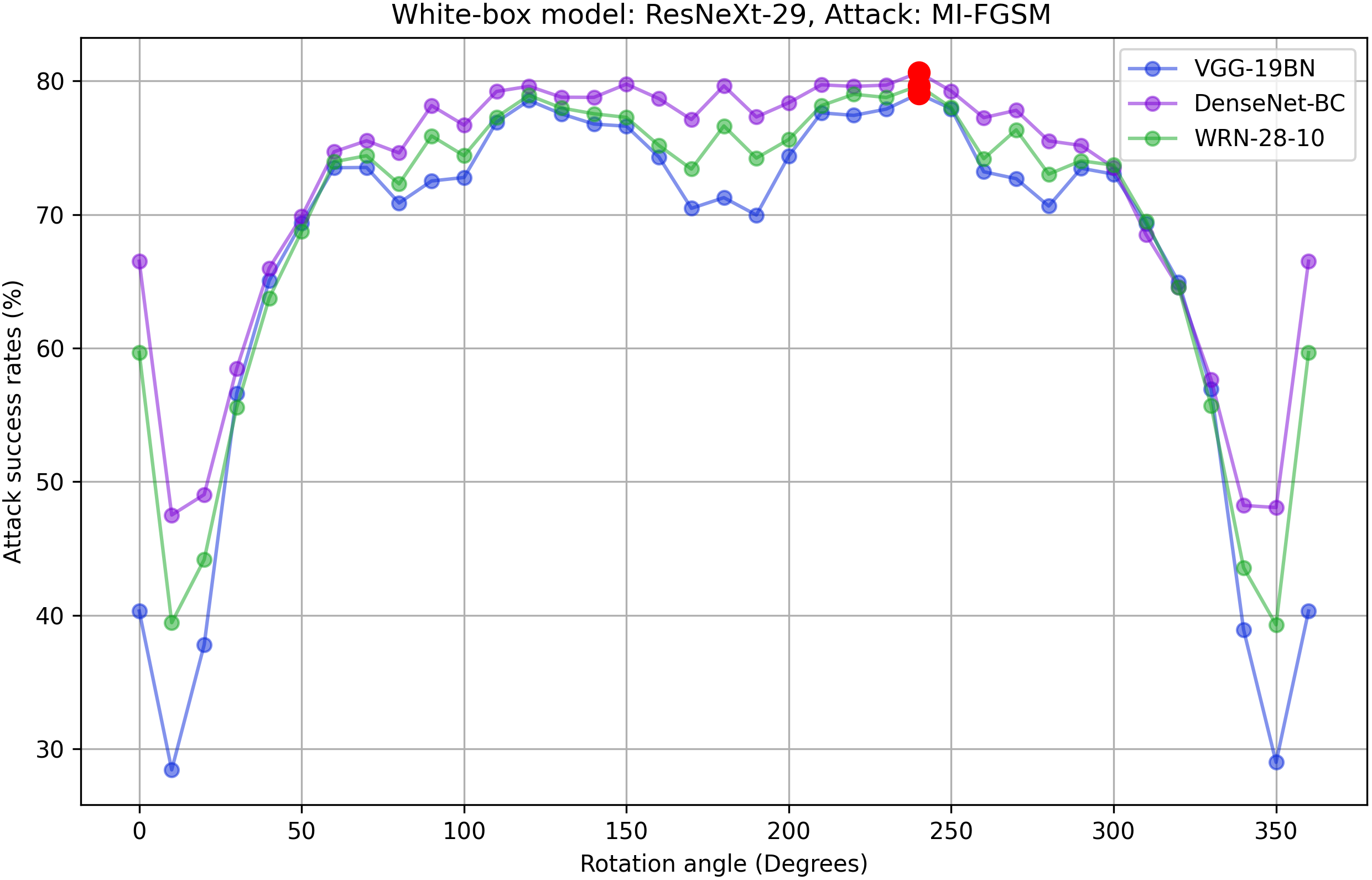} \quad \quad & \includegraphics[width=0.45\linewidth]{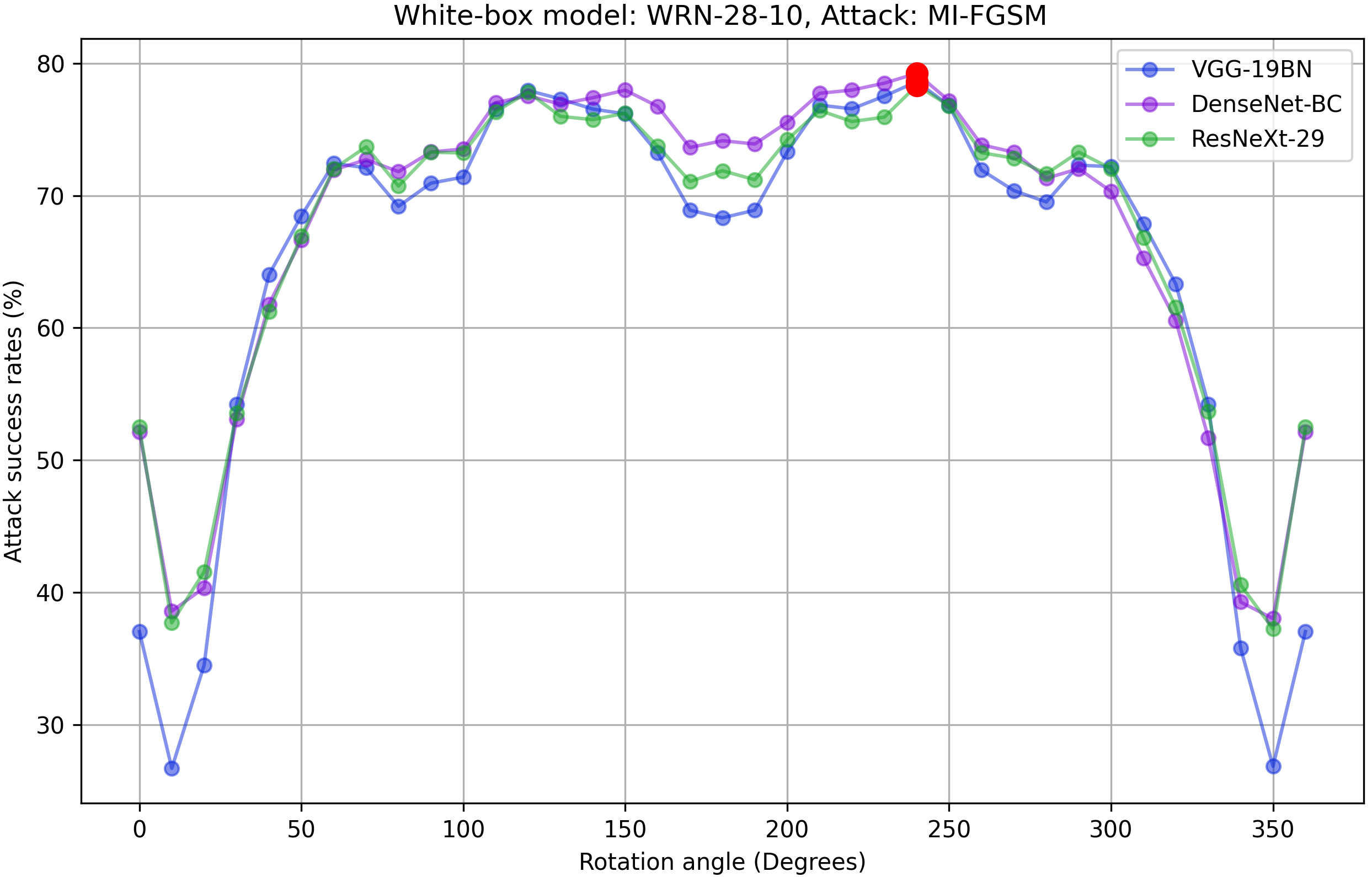} \\
		$(1)$ ResNeXt-29, MI-FGSM & $(2)$ WRN-28-10, MI-FGSM \\ 
	\end{tabular}
	\caption{Identify the optimal rotation angle for maximizing transferability across three black-box models (CIFAR-10). The left and right figures use the ResNeXt-29 and WRN-28-10 as the white-box model, respectively. Both use MI-FGSM as the baseline attack. The three curves depict the fluctuations in attack success rates for each black-box model as the rotation angle varies. A red dot indicates the global maximum transferability on that curve.}
	\label{supp:fig:optimal_angle:CIFAR10:Inc-v3_Res-101_MI-FGSM}
\end{figure*}

In Section \ref{sec:further_study:optimal_rotation}, we presented three representative subplots. Here, we extend the analysis by illustrating additional subplots (Figures \ref{supp:fig:optimal_angle:NIPS17:Inc-v3_DIM}, \ref{supp:fig:optimal_angle:NIPS17:Inc-v3_Res-101_BSR}, \ref{supp:fig:optimal_angle:NIPS17:Inc-v3_Res-101_GI-FGSM}, and \ref{supp:fig:optimal_angle:CIFAR10:Inc-v3_Res-101_MI-FGSM}) generated from various combinations of white-box models and baseline attacks. As in Section \ref{sec:further_study:optimal_rotation}, these subplots assume unlimited query conditions as a prerequisite for the experiments. For detailed findings and further context, please refer to Section \ref{sec:further_study:optimal_rotation}.

\end{document}